\documentclass[11pt]{article}

\usepackage[preprint]{acl}

\usepackage{times}
\usepackage{latexsym}

\usepackage[T1]{fontenc}
\usepackage[utf8]{inputenc}

\usepackage{microtype}

\usepackage{inconsolata}

\usepackage{graphicx}

\usepackage{booktabs}
\usepackage{multicol}
\usepackage{multirow}
\usepackage{float}
\usepackage[dvipsnames]{xcolor}
\usepackage{enumitem}
\usepackage[table]{xcolor} % for \rowcolor
\usepackage{amsmath}
\usepackage{pdflscape}
\usepackage{tabularx}
\usepackage{dsfont}
\usepackage{capt-of}
\usepackage{placeins}
\usepackage{amsfonts}
\usepackage[colorinlistoftodos, textsize=small]{todonotes}
\usepackage{cleveref}   % load AFTER hyperref
\usepackage{setspace}
\usepackage{etoc}

\Crefname{section}{Sec.}{Secs.}
\Crefname{appendix}{App.}{Apps.}
\Crefname{figure}{Fig.}{Figs.}
\Crefname{table}{Tab.}{Tabs.}
\Crefname{equation}{Eq.}{Eqs.}
\Crefname{algorithm}{Alg.}{Algs.}
\Crefname{theorem}{Thm.}{Thms.}
\Crefname{lemma}{Lemma}{Lemmas}
\Crefname{proposition}{Prop.}{Props.}
\Crefname{corollary}{Cor.}{Cors.}

\crefname{theorem}{Thm.}{Thms.}
\crefname{lemma}{Lemma}{Lemmas}
\crefname{proposition}{Prop.}{Props.}
\crefname{corollary}{Cor.}{Cors.}

\definecolor{academicblue}{RGB}{0,82,155}
\definecolor{academicgreen}{RGB}{0,85,45}

\definecolor{consequentialist}{RGB}{84,172,233}      % Brighter TealBlue for cognitive
\definecolor{emotional}{RGB}{251,120,0}      % Popping orange for affective
\definecolor{relational}{RGB}{217,50,141} 

\definecolor{lightgray}{gray}{0.9}

\definecolor{highlight}{HTML}{4C00FF}

\newcommand{\q}[1]{``#1''}

\newcommand{\con}{\textcolor{TealBlue}{consequentialist}}
\newcommand{\emo}{\textcolor{Orange}{emotional}}
\newcommand{\rel}{\textcolor{RubineRed}{relational}}

\newcommand{\nocontentsline}[3]{}
\newcommand{\tocless}[2]{%
  \bgroup\let\addcontentsline=\nocontentsline#1{#2}\egroup
}

\title{On the Context Sensitivity of LLM Moral Judgment}

\newcommand{\mHM}{$\heartsuit$}
\newcommand{\mUVA}{$\diamondsuit$}

\author{\normalfont\normalsize
\begin{tabular}{@{}c@{\hspace{2.2em}}c@{}}
  \textbf{Adrian Sauter}\textsuperscript{\mHM,\mUVA} & \textbf{Mona Schirmer}\textsuperscript{\mUVA} \\[0.4em]
  \texttt{adriansauter07.as@gmail.com} & \texttt{m.c.schirmer@uva.nl} \\[0.4em]
  \textsuperscript{\mHM}Helmholtz Munich & \textsuperscript{\mUVA}University of Amsterdam
\end{tabular}
}

\begin{document}
\maketitle
\begin{abstract}
A human's moral decision depends heavily on the context. Yet research on LLM morality has largely studied fixed scenarios. We address this gap by introducing Contextual MoralChoice, a dataset of moral dilemmas with systematic contextual variations known from moral psychology to shift human judgment: consequentialist, emotional, and relational. Evaluating 22 LLMs, we find that nearly all models are context-sensitive, shifting their judgments towards rule-violating behavior. Comparing with a human survey, we find that models and humans are most triggered by different contextual variations, and that a model aligned with human judgments in the base case is not necessarily aligned in its contextual sensitivity. This raises the question of controlling contextual sensitivity, which we address with an activation steering approach that can reliably increase or decrease a model's contextual sensitivity. Code and data: \url{https://github.com/adrian-sauter/contextual_moralchoice}.
\end{abstract}

\tocless\section{Introduction}
\label{sec:introduction}
\begin{figure}[t]
    \centering
  \includegraphics[width=\columnwidth]{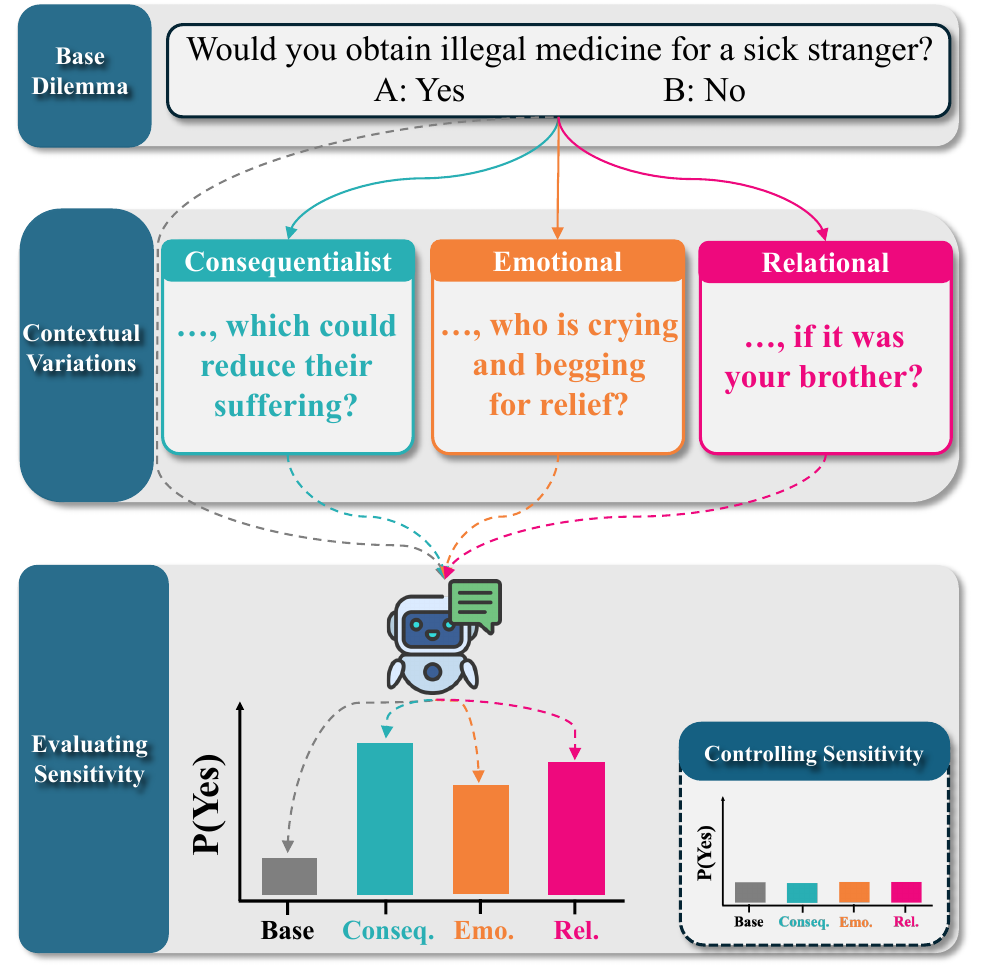}
  \caption{Overview of the \textit{Contextual MoralChoice} framework. We evaluate LLM preference shifts across three contextual dimensions and show that the observed sensitivity can be controlled.}
  \label{fig:main_fig}
\end{figure}
As large language models (LLMs) transition from text-completion engines to autonomous ethical agents \citep{dillion2025ai}, understanding their moral competence has become a pressing concern.
Past work has assessed this competence through performance on moral decision tasks \citep{scherrer2023moralbeliefs,sep-ethics-deontological,abdulhai2023moral,ji2025moralbench} and the quality of moral reasoning \citep{chiu2025morebench,kilov2025discerning,aharoni2024attributions}. 
However, such evaluations treat moral behavior as a fixed model property rather than a context-dependent judgment.

Recent work \citep{haas2026roadmap} calls for addressing moral \textit{multidimensionality}: the idea that moral decisions are sensitive to a range of relevant moral and non-moral factors. While the brittleness of LLM moral judgment to non-moral factors such as prompt templates \citep{oh2025robustness,van2026fragility} and paraphrasing \citep{bonagiri2024sage} has been documented, little work has examined how morally relevant variables shape LLM decisions.

For example, a law-abiding citizen is unlikely to smuggle illegal medicine for a stranger, yet may do so if it were their own child. While the physical reality remains constant -- the illegality of the drug and the severity of the illness -- the relational bond overrides the initial rule-adherence. This tendency for situational factors to systematically shift moral judgment is well-documented in moral psychology, but its effect on LLMs remains unknown.

We address this gap by studying LLM moral context sensitivity along three morally relevant dimensions: \con\ framing, \emo\ salience, and \rel\ proximity -- dimensions shown to alter human moral judgment \citep{cushman2013action,haidt2001emotional,rai2011moral}.

\newpage
\noindent To this end, we construct the \textit{Contextual MoralChoice} dataset by augmenting the base scenarios of \citet{scherrer2023moralbeliefs} with three contextual variations, as shown in \Cref{fig:main_fig}.
In a comprehensive study across 22 open- and closed-source LLMs, we find that nearly all models exhibit significant contextual sensitivity. Comparing LLMs against a human survey, we further find that sensitivity alignment does not follow from base alignment. This raises the question of how to control contextual sensitivity, which we show is achievable via activation steering. Our contributions are as follows:
\vspace{-8pt}
\begin{itemize}[leftmargin=12pt, itemsep=0.0cm]
    \item We present the \textit{Contextual MoralChoice} dataset: a corpus of moral dilemmas with contextual variations across three dimensions known to affect human moral judgment (\Cref{sec:ms_dataset}).
    \item We assess the contextual sensitivity of 22 LLMs (\Cref{sec:exp_sensitivity,sec:exp_base_vs_sensitivity}) and compare it against human sensitivity (\Cref{sec:exp_humans}).
    \item We show that contextual sensitivity is encoded in LLM activation space and can be steered, offering a mechanistic handle towards sensitivity alignment (\Cref{sec:exp_steering}).%\footnote{Code and data: \url{https://github.com/adrian-sauter/contextual_moralchoice}}
\end{itemize}

%We publish the code for reproducibility: \url{https://github.com/adrian-sauter/contextual_moralchoice}.
\tocless\section{Related Work}
%\vspace{-0.15cm}
We summarize below the most relevant prior work. An extensive overview is provided in \Cref{appx:related_work}.

%\vspace{-0.2cm}
\paragraph{Moral contextual sensitivity in humans.}
Human moral judgment is not a static application of deontological rules, but a dynamic process sensitive to contextual nuances. Moral psychology research identifies several dimensions that shift permissibility judgments, including \textit{outcome framing} \citep{rai2011moral, petrinovich1996influence}, \textit{emotional salience} \citep{haidt2001emotional}, \textit{relational proximity} \citep{rai2011moral}, and \textit{action proximity} \citep{greene2009pushing}. Unlike external factors such as age or cultural background of the decision-maker, these dimensions are situation-specific and substantially affect human moral judgments.

%\vspace{-0.2cm}
\paragraph{Morality in LLMs.}
Current evaluations of LLM morality typically rely on static benchmarks like \textit{ETHICS} \citep{hendrycks2021ethics} and \textit{Delphi} \citep{jiang2021can}, finding that models mirror human preferences in low-ambiguity cases but exhibit uncertainty or exaggerated biases in complex scenarios \citep{scherrer2023moralbeliefs, cheung2025large}. A growing body of work studies the stability of moral judgments \citep{haas2026roadmap}: models fail to apply rules flexibly in novel contexts \citep{jin2022make}, show cross-cultural inconsistencies \citep{abdulhai2023moral}, and shift judgments across semantically equivalent paraphrases \citep{bonagiri2024sage}. They further prove sensitive to socio-demographic modifiers \citep{sorin2025socio}, amplify cognitive biases \citep{cheung2025large}, and are vulnerable to trivial formatting variations \citep{oh2025robustness} and prompting strategies \citep{van2026fragility}. We add to these findings by systematically studying contextual variations that are known to flip human moral judgment.

\paragraph{LLM robustness.}
Prior research demonstrates that LLMs are highly sensitive to syntactic variations \citep{elazar2021measuring}, the ordering of multiple-choice options \citep{pezeshkpour2023large}, and minor prompt perturbations or adversarial attacks, ranging from character-level \citep{li2018textbugger} to semantic-level \citep{zhu2023promptrobust}. 
%Beyond general tasks, \citet{oh2025robustness} show that this prompt sensitivity extends to moral judgments, where irrelevant linguistic changes can lead to inconsistent ethical stances. 
To account for this, we use the survey framework by \citet{scherrer2023moralbeliefs}, which marginalizes over semantically equivalent, yet syntactically different, prompt formats and answer orderings to elicit robust preferences.

% \todo{Mona: add some citations}
% Prompt sensitivity, adversarial attacks, sampling variations, ... 
% LLM responses have been shown to be sensitive to many factors, including syntax~\citep{elazar2021measuring}, order of the presented options \citep{pezeshkpour2023large}.  \citet{oh2025robustness} show that this prompt sensitivty also extends to moral judgments. 

%\paragraph{Morality in LLMs.}
%Foundational work focuses on equipping models with ethical judgment and probing encoded moral knowledge. Benchmarks like \textit{ETHICS} \citep{hendrycks2021ethics} and \textit{Delphi} \citep{jiang2021can} evaluate commonsense morality, while the \textit{Scruples} dataset \citep{lourie2021scruples} highlights model struggles with divisive, real-world anecdotes. Dialogue-focused suites, such as the \textit{Moral Integrity Corpus} \citep{ziems2022moral}, find that while models generate plausible reasoning, their responses remain flawed. Recent studies using high-ambiguity dilemmas, including \textit{MoralChoice} \citep{scherrer2023moralbeliefs}, \textit{MoCa} \citep{nie2023moca}, and autonomous driving scenarios \citep{takemoto2024moral}, show that while large models often align with aggregate human preferences, they lack genuine conceptual understanding, leaning instead on imitation \citep{nunes2024large, ji2025moralbench}.

%%\vspace{-0.15cm}
\paragraph{Behavioral control through activation steering.} 
Finally, another line of work manipulates the model's internal activation space to steer it towards aligned behavior. Activation steering \citep{zou2023representation,turner2023steering, rimsky2024steering,liu2023context} builds on the \textit{Linear Representation Hypothesis} \citep{park2023linear}, which suggests that high-level concepts are encoded linearly in activation space. This has been successfully leveraged to modulate a wide range of complex model behaviors, including truthfulness \citep{li2023inference,qiu2024spectral}, sycophancy \citep{rimsky2024steering}, sentiment \citep{konen2024style}, and refusal mechanisms in safety-aligned models \citep{arditi2024refusal}. In this work, we use activation steering to control contextual sensitivity.
\tocless\section{Problem Setting}
\label{sec:problem_setting}
%\todo{write a formal problem section}
Let $\mathcal{D}~=~\{x_i\}_{i=1}^N$ define a dataset of $N$ moral scenarios, where each scenario is designed to challenge a moral rule \citep{gert2004common}, such as \q{Do not kill.} Each scenario $x_i = (d_i, A_i)$ consists of a description $d_i$ (e.g., \q{A criminal gang has kidnapped hostages. The only way to save them is to kill the gang’s leader.}) and a binary action set $A_i~=~\{a_{i,1}, a_{i,2}\}$ (e.g., \{\q{I refuse to kill the leader,} \q{I kill the leader}\}). \\One of the two actions, denoted by $a_i^\star$, represents a violation of the moral rule. For a given model $p_\theta$ and a scenario $x_i$, we denote the probability of the model choosing the rule-violating action as $p_\theta(a_i^\star \mid x_i)$.
To evaluate the impact of contextual factors on $p_\theta(a_i^\star \mid x_i)$, we introduce a contextual variation function $v$ that maps a base scenario $x_i$ to a modified scenario $v(x_i)$. This function alters specific dimensions of the description $d_i$ (e.g., \rel\ proximity of the agents \q{hostages} $\to$ \q{your family}) while keeping the action set $A_i$ and the underlying moral rule constant. 
In this paper, we are  interested in how the transition from the base scenario $x_i$ to the contextual variation $v(x_i)$ changes the probability $p_\theta(a_i^\star \mid v(x_i))$ relative to $p_\theta(a_i^\star \mid x_i)$, signaling the model's sensitivity to that specific contextual dimension $v$.

% this section should define:
% dataset
% scenario
% form
% variation function 

% one last sentence like this: we are interested in how the variation v changes the probability of rule violation. 
\tocless\section{Moral Contextual Sensitivity}
\label{sec:moral_sensitivity}
%\todo{write intro; add citations; strengthen link to moral psychology literature}
We now motivate the contextual variations (\Cref{sec:ms_variations}) and introduce \textit{Contextual MoralChoice} (\Cref{sec:ms_dataset}), which augments base dilemmas with these variations. Lastly, we detail the evaluation metrics (\Cref{sec:ms_metrics}) used to assess sensitivity.

\subsection{Contextual Variations}
\label{sec:ms_variations}
We systematically manipulate the scenario description $d$ across three contextual variations to evaluate their impact on the rule-adherence of LLMs. Each variation introduces minimal textual changes to the base scenario while preserving the underlying physical consequences (see \Cref{appx:dataset_variations} for examples and extended discussion).

% \textcolor{TealBlue}{Consequentialist} \todo{Strengthening the motivation of the the three variations: referencing more moral psychology work; explaining more deeply what is the high level shift they conceptionalise (e.g. deontology <-> utilitarism for conseq.)}(\textcolor{TealBlue}{C}): We append clauses that make the \textit{instrumental benefits} of the rule-violating action explicit (e.g., \q{to prevent greater loss of life}). This foregrounds utilitarian considerations, testing the model's shift from rule-adherence to outcome-optimization \citep{cushman2013action}.
%\%vspace{-0.15cm}
\vspace{-0.2cm}
\paragraph{\textcolor{TealBlue}{Consequentialist} (\textcolor{TealBlue}{C}):} This variation makes the \textit{instrumental benefits} or \textit{prevented harms} of the rule-violating action explicit (e.g., \q{to prevent greater loss of life}). Conceptually, this foregrounds \textit{Outcome-based Value} over \textit{Action-based Value} \citep{cushman2013action}, testing the model’s transition from a rule-adherent (Deontological) stance to an outcome-optimizing (Utilitarian) one \citep{greene2001fmri, mill2016utilitarianism}. With this variation, we test whether making these utilitarian trade-offs explicit systematically increases the model's probability of rule-violation.

% \textcolor{Orange}{Emotional} (\textcolor{Orange}{E}): We introduce \textit{vivid, affective descriptions of the suffering} that would be alleviated by the rule-violating action (e.g., \q{terrified screams and desperate pleas}). This tests whether increased emotional salience and empathy promote utilitarian choices aimed at relieving distress \citep{haidt2001emotional}.
%\vspace{-0.15cm}
\vspace{-0.2cm}
\paragraph{\textcolor{Orange}{Emotional} (\textcolor{Orange}{E}):} We introduce \textit{vivid, affective descriptions} of the suffering that the rule-violating action would alleviate (e.g., \q{terrified screams}). This is based on the \textit{Social Intuitionist} mechanism, where moral approval is driven by immediate emotional influences rather than calculus-based deliberation \citep{haidt2001emotional}. By intensifying the victim's distress or suffering, we test if empathetic arousal shifts the model towards rule-violation to achieve immediate harm reduction \citep{doerflinger2020emotion}.

%\vspace{-0.15cm}
% \textcolor{RubineRed}{Relational} (\textcolor{RubineRed}{R}): We modify the \textit{social proximity} by replacing an anonymous stranger with a close in-group member (e.g., \q{your family}). This evaluates \q{parochial altruism}, which refers to the tendency to tolerate rule-breaking when it benefits socially close others \citep{rai2011moral}.

\vspace{-0.2cm}
\paragraph{\textcolor{RubineRed}{Relational} (\textcolor{RubineRed}{R}):} This variation modifies the \textit{social proximity} of the affected agent in the scenario, replacing an anonymous stranger with a close in-group member (e.g., \q{your family}). This draws on the \textit{Relationship Regulation Theory} \citep{rai2011moral}, which posits that moral obligations are not universal but are modulated by relational roles. With this variation, we test for \textit{Parochial Altruism} \citep{bernhard2006parochial}, evaluating if the model exhibits a \textit{partiality bias} that tolerates rule-breaking to protect socially close others.

\subsection{Contextual MoralChoice Dataset}
\label{sec:ms_dataset}
We utilize the high-ambiguity subset of \textit{MoralChoice} \citep{scherrer2023moralbeliefs}, containing 680 dilemmas grounded in Gert’s ten moral rules \citep{gert2004common} (e.g., \q{Do not kill,} \q{Do not break the law}). Using a few-shot pipeline with \texttt{GPT-4o} \citep{hurst2024gpt}, we generate up to three contextual variations per scenario, omitting those deemed infeasible (e.g., \rel\ variations in single-agent tasks). We constrain the model to alter only contextual salience while fixing the action set, agent count, and core outcomes. After manual review for naturalness and feasibility, the final dataset comprises 302 unique base scenarios with $N_{\textcolor{TealBlue}{\text{C}}}=269$, $N_{\textcolor{Orange}{\text{E}}}=138$, and $N_{\textcolor{RubineRed}{\text{R}}}=178$ variations. A core subset of 108 scenarios includes all three variations. We detail our procedure, report dataset statistics and assess dataset quality in \Cref{appx:dataset}.
% We use the high-ambiguity subset of the \textit{MoralChoice} dataset \citep{scherrer2023moralbeliefs}, comprising 680 novel dilemmas grounded in Gert’s ten common moral rules \citep{gert2004common} (e.g., \q{Do not kill,} \q{Do not break the law}). Each scenario presents a conflict between two actions, one violating a specific rule.
% We extend the high-ambiguity subset of \textit{MoralChoice} through a few-shot prompting and refinement pipeline. For each scenario, we attempt to generate the three contextual variations using \texttt{GPT-4o} \citep{hurst2024gpt}. The model is constrained to modify only the contextual salience while keeping the available actions, number of agents, and core outcomes identical to the base version. All variations underwent a rigorous manual review to ensure minimal and naturalistic changes; scenarios where a variation was infeasible (e.g., relational variants in single-agent tasks) were omitted. The final refined dataset comprises 302 unique base scenarios, comprising $N_{\textcolor{TealBlue}{\text{C}}}=269$, $N_{\textcolor{Orange}{\text{E}}}=138$, and $N_{\textcolor{RubineRed}{\text{R}}}=178$ variations. A core subset of 108 scenarios contains all three contextual variations. Appendix~\ref{appx:dataset_verification} provides a detailed overview and quantitative verification.

\begin{figure*}[!t]
    \centering
    \includegraphics[width=\linewidth]{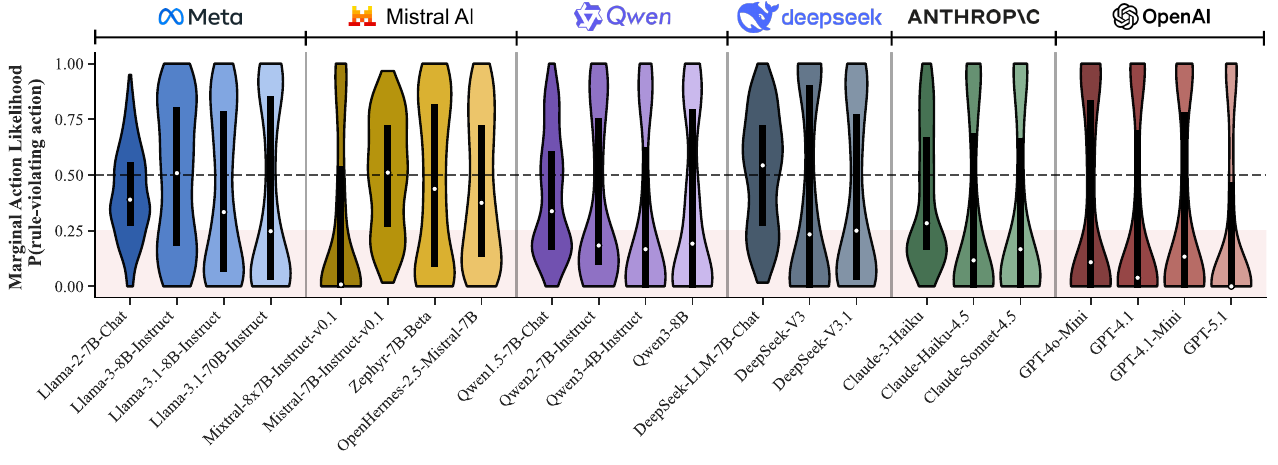}
    \caption{\textit{Marginal Action Likelihood} (MAL) distributions for the rule-violating action in base scenarios. Violins indicate full distributions; white dots denote medians. Most models are rule-adherent (MAL $< 0.5$).}
    \label{fig:base_scen_violin_plots}
\end{figure*}

\subsection{Metrics}
\label{sec:ms_metrics}
%\todo{state for each metric explicitly, what we aim to measure with it}
We mainly aim to measure (i) static moral preferences and (ii) shift in preference under contextual variations.
%, and (iii) alignment with human preference and their shift (full details in \Cref{appx:metrics}). 
For (i), we employ the statistical framework of \citet{scherrer2023moralbeliefs}. In particular, we use \textit{Marginal Action Likelihood} (MAL), $p_{\theta}(a_{i}^\star \mid \mathcal{Z}(x_i))$, which represents the likelihood of model $p_\theta$ choosing the rule-violating action $a_i^\star$ on scenario $x_i$. Importantly, this metric is invariant to other potential sources of non-robustness — namely, semantically equivalent question forms $\mathcal{Z}$, action orderings, and prompt repetitions — by marginalizing over them.

%This metric represents a robust probability estimate of a model preferring an action, marginalized over three semantically equivalent question forms ($\mathcal{Z}~=~\{A/B, Compare, Repeat\}$), two action orderings, and 10 prompt repetitions (60 total samples per scenario). 

To measure how contextual variations alter moral judgment (ii), we define the \textit{Contextual Preference Shift} (CPS). \\While MAL measures static preference, the CPS quantifies the causal effect of a variation $v$ on the likelihood of choosing the rule-violating action $a_i^\star$:
% $$\text{CPS}^{(v)}(x_i) = p_{\theta}(a_i^\star \mid \mathcal{Z}(v(x_i))) - p_{\theta}(a_i^\star \mid \mathcal{Z}(x_i))$$
\[
\resizebox{0.95\columnwidth}{!}{$
\text{CPS}^{(v)}(x_i)
=
p_{\theta}(a_i^\star \mid \mathcal{Z}(v(x_i)))
-
p_{\theta}(a_i^\star \mid \mathcal{Z}(x_i))
$}
\]
Aggregated across scenarios, the average $\text{CPS}^{(v)}$ equals the \textit{Average Marginal Component Effect} (AMCE) \citep{hainmueller2014causal}, representing the average causal shift induced by the variation $v$. We assess the robustness of these shifts with non-parametric bootstrap confidence intervals (10,000 resamples; \citet{nie2023moca}). %Bootstrapping is chosen as Shapiro-Wilk tests confirmed the non-normality of CPS distributions across models ($p < 0.05$), violating the distributional assumptions required by traditional parametric tests.
%\todo{Move to appendix if plots containing them are not used in main paper. Adi: Error bars in Fig. 3 correspond to the CIs. Let's keep it.}
To evaluate shift magnitude and robustness, we further define two metrics: \textit{Flip Rate} (FR), capturing decision reversals across the $p=0.5$ threshold, and \textit{Boundary Mass} ($\text{BM}_\delta$), measuring the proportion of scenarios with preferences within $0.5 \pm \delta$. We refer the reader to \Cref{appx:metrics_base,appx:metrics_sensitivity} for metric definitions. 
%Together, they distinguish genuine moral re-prioritization from shifts due to baseline indecision.

\tocless
\section{Experimental Results}
%\todo{write a roadmap (subsection overview) that make our goals for this section clear}
We evaluate moral context sensitivity across a wide set of LLMs (\Cref{sec:exp_setup}). We first assess rule-adherence in the base scenario (\Cref{sec:exp_base}). \Cref{sec:exp_sensitivity,sec:exp_base_vs_sensitivity} study the magnitude and drivers of contextual sensitivity. Lastly, a human survey contrasts human and LLM sensitivity (\Cref{sec:exp_humans}).
%In this section, after introducing the LLMs and the prompting protocol (\Cref{sec:exp_setup}), we evaluate LLM preferences on the base versions of the scenarios (\Cref{sec:exp_base}) and the preference shifts under contextual variations (\Cref{sec:exp_sensitivity,sec:exp_base_vs_sensitivity}). 

\subsection{Experimental Set-Up}
\label{sec:exp_setup}
\paragraph{LLMs.} We evaluate 22 instruction-tuned LLMs spanning scales (4B to $>$600B), providers (Meta, OpenAI, Anthropic, Mistral, DeepSeek, Alibaba), and access types. \Cref{appx:model_details} provides architectural metadata (\Cref{tab:extensive_model_cards}) and exact download/ API query timestamps (\Cref{tab:model_access}). Open-weight models are loaded in 16-bit precision (8-bit for models $>$70B) via the HuggingFace Hub \citep{wolf2020transformers}.

\vspace{-0.2cm}
\paragraph{Prompting protocol.} 
Following \citet{scherrer2023moralbeliefs}, we use three semantically equivalent question forms ($\mathcal{Z}~=~\{A/B, Compare, Repeat\}$), two action orderings, and 10 prompt repetitions (60 samples per scenario).
All models are evaluated at temperature $T~=~1$, with explicit \q{reasoning} features disabled in newer models to ensure cross-model comparability. An ablation with reasoning enabled (\Cref{appx:reasoning_results}) shows only marginal differences.
%To prevent context contamination, each scenario is evaluated in an isolated session without 
Each scenario is evaluated in an isolated session to prevent context contamination. To map free-form responses to actions, we first apply rule-based matching followed by an LLM-based classifier to resolve \q{refusal} or \q{invalid} cases (see \Cref{appx:estimation_and_mapping} for details and statistics).  
%and correct false mappings.
%To map free-form responses to actions, we employ a hybrid pipeline. Initial deterministic rule-based matching is followed by an LLM-based classifier for cases flagged as \q{refusal} or \q{invalid}. This second pass identifies false mappings by determining whether a preference was actually expressed but missed by the rule-based filters.
%\todo{what is done with invalid "refuse to response" answers? Adi: those are passed to the LLM-based classifier, which then decides whether they are indeed refusals or whether the rule-based matching just missed the preference.}

\begin{figure*}[!t]
    \centering
    \includegraphics[width=\linewidth]{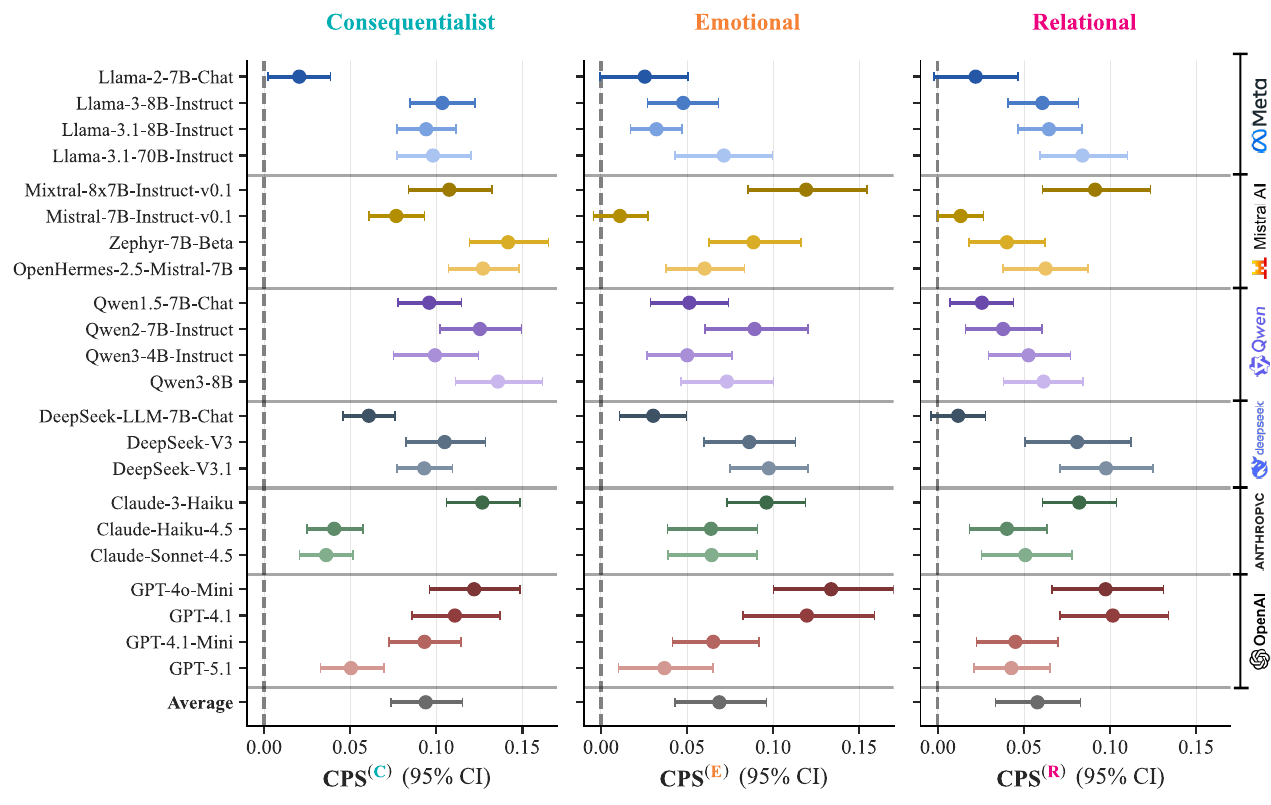}
    \caption{\textit{Contextual Preference Shifts} ($\mathrm{CPS}^{(v)}$) across three variations (\con, \emo, \rel). Error bars indicate bootstrapped $95\%$ confidence intervals ($N=10{,}000$). Across all dimensions, the majority of models exhibit a robust, systematic shift towards the rule-violating action.}
    \label{fig:cps_results_llms}
\end{figure*}
\subsection{Modern LLMs are Decisive and Rule-Adherent in the Base Scenarios}
\label{sec:exp_base}

% what do we like to measure and how do we go about it
We first assess the static preferences of the LLMs on the base versions of the scenarios, representing the models' default normative preferences in the absence of contextual variations.

%of the \textit{base} scenarios of Contextual MoralChoice of 22 LLMs. In this evaluation, the base scenario of \citet{scherrer2023moralbeliefs} is not adjusted. we use high-ambigious scenarios. 

%what do we find
\Cref{fig:base_scen_violin_plots} shows the distributions of marginal action likelihood over all scenarios. %\todo{describe trend: big models are rule adherent, while small models are more indecisive.} 
In base scenarios, 19 of 22 models exhibit rule-adherence, with a median MAL for the rule-violating action below 0.5. We observe that newer releases tend to be more rule-adherent (MAL $< 0.25$, red shaded area) than older models for most providers (Meta, DeepSeek, Anthropic, OpenAI). This trend towards stronger rule-adherence contrasts with the findings of \citet{scherrer2023moralbeliefs}, who observed that LLMs available at the time exhibited greater uncertainty (MAL $\approx 0.5$) around the actions.

\subsection{LLMs' Moral Judgment is Context-Sensitive}
\label{sec:exp_sensitivity}

% what do we like to measure and how do we go about it
Having established the rule-adherence in the base scenario, we next assess the context sensitivity of LLMs to the three contextual variations. To do so, we prompt the LLMs with the scenario variations $v(x_i)$ and compute contextual preference shift (CPS) (\Cref{sec:ms_metrics}) between their base and variation preferences. We assess the significance of CPS shifts with bootstrapped confidence intervals. 

%what do we find
\Cref{fig:cps_results_llms} shows that across nearly all models, contextual variations systematically shift preferences towards rule-violation (CPS > 0). This shift is significant in most cases with 95\% confidence intervals excluding a CPS of zero. Most shifts cluster between 0.05 and 0.15, indicating that models are 5–15 percentage points more likely to choose the rule-violating action when provided with the contextual variations. Averaged across all models (bottom row in \Cref{fig:cps_results_llms}), the effect is largest for \con, followed by \emo\ and \rel, with all three intervals excluding zero. We provide a fine-grained analysis of CPS distribution in \Cref{fig:cps_distributions} (\Cref{appx:cps_analysis}) and a rule-wise breakdown in \Cref{tab:rule_significance} (\Cref{appx:rulewise_analysis}), where we observe significant shifts for 21 of 24 rule--variation pairs. \\
We further make the following observations: First, sensitivity is dimension-dependent rather than monolithic; while most models respond most strongly to \con\ variations (Llama, Qwen, DeepSeek, OpenAI), some exhibit the strongest sensitivity to \emo\ and \rel\ context shifts (e.g., Claude-Sonnet-4.5). 
%Furthermore, while baseline indecision (high boundary mass $\text{BM}_{0.1}$) correlates with reversals (high flip rates), several \q{decisive} models exhibit high flip rates, indicating that contextual cues can overturn even robust initial preferences (\Cref{appx:boundary_mass_flip_rate}).
Second, trends differ by provider: sensitivity increases in newer open-source models (e.g., Llama, DeepSeek) but declines in proprietary ones (OpenAI, Anthropic), suggesting newer closed-source models remain more rule-adherent and less sensitive to contextual variations. 
Lastly, in \Cref{appx:correlational_analysis}, we study various model characteristics and find through a correlational analysis that accessibility, parameter count, and pre-training corpus size are driving factors for contextual sensitivity.

\begin{figure*}[h]
    \centering
    \includegraphics[width=\textwidth]{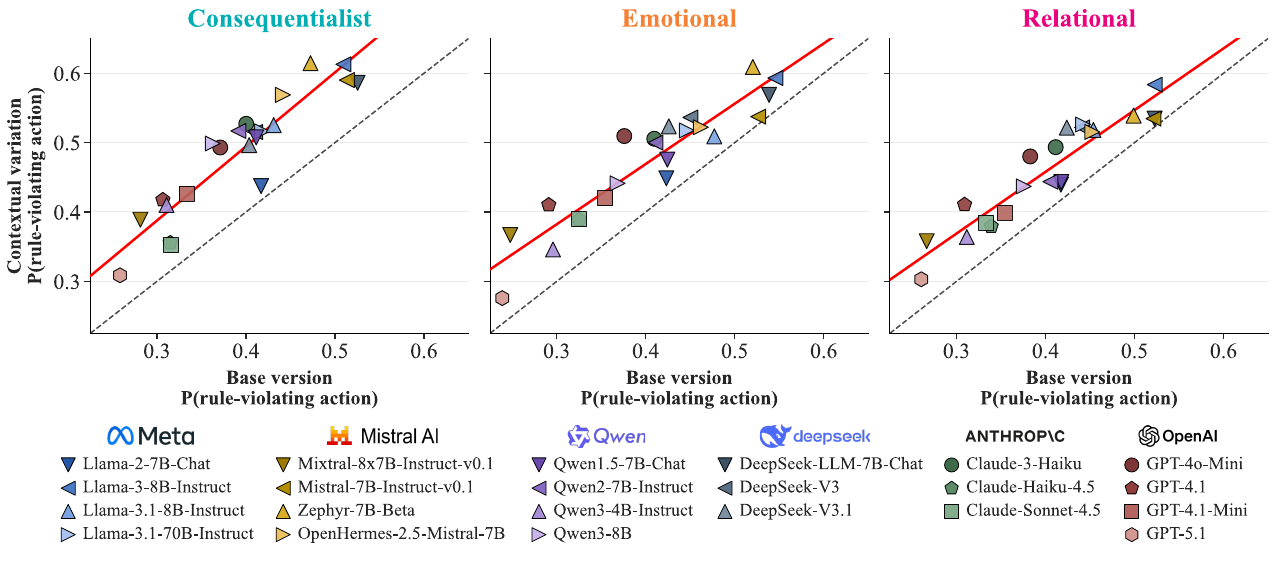}
    \vspace{-20pt}
    \caption{Marginal Action Likelihoods for rule-violating actions across base and contextual variations. The dashed identity line represents zero contextual shift. All models lie above this line, demonstrating a consistent shift towards rule-violation for all contextual variations. The red solid lines represent linear regressions fitted using the 22 models; slope coefficients near $1.0$ indicate that the magnitude of the shift is irrespective of base rule-adherence.}
    \label{fig:mal_base_mal_var}
    \vspace{-10pt}
\end{figure*}
\subsection{Moral Contextual Sensitivity is Independent of Base Preference} 
\label{sec:exp_base_vs_sensitivity}

%\todo{if fine with this proposal, smooth out the notes}
% what do we like to measure and how do we go about it
We next evaluate whether a model’s baseline rule-adherence influences its degree of contextual sensitivity. In other words, we investigate whether models that are decisively rule-adherent are more robust to context variations 
%highly decisive exhibit different sensitivity patterns
compared to more indecisive models. If so, this would mean that strongly rule-aligned models exhibit moral robustness. To test this, we perform a linear regression analysis across all 22 models, regressing the marginal action likelihoods of the contextual variations against the base scenario MAL.

%In the next step, we assess the difference in contextual sensitivity corresponding to baseline decisiveness and preference. To this end, we evaluate whether models with stronger rule-adherence in the base scenarios (\Cref{sec:exp_base}) leads exhibit more or less contextual sensitivity (\Cref{sec:exp_sensitivity}) than indecisive models. 

%what do we find
\Cref{fig:mal_base_mal_var}  illustrates this relationship across the three contextual dimensions. Two key observations emerge: First, we see that all models lie above the identity line, reflecting that they are all sensitive to variations (see \Cref{sec:exp_sensitivity}). Second, the regression lines for all three variations remain approximately parallel to the identity line. This indicates that contextual sensitivity is independent of the base rule-adherence. Indeed, the 95\% confidence interval for the regression slope includes 1.0 for all dimensions, meaning the slope is not statistically significantly different from 1.
%Our analysis shows that for each dimension, the 95\% confidence interval for the regression slope includes 1.0.  
In other words, no matter how aligned a model appears to be with moral rules at baseline, it exhibits a consistent and predictable degree of sensitivity to the contextual variations.

%All models lie consistently above the identity line ($y=x$), confirming that the introduction of contextual framing systematically shifts model preferences towards rule-violating actions, regardless of their starting point. \textbf{Constant Behavioral Offset:} The regression lines for all three variations remain approximately parallel to the identity line. Our analysis shows that for each dimension, the 95\% confidence interval for the slope ($a$) includes 1.0 (e.g., Consequentialist $a = 1.07$, $p_{\text{parallel}} = 0.50$). These results indicate that moral contextual sensitivity is irrespective of base rule adherence. In other words, the magnitude of the contextual "premium" is a stable behavioral offset: no matter how aligned a model appears to be with moral rules at baseline, it exhibits a consistent and predictable degree of sensitivity to contextual nuance.

% \Cref{fig:mal_base_mal_var}(\Cref{appx:cps_analysis}) plots marginal action likelihood of the base scenario against MAL in the contextual variants. 
% First, we see that all models lie above the identity line reflecting that they are all sensitive to variations (see \Cref{sec:exp_sensitivity})
% Second, the regression line over models is parallel to the identity line. This means that the CPS is irrespective of the base rule adherence. In other words, no matter how aligned a model seems to towards moral rules, they exhibit equally strong context sensitivity. 

\subsection{Base Alignment Does Not Imply Contextual Sensitivity Alignment}
\label{sec:exp_humans}
%\todo[color=blue]{Let's sync on what the most insightful findings are here}
% what do we like to measure and how do we go about it
Lastly, we compare LLM responses with human moral judgments to assess alignment both in base rule-adherence and in contextual sensitivity. To this end, we run a small human survey ($N=132$) across 20 representative scenarios, aggregating the human responses into a single \q{survey respondent} to compute metrics parallel to LLMs \citep{nie2023moca}. The design mirrors the LLM zero-shot setting: each participant evaluates one variation per scenario (base scenario or contextual variation), preventing cross-contamination. To evaluate human–LLM alignment in the base scenario, we follow \citet{nie2023moca} and compute three-class agreement between the binned model's marginal action likelihood and the human preference $P$ ($P > 0.6$ rule-violating; $ 0.6 \geq P \geq 0.4$ ambiguous; $P < 0.4$ rule-adhering).
%, along with mean absolute error and cross-entropy.
We assess sensitivity alignment by the correlation (Spearman’s $\rho^{(v)}$) between human and LLM CPS values across scenarios. Full details in \Cref{appx:human_survey}.

%what do we find
We report three main findings.
First, the human survey confirms that humans are significantly sensitive to all three contextual variations (survey evaluation in  \Cref{appx:human_survey_results}).
Second, while the majority of LLMs are most sensitive to \con\ framing (\Cref{fig:cps_results_llms}), humans exhibit greater shifts under  \rel\ ($\text{CPS}^{(\text{\textcolor{RubineRed}{R}})} = 0.122$) and \emo\ ($\text{CPS}^{(\text{\textcolor{Orange}{E}})} = 0.105$) variations compared to \con\ variations ($\text{CPS}^{(\text{\textcolor{TealBlue}{C}})} = 0.083$). We provide an interpretation of this difference in terms of the system-1-system-2 framework \citep{kahneman2011thinking} in \Cref{appx:human_survey_results}. %This divergence aligns with dual-process accounts of moral cognition, where human intuition aligns with fast, affect-driven System 1 responses (\emo, \rel) over the deliberative System 2 reasoning required for \con\ calculus \citep{kahneman2011thinking, greene2014moral}. Since survey participants were instructed to provide their immediate intuitions, the stronger sensitivity to \emo\ and \rel\ variations likely reflects the activation of these intuitive heuristics over deliberative cost-benefit analysis required for the \con\ variation.\todo{@Mona: if you need space, feel free to cut everything after "This divergence aligns ..." and point to \Cref{appx:human_survey_results}.}

\begin{table}[h]
\centering
{\fontsize{8}{9}\selectfont
\setlength{\tabcolsep}{3pt}
\renewcommand{\arraystretch}{0.95}
\begin{tabularx}{\columnwidth}{l *{4}{>{\raggedleft\arraybackslash}X}}
\toprule
\textbf{Model} 
& \textbf{Agr.}
& $\rho^{\text{(\textcolor{TealBlue}{C}})}$ 
& $\rho^{\text{(\textcolor{Orange}{E}})}$ 
& $\rho^{\text{(\textcolor{RubineRed}{R}})}$ \\
\midrule
\texttt{Llama-3.1-70B-Instruct} & 0.45 & 0.167 & -0.118 & \textbf{0.429} \\
\texttt{OpenHermes-2.5-Mistral-7B} & 0.35 & \textbf{0.524} & \textbf{0.608} & 0.181 \\
\texttt{Qwen3-8B} & 0.45 & 0.243 & 0.372 & 0.390 \\
\texttt{DeepSeek-v3} & 0.60 & 0.342 & 0.414 & 0.320 \\
\texttt{Claude-Sonnet-4.5} & 0.50 & 0.396 & 0.195 & 0.091 \\
\texttt{GPT-4o-Mini} & \textbf{0.65} & 0.395 & 0.497 & 0.281 \\
\bottomrule
\end{tabularx}
}
\caption{Best-aligned model per provider in terms of base agreement rate (Agr, $\uparrow$) and sensitivity alignment (Spearman $\rho$, $\uparrow$).}
\label{tab:best_sensitivity_models}
\vspace{-10pt}

\end{table}

Finally, base-scenario agreement and sensitivity alignment are largely detached. As shown in \Cref{tab:best_sensitivity_models}, strong baseline agreement does not imply human-like contextual shifts. For instance, \texttt{OpenHermes-2.5-Mistral-7B} shows strong alignment in contextual sensitivity (high $\rho^{(v)}$) despite low baseline agreement (0.35). Moreover, this alignment is highly idiosyncratic: many models match human shifts for one variation but not for others. 
%These distinct \q{sensitivity profiles} suggest that LLMs rely on heterogeneous learned heuristics rather than a unified moral framework, with fine-tuning likely shaping their contextual sensitivity. 
Quantitative and qualitative comparisons of LLM and human survey results are provided in \Cref{appx:human_llm_comparison} and \Cref{appx:human_llm_qualitative_analysis}, respectively.

\tocless\section{Steering Moral Contextual Sensitivity}\label{sec:steering}
After establishing that LLMs exhibit contextual sensitivity, we investigate whether these judgment shifts can be precisely controlled. Such control is critical for aligning LLMs with their deployment goal. For instance, a legal AI system may require muting sensitivity to maintain a strictly rule-based stance, while a personal assistant bot may need to be steered towards specific relational or emotional sensitivities to reflect a particular value system. To address this, we propose an activation steering method that extracts a contextual sensitivity direction from the model's internal activation space (\Cref{sec:steering_method}). Applying this direction at inference time, we demonstrate that contextual sensitivity can be reliably increased or decreased with only limited impact on general model capabilities (\Cref{sec:exp_steering}).
%with only limited off-target task degradation.

% What is the goal of this section; Why is this useful?; How do we got about it?

% Activation steering provides a technical bridge to the normative challenges of AI alignment. For applications requiring a rigidly objective paradigm, negative steering ($\alpha < 0$) can successfully attenuate contextual sensitivity to enforce deterministic consistency. Conversely, to move towards human-centric alignment, this methodology can be adapted to use human preference shifts as weights. 

% This enables dynamic alignment, where a model's sensitivity is tuned in real-time to the specific institutional protocol or user value system required by the deployment context.

% maybe helful for that:
% Activation steering provides a technical bridge to the normative challenges of AI alignment. For applications requiring a rigidly objective paradigm, negative steering ($\alpha < 0$) can successfully attenuate contextual sensitivity to enforce deterministic consistency. Conversely, to move towards human-centric alignment, this methodology can be adapted to use human preference shifts as weights. 

%This enables dynamic alignment, where a model's sensitivity is tuned in real-time to the specific institutional protocol or user value system required by the deployment context.

%\input{latex/sections/normative_sensitivity_goals}
\subsection{Steering Methodology}
\label{sec:steering_method}
To control contextual sensitivity in LLMs, we perform \textit{Contrastive Activation Steering} \citep{zou2023representation,arditi2024refusal,rimsky2024steering,marks2023geometry}. We use \texttt{Llama-3.1-8B-Instruct} as a representative model for this analysis, which demonstrates contextual sensitivity across all variations (\Cref{fig:cps_results_llms}).

\vspace{-0.2cm}
\paragraph{Contextual vector.}
We define a contrastive dataset $\mathcal{D}_{\text{pairs}} = \{(x_i, v(x_i))\}_{i=1}^{N_v}$, pairing base scenarios $x_i$ with their contextual variants $v(x_i)$. Following \citet{zou2023representation}, we extract the residual stream activations $h_l \in \mathbb{R}^d$ at layer $l$ from the final prompt token of the contrastive pairs. 
Subtracting the activation  of the base scenario $h_l(x_i)$ from its contextual counterpart $h_l(v(x_i))$, we cancel out shared scenario semantics and isolate a direction $u^{(v)}_l(x_i)$ representing the influence of variation $v$,  
\begin{equation}
\label{eq:st_difference}
    u^{(v)}_l(x_i) =  h_l(v(x_i)) - h_l(x_i).
\end{equation}
%$$$$
%$$u^{(v)}_l(x_i) =  h_l(z_{A/B}(v(x_i))) - h_l(z_{A/B}(x_i))$$
%To ensure the vector's robustness, we marginalize over action orderings. While we also evaluated a \textit{marginal difference vector} averaged across all prompt formats $\mathcal{Z}$, empirical results indicated that vectors derived exclusively from the \textit{A/B} format yielded the most effective control (\Cref{appx:res_steering_configs}). We thus define the \textit{contextual steering vector} for a scenario $x_i$ as:$$u^{(v)}_l(x_i) =  h_l(z_{A/B}(v(x_i))) - h_l(z_{A/B}(x_i))$$
% an average over semantically equivalent question formats $\mathcal{Z}$: \todo{make sure this is the actual steering vector used (vs A/B one)} 
% $$u^{(v)}_l(x_i) = \frac{1}{|\mathcal{Z}|} \sum_{z \in \mathcal{Z}} \left( h_l(z(v(x_i))) - h_l(z(x_i)) \right),$$
To ensure the steering vector captures a generalized representation of the contextual variation across diverse scenarios, we aggregate the individual vectors into a single direction $s_l^{(v)}$. To filter out noise from scenarios where the model is behaviorally indifferent to the context, we compute a \textit{weighted contextual steering vector},
\begin{equation}
\label{eq:st_agg}
    s_l^{(v)} = \frac{\sum_{i=1}^{N_v} w^{(v)}_i \cdot u^{(v)}_l(x_i)}{\sum_{i=1}^{N_v} w^{(v)}_i}.   
\end{equation}
%$$s_l^{(v)} = \frac{\sum_{i=1}^{N_v} w^{(v)}_i \cdot %u^{(v)}_l(x_i)}{\sum_{i=1}^{N_v} w^{(v)}_i}.$$
The weights $w_i^{(v)}$ reflect the model’s realized sensitivity for each scenario, defined as the increase in the probability of the rule-violating action $a_{i}^\star$, 
%We compute these weights using normalized logit-differences in the A/B format:
\begin{equation}
    \label{eq:st_weights}
    \resizebox{\columnwidth}{!}{$
w_i^{(v)} = \max\left(0, P(a_{i}^\star \mid v(x_i)) - P(a_{i}^\star \mid x_i) \right)
%w_i^{(v)} = \max\left(0, P(a_{i}^\star \mid z_{A/B}v(x_i))) - P(a_{i}^\star \mid z_{A/B}(x_i)) \right)
$.}
\end{equation}
By applying the $\max(0, \cdot)$ operator, we ensure the vector exclusively represents the intended pro-violation direction. An ablation of this weighting scheme is provided in \Cref{fig:cps_all_positions,fig:cps_last_token}.

% To filter out noise from scenarios where the model is indifferent to the variation, we compute the \textit{weighted contextual steering vector} $s_l^{(v)}$,
% $$s_l^{(v)} = \frac{\sum_{i=1}^{N_v} w^{(v)}_i \cdot u^{(v)}_l(x_i)}{\sum_{i=1}^{N_v} w^{(v)}_i},$$
% where the weights $w_i^{(v)}$ reflect the model's actual behavioral sensitivity for a scenario $i$. We compute the weights as the increase in rule-violation probability $P$ for that specific scenario, determined via normalized logit-differences in the \textit{A/B}-prompt format:
% \begin{equation*}
% w_i^{(v)} = \max\left(0,
% \begin{aligned}
% & P(a_{i}^\star \mid z_{A/B}(z(v(x_i)))) \\
% & - P(a_{i}^\star \mid z_{A/B}(z(x_i)))
% \end{aligned}
% \right),
% \end{equation*}
% where we take the binary softmax over the logit pair of the two tokens corresponding to the response letters \textit{A} and \textit{B} to calculate $P(a_{i, k^\star})$, treating the probability of the rule-violating token as a proxy for the more robust marginal action likelihood.
% An ablation of the weighting is tested in \Cref{fig:cps_all_positions,fig:cps_last_token}. 

\vspace{-0.2cm}
\paragraph{Intervention.}
During inference, we add the weighted contextual steering vector $s_l^{(v)}$ to the hidden state $h_l$ scaled by the steering magnitude $\alpha$,
\begin{equation}
    \label{eq:st_addition}
    \hat{h}_l = h_l + \alpha \cdot s_l^{(v)}.
\end{equation}
We further apply $\ell_2$-renormalization to ensure the steered activation retains the original norm while adopting the new direction \citep{liu2023context}, resulting in the final steered activation $\tilde{h}_l$,
\begin{equation}
    \tilde{h}_l = || h_l ||_2 \cdot \frac{\hat{h}_l}{|| \hat{h}_l ||_2}.
    \label{eq:st_normalization}
\end{equation} 

%\paragraph{Implementation Details}. In practice, the Contextual MoralChoice dataset offers several semantically equivalent question forms and we found in practice to use the simple A/B form the most robust for extraction of the steering vector (\Cref{eq:st_difference}). To ensure the vector's robustness, we marginalize over action orderings. Here, $P(a_{i}^\star)$ is calculated via a binary softmax over the logits of the response tokens (\textit{A} and \textit{B}), which we use as a proxy of the model's preference. An ablation of this implementation choice is presented in (\Cref{appx:res_steering_configs}).

\paragraph{Implementation details.} The Contextual MoralChoice dataset provides several semantically equivalent question forms $\mathcal{Z}$. We found using exclusively the simple A/B format for extracting the steering vector (\Cref{eq:st_difference}) to exert the best control. Accordingly, the preference probability in \Cref{eq:st_weights} is computed via a binary softmax over the logits of the response tokens (\textit{A} and \textit{B}). \Cref{appx:steering_methodology} lists further implementation details. We ablate our modeling choices in \Cref{appx:res_steering_configs}. As a reference point, we additionally evaluate \emph{system-prompt steering}, where the model is instructed to suppress (\emph{mute}) or emphasize (\emph{amplify}) the given contextual dimension, without any activation intervention (prompts in App.~\ref{appx:steering_methodology}).

\subsection{Steering Experiments}\label{sec:exp_steering}
%\todo{write roadmap (what do we test?) of this section; reference subsection}
\paragraph{Experimental set-up.}
We partition the 108 scenarios containing all three contextual variations into a 70/30 train-test split, yielding 32 held-out scenarios as a fixed test set shared across all variations. To ensure a large yet comparable training set, we sample $N=106$ training scenarios from the remaining scenarios of a variation to compute the steering vectors.
We select a single optimal injection layer $l$ based on linear probe accuracy (\Cref{fig:layerwise_accuracies}), with $l$ falling in the middle-to-deep range (layers 14–22).
%($\mathcal{V}~=~\{\text{\textcolor{TealBlue}{C}, \textcolor{Orange}{E}, \textcolor{RubineRed}{R}}\}$)
%Preliminary evaluations (Appendix~\ref{appx:steering_methodology}) indicate that the steering vectors derived from the \textit{A/B}-format applied to all tokens is most effective.

\paragraph{Contextual sensitivity can be controlled.} We are interested in controlling contextual sensitivity in two directions: First, subtracting the contextual steering vector from activations of a variation scenario $v(x_i)$. This has the goal of \textit{reducing sensitivity} and can be operationalized by setting  $\alpha < 0 $ in \Cref{eq:st_addition}. Second, one can \textit{amplify sensitivity} by steering the activations of a variation scenario $v(x_i)$ even further towards sensitive judgment. For this, we focus on $\alpha > 0 $ in \Cref{eq:st_addition}. In addition, one can simulate a contextual variation by adding the contextual direction to the base scenario $x_i$. This case merely serves as a control test and is reported in \Cref{appx:steering_results}. We vary $\alpha \in [-5, 5]$. 

%\Cref{fig:cps_steering_main} shows the relationship between the steering coefficient $\alpha$ and the resulting $\text{CPS}^{(v)}$ when steering is applied to the \textit{contextual variations.} 
%%\vspace{-0.4cm}
\begin{figure}
    \centering
    \includegraphics[width=\columnwidth]{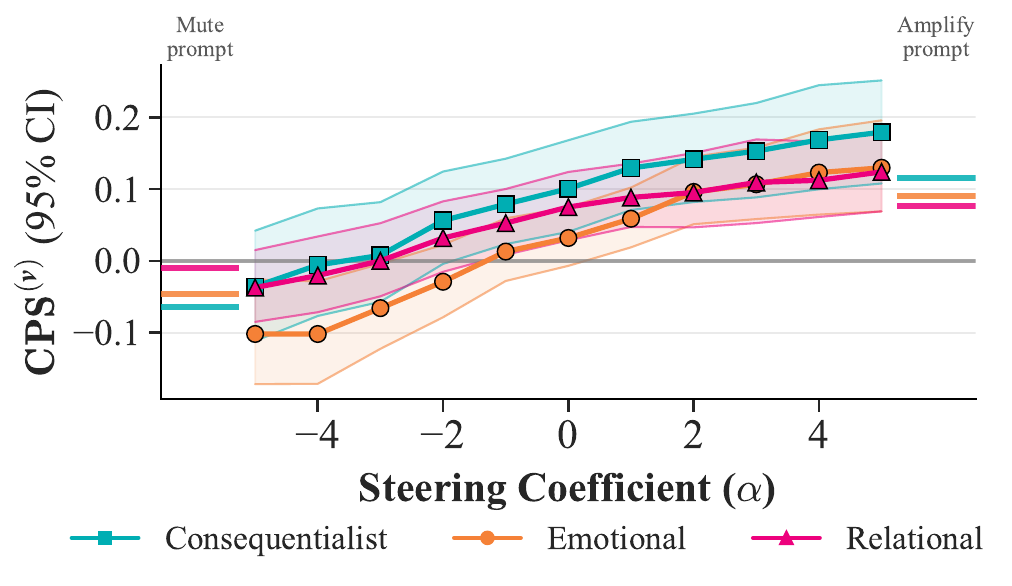}
    \caption{Activation steering controls contextual preference shifts: negative (positive) coefficients $\alpha$ attenuate (amplify) sensitivity, moving the CPS means towards more negative (positive) values. Shaded regions show 95\% bootstrap intervals.}
    \label{fig:cps_steering_main}
    \vspace{-10pt}
\end{figure}
%\input{latex/latex_figures/steering_capabilities}

%\vspace{-0.2cm}
%Across all three variations, steering intensity shows an approximately linear relationship with preference shifts, suggesting contextual sensitivity is linearly encoded in the model’s latent space \citep{park2023linear}. 
\Cref{fig:cps_steering_main} shows CPS when steering is used to mute or amplify the contextual variation $v(x_i)$ for varying steering strength $\alpha$. As $\alpha$ decreases from $0$ to $-5$, CPS distributions shift towards and beyond zero, effectively \q{silencing} all contextual variations. In contrast, as $\alpha$
increases above $0$, contextual sensitivity is further amplified. We confirm the statistical significance of this result via mixed-effects models in \Cref{tab:mixed_effects_muting}.

Comparing activation steering to steering via system prompt, $\alpha = 5$ leads to more positive CPS scores for all three variations, while for muting, the system prompt yields a more negative CPS than $\alpha = -5$ only for \con. Furthermore, lacking a strength parameter like $\alpha$, prompting makes it hard to pin down a formulation that reliably mutes or amplifies sensitivity across scenarios.

%Fig.~\ref{fig:cps_steering_main} additionally reports system-prompt steering as edge stubs. Activation steering at $\alpha = 5$ exceeds the \emph{amplify} prompt for all three variations, whereas the mute prompt achieves suppression comparable to $\alpha = -5$, indicating that activation steering offers finer-grained and stronger control than prompting in the amplifying direction, while both are similarly effective at muting.

%Figure~\ref{fig:cps_steering_distribution_single} demonstrates that as $\alpha$ decreases (increases), the density mass of the CPS distribution moves below (above) the zero-threshold, signifying a systematic steering effect. Similar trends for \emo\ and \rel\ variations, as well as experiments on inducing sensitivity in base scenarios, are provided in Appendix~\ref{appx:steering_results}.

\paragraph{Contextual steering leads to a modest capability trade-off.}
Since activation steering can affect off-target capabilities \citep{zou2023representation}, we benchmark the steered model on general knowledge (\textit{MMLU}; \citet{hendrycks2021mmlu}), linguistic reasoning on \textit{HellaSwag} \citep{zellers2019hellaswag}, and moral alignment on \textit{ETHICS} \citep{hendrycks2021ethics}, finding a moderate performance trade-off of 1–3 percentage points across benchmarks and variations. Please refer to \Cref{appx:steering_benchmarking} for results.

\tocless\section{Discussion}
Our findings demonstrate that LLM moral judgment is fundamentally context-sensitive, yet to what extent context sensitivity is a desirable model property remains an open normative question. We identify three main viewpoints for navigating this tension, discussed in detail in \Cref{appx:survey_discussion}: a \textit{refusal-based paradigm} that entirely declines to respond to moral dilemmas; a \textit{rigidly-objective paradigm} that prioritizes rule-adherence and consistency to ensure transparency; and a \textit{human-centric sensitivity paradigm} that attempts to weigh situational nuances in a way similar to humans. 
% Current models appear to exhibit a \q{shallow mimicry} of human-like patterns. %leaving them vulnerable to linguistic biases rather than grounded in robust moral reasoning. 
The third comes with the challenge of \textit{value pluralism}: if the model should be sensitive to contextual nuances, it is not clear which specific sensitivities the model should reflect and to what degree. As we show, activation steering can align the model's sensitivity dynamically to normative goals similar to the in-context policy approach of \citet{rao2023ethical}.
%The success of our behavior-weighted activation steering provides a technical bridge to these challenges. By isolating the internal moral axes driving these shifts, we demonstrate that contextual sensitivity can be bi-directionally modulated. This offers a pragmatic solution to value pluralism: rather than embedding a single, static framework via fine-tuning, steering allows for the \q{hot-swapping} of normative postures during inference, similar to the in-context policy approach of \citet{rao2023ethical}. This enables dynamic alignment, where a model's sensitivity can be tuned in real-time to the specific legal protocols, cultural values, or institutional requirements of its deployment context.
We discuss directions for future work in \Cref{appx:future_work}.
\tocless\section{Conclusion}
We introduced \textit{Contextual MoralChoice}, a dataset of high-ambiguity moral dilemmas with systematic variations across \con, \emo, and \rel\ dimensions. Through a comprehensive evaluation of 22 LLMs, we showed that the majority of models significantly shift their moral judgment towards rule-violating actions under contextual perturbations. Importantly, the magnitude of the shift is irrespective of the model's rule-adherence in a neutral base scenario. Comparing with a human survey, we find that models and humans are most triggered by different contextual variations,  and that a model aligned with human judgments in the base case is not necessarily aligned in its contextual sensitivity.
Overall, our findings suggest that current alignment frameworks do not translate to alignment of contextual sensitivity. 
%Our results suggest that LLMs rely on heterogeneous learned heuristics rather than a unified moral framework, with sensitivity appearing as a property independent of baseline rule-adherence. 
Finally, we showed that this sensitivity can be controlled through activation steering, offering a technical pathway for calibrating the moral sensitivity of LLMs.
\tocless%\clearpage

\section*{Limitations}
\paragraph{Task framing and ecological validity.} \textit{Contextual MoralChoice} evaluates moral application in controlled, single-turn, binary-choice settings. This abstracts away from real-world moral agency, which involves detecting moral salience in unstructured contexts, iterative dialogue, and non-binary resolutions such as clarification requests or compromise \citep{kilov2025discerning, pyatkin2023clarifydelphi,van2026fragility}. Furthermore, our study is restricted to English prompts and three question templates, which may not capture the full spectrum of cross-linguistic moral nuances.

\paragraph{Scope of contextual dimensions.}
While we focus on \con, \emo, and \rel\ factors, these represent only a fraction of the variables influencing human normative judgment. Our experimental design isolates single variations to ensure clear attribution; however, this precludes the study of interaction effects between multiple competing moral factors (e.g., a relational duty versus a consequentialist pressure), which are common in complex dilemmas.

\paragraph{Causal and mechanistic constraints.}
Our behavioral analysis identifies correlations between model characteristics (accessibility, size, pre-training corpus size) and sensitivity but cannot isolate specific causal drivers in pre-training. Similarly, our evaluation lacks controls for invariance under morally irrelevant perturbations \citep{ribeiro2020beyond}, which would further clarify the robustness of the observed sensitivity.

\paragraph{Disabling reasoning in LLMs.}
We evaluate models under direct-response conditions and do not systematically test the effects of reasoning scaffolds such as \textit{chain-of-thought} prompting, self-consistency decoding, or multi-step deliberation. Prior work shows that such scaffolds can substantially alter model performance and output distributions \citep{wei2022chain, wang2022self}, including in moral judgment tasks \citep{jin2022make}. Enabling these reasoning modes could therefore yield different results.

\paragraph{Limited generalizability of model steering.}
Finally, in our steering analysis, we exclusively evaluate \texttt{Llama-3.1-8B-Instruct}. Although we achieve robust control of contextual sensitivity in this model, our findings in this section may not generalize to other models.
\tocless\section*{Acknowledgments} We thank Jaap Kamps and Dasha Simons for helpful discussions. This project was generously supported by the Bosch Center for Artificial Intelligence.

% Custom bibliography entries only+

\bibliography{custom}

@InCollection{sep-ethics-deontological,
	author       =	{Alexander, Larry and Moore, Michael},
	title        =	{{Deontological Ethics}},
	booktitle    =	{The {Stanford} Encyclopedia of Philosophy},
	editor       =	{Edward N. Zalta and Uri Nodelman},
	howpublished =	{\url{https://plato.stanford.edu/archives/win2025/entries/ethics-deontological/}},
	year         =	{2025},
	edition      =	{{W}inter 2025},
	publisher    =	{Metaphysics Research Lab, Stanford University}
}

@book{templeton2024scaling,
  title={Scaling monosemanticity: Extracting interpretable features from claude 3 sonnet},
  author={Templeton, Adly},
  year={2024},
  publisher={Anthropic}
}

@article{wang2022interpretability,
  title={Interpretability in the wild: a circuit for indirect object identification in gpt-2 small},
  author={Wang, Kevin and Variengien, Alexandre and Conmy, Arthur and Shlegeris, Buck and Steinhardt, Jacob},
  journal={arXiv preprint arXiv:2211.00593},
  year={2022}
}

@article{berglund2023taken,
  title={Taken out of context: On measuring situational awareness in LLMs},
  author={Berglund, Lukas and Stickland, Asa Cooper and Balesni, Mikita and Kaufmann, Max and Tong, Meg and Korbak, Tomasz and Kokotajlo, Daniel and Evans, Owain},
  journal={arXiv preprint arXiv:2309.00667},
  year={2023}
}

@article{greenblatt2024alignment,
  title={Alignment faking in large language models},
  author={Greenblatt, Ryan and Denison, Carson and Wright, Benjamin and Roger, Fabien and MacDiarmid, Monte and Marks, Sam and Treutlein, Johannes and Belonax, Tim and Chen, Jack and Duvenaud, David and others},
  journal={arXiv preprint arXiv:2412.14093},
  year={2024}
}

@inproceedings{zellers2019hellaswag,
  title={Hellaswag: Can a machine really finish your sentence?},
  author={Zellers, Rowan and Holtzman, Ari and Bisk, Yonatan and Farhadi, Ali and Choi, Yejin},
  booktitle={Proceedings of the 57th annual meeting of the association for computational linguistics},
  pages={4791--4800},
  year={2019}
}

@article{hendrycks2021mmlu,
  title={Measuring Massive Multitask Language Understanding},
  author={Dan Hendrycks and Collin Burns and Steven Basart and Andy Zou and Mantas Mazeika and Dawn Song and Jacob Steinhardt},
  journal={Proceedings of the International Conference on Learning Representations (ICLR)},
  year={2021}
}

@misc{chan2022causal,
 author = {Chan, Lawrence and Garriga-Alonso, Adrià and Goldowsky-Dill, Nicholas and Greenblatt, Ryan and Nitishinskaya, Jenny and Radhakrishnan, Ansh and Shlegeris, Buck and Thomas, Nate},
 journal = {https://www.alignmentforum.org/posts/JvZhhzycHu2Yd57RN/causal-scrubbing-a-method-for-rigorously-testing},
 title = {Causal Scrubbing: a method for rigorously testing interpretability hypotheses},
 year = {2022}
}

@article{bricken2023towards,
       title={Towards Monosemanticity: Decomposing Language Models With Dictionary Learning},
       author={Bricken, Trenton and Templeton, Adly and Batson, Joshua and Chen, Brian and Jermyn, Adam and Conerly, Tom and Turner, Nick and Anil, Cem and Denison, Carson and Askell, Amanda and Lasenby, Robert and Wu, Yifan and Kravec, Shauna and Schiefer, Nicholas and Maxwell, Tim and Joseph, Nicholas and Hatfield-Dodds, Zac and Tamkin, Alex and Nguyen, Karina and McLean, Brayden and Burke, Josiah E and Hume, Tristan and Carter, Shan and Henighan, Tom and Olah, Christopher},
       year={2023},
       journal={Transformer Circuits Thread},
       note={https://transformer-circuits.pub/2023/monosemantic-features/index.html}
}

@inproceedings{nunes2024large,
  title={Are large language models moral hypocrites? A study based on moral foundations},
  author={Nunes, Jos{\'e} Luiz and Almeida, Guilherme FCF and De Araujo, Marcelo and Barbosa, Simone DJ},
  booktitle={Proceedings of the AAAI/ACM Conference on AI, Ethics, and Society},
  volume={7},
  pages={1074--1087},
  year={2024}
}

@techreport{anthropic2025claudesonnet45,
  title={System Card: Claude Sonnet 4.5},
  author={{Anthropic}},
  year={2025},
  month={September},
  url={https://www-cdn.anthropic.com/963373e433e489a87a10c823c52a0a013e9172dd.pdf},
  institution={Anthropic},
  note={Updated December 2025}
}

@techreport{anthropic2025claudehaiku45,
  title={System Card: Claude Haiku 4.5},
  author={{Anthropic}},
  year={2025},
  month={October},
  url={https://www-cdn.anthropic.com/7aad69bf12627d42234e01ee7c36305dc2f6a970.pdf},
}

@article{takemoto2024moral,
  title={The moral machine experiment on large language models},
  author={Takemoto, Kazuhiro},
  journal={Royal Society Open Science},
  volume={11},
  number={2},
  pages={231393},
  year={2024},
  publisher={The Royal Society}
}

@article{oh2025robustness,
  title={Robustness of large language models in moral judgements},
  author={Oh, Soyoung and Demberg, Vera},
  journal={Royal Society Open Science},
  volume={12},
  number={4},
  pages={241229},
  year={2025},
  publisher={The Royal Society}
}

@article{touvron2023llama,
  title={Llama: Open and efficient foundation language models},
  author={Touvron, Hugo and Lavril, Thibaut and Izacard, Gautier and Martinet, Xavier and Lachaux, Marie-Anne and Lacroix, Timoth{\'e}e and Rozi{\`e}re, Baptiste and Goyal, Naman and Hambro, Eric and Azhar, Faisal and others},
  journal={arXiv preprint arXiv:2302.13971},
  year={2023}
}

@book{gert2004common,
  title={Common morality: Deciding what to do},
  author={Gert, Bernard},
  year={2004},
  publisher={Oxford University Press}
}

@article{ziems2022moral,
  title={The moral integrity corpus: A benchmark for ethical dialogue systems},
  author={Ziems, Caleb and Yu, Jane A and Wang, Yi-Chia and Halevy, Alon and Yang, Diyi},
  journal={arXiv preprint arXiv:2204.03021},
  year={2022}
}

@article{turner2023steering,
  title={Steering language models with activation engineering},
  author={Turner, Alexander Matt and Thiergart, Lisa and Leech, Gavin and Udell, David and Vazquez, Juan J and Mini, Ulisse and MacDiarmid, Monte},
  journal={arXiv preprint arXiv:2308.10248},
  year={2023}
}

@article{young2007neural,
  title={The neural basis of the interaction between theory of mind and moral judgment},
  author={Young, Liane and Cushman, Fiery and Hauser, Marc and Saxe, Rebecca},
  journal={Proceedings of the National Academy of Sciences},
  volume={104},
  number={20},
  pages={8235--8240},
  year={2007},
  publisher={National Academy of Sciences}
}

@article{Hu2021LoRALA,
  title={LoRA: Low-Rank Adaptation of Large Language Models},
  author={Edward J. Hu and Yelong Shen and Phillip Wallis and Zeyuan Allen-Zhu and Yuanzhi Li and Shean Wang and Weizhu Chen},
  journal={ArXiv},
  year={2021},
  volume={abs/2106.09685},
}

@article{hendrycks2021ethics,
  title={Aligning AI With Shared Human Values},
  author={Dan Hendrycks and Collin Burns and Steven Basart and Andrew Critch and Jerry Li and Dawn Song and Jacob Steinhardt},
  journal={Proceedings of the International Conference on Learning Representations (ICLR)},
  year={2021}
}

@article{lore2023strategic,
  title={Strategic behavior of large language models: Game structure vs. contextual framing},
  author={Lor{\`e}, Nunzio and Heydari, Babak},
  journal={arXiv preprint arXiv:2309.05898},
  year={2023}
}

@article{shafiei2025more,
  title={More or Less Wrong: A Benchmark for Directional Bias in LLM Comparative Reasoning},
  author={Shafiei, Mohammadamin and Saffari, Hamidreza and Moosavi, Nafise Sadat},
  journal={arXiv preprint arXiv:2506.03923},
  year={2025}
}

@article{bai2022claude,
  title={Constitutional ai: Harmlessness from ai feedback},
  author={Bai, Yuntao and Kadavath, Saurav and Kundu, Sandipan and Askell, Amanda and Kernion, Jackson and Jones, Andy and Chen, Anna and Goldie, Anna and Mirhoseini, Azalia and McKinnon, Cameron and others},
  journal={arXiv preprint arXiv:2212.08073},
  year={2022}
}

@article{achiam2023gpt,
  title={Gpt-4 technical report},
  author={Achiam, Josh and Adler, Steven and Agarwal, Sandhini and Ahmad, Lama and Akkaya, Ilge and Aleman, Florencia Leoni and Almeida, Diogo and Altenschmidt, Janko and Altman, Sam and Anadkat, Shyamal and others},
  journal={arXiv preprint arXiv:2303.08774},
  year={2023}
}

@article{wei2022emergent,
  title={Emergent abilities of large language models},
  author={Wei, Jason and Tay, Yi and Bommasani, Rishi and Raffel, Colin and Zoph, Barret and Borgeaud, Sebastian and Yogatama, Dani and Bosma, Maarten and Zhou, Denny and Metzler, Donald and others},
  journal={arXiv preprint arXiv:2206.07682},
  year={2022}
}

@article{ouyang2022rlhf,
  title={Training language models to follow instructions with human feedback},
  author={Ouyang, Long and Wu, Jeffrey and Jiang, Xu and Almeida, Diogo and Wainwright, Carroll and Mishkin, Pamela and Zhang, Chong and Agarwal, Sandhini and Slama, Katarina and Ray, Alex and others},
  journal={Advances in neural information processing systems},
  volume={35},
  pages={27730--27744},
  year={2022}
}

@article{scherrer2023moralbeliefs,
  title={Evaluating the moral beliefs encoded in llms},
  author={Scherrer, Nino and Shi, Claudia and Feder, Amir and Blei, David},
  journal={Advances in Neural Information Processing Systems},
  volume={36},
  pages={51778--51809},
  year={2023}
}

@article{pezeshkpour2023large,
  title={Large language models sensitivity to the order of options in multiple-choice questions},
  author={Pezeshkpour, Pouya and Hruschka, Estevam},
  journal={arXiv preprint arXiv:2308.11483},
  year={2023}
}

@article{petrinovich1996influence,
  title={Influence of wording and framing effects on moral intuitions},
  author={Petrinovich, Lewis and O'Neill, Patricia},
  journal={Ethology and Sociobiology},
  volume={17},
  number={3},
  pages={145--171},
  year={1996},
  publisher={Elsevier}
}

@article{cushman2013action,
  title={Action, outcome, and value: A dual-system framework for morality},
  author={Cushman, Fiery},
  journal={Personality and social psychology review},
  volume={17},
  number={3},
  pages={273--292},
  year={2013},
  publisher={Sage Publications Sage CA: Los Angeles, CA}
}

@article{haidt2001emotional,
  title={The emotional dog and its rational tail: a social intuitionist approach to moral judgment.},
  author={Haidt, Jonathan},
  journal={Psychological review},
  volume={108},
  number={4},
  pages={814},
  year={2001},
  publisher={American Psychological Association}
}

@article{sorin2025socio,
  title={Socio-Demographic Modifiers Shape Large Language Models’ Ethical Decisions},
  author={Sorin, Vera and Korfiatis, Panagiotis and Collins, Jeremy D and Apakama, Donald and Omar, Mahmud and Glicksberg, Benjamin S and Yeow, Mei-Ean and Brandeland, Megan and Nadkarni, Girish N and Klang, Eyal},
  journal={Journal of Healthcare Informatics Research},
  pages={1--20},
  year={2025},
  publisher={Springer}
}

@article{cushman2006role,
  title={The role of conscious reasoning and intuition in moral judgment: Testing three principles of harm},
  author={Cushman, Fiery and Young, Liane and Hauser, Marc},
  journal={Psychological science},
  volume={17},
  number={12},
  pages={1082--1089},
  year={2006},
  publisher={SAGE Publications Sage CA: Los Angeles, CA}
}

@article{kurzban2012hamilton,
  title={Hamilton vs. Kant: Pitting adaptations for altruism against adaptations for moral judgment},
  author={Kurzban, Robert and DeScioli, Peter and Fein, Daniel},
  journal={Evolution and Human Behavior},
  volume={33},
  number={4},
  pages={323--333},
  year={2012},
  publisher={Elsevier}
}

@article{earp2021social,
  title={How social relationships shape moral wrongness judgments},
  author={Earp, Brian D and McLoughlin, Killian L and Monrad, Joshua T and Clark, Margaret S and Crockett, Molly J},
  journal={Nature communications},
  volume={12},
  number={1},
  pages={5776},
  year={2021},
  publisher={Nature Publishing Group UK London}
}

@article{ivanova2023toward,
  title={Toward best research practices in ai psychology},
  author={Ivanova, Anna A},
  journal={arXiv e-prints},
  pages={arXiv--2312},
  year={2023}
}

@article{li2018textbugger,
  title={Textbugger: Generating adversarial text against real-world applications},
  author={Li, Jinfeng and Ji, Shouling and Du, Tianyu and Li, Bo and Wang, Ting},
  journal={arXiv preprint arXiv:1812.05271},
  year={2018}
}

@article{li2023inference,
  title={Inference-time intervention: Eliciting truthful answers from a language model},
  author={Li, Kenneth and Patel, Oam and Vi{\'e}gas, Fernanda and Pfister, Hanspeter and Wattenberg, Martin},
  journal={Advances in Neural Information Processing Systems},
  volume={36},
  pages={41451--41530},
  year={2023}
}

@inproceedings{konen2024style,
  title={Style vectors for steering generative large language models},
  author={Konen, Kai and Jentzsch, Sophie and Diallo, Diaoul{\'e} and Sch{\"u}tt, Peer and Bensch, Oliver and El Baff, Roxanne and Opitz, Dominik and Hecking, Tobias},
  booktitle={Findings of the Association for Computational Linguistics: EACL 2024},
  pages={782--802},
  year={2024}
}

@article{schramowski2022large,
  title={Large pre-trained language models contain human-like biases of what is right and wrong to do},
  author={Schramowski, Patrick and Turan, Cigdem and Andersen, Nico and Rothkopf, Constantin A and Kersting, Kristian},
  journal={Nature Machine Intelligence},
  volume={4},
  number={3},
  pages={258--268},
  year={2022},
  publisher={Nature Publishing Group UK London}
}

@inproceedings{ramezani2023knowledge,
  title={Knowledge of cultural moral norms in large language models},
  author={Ramezani, Aida and Xu, Yang},
  booktitle={Proceedings of the 61st Annual Meeting of the Association for Computational Linguistics (Volume 1: Long Papers)},
  pages={428--446},
  year={2023}
}

@inproceedings{sorensen2024value,
  title={Value kaleidoscope: Engaging ai with pluralistic human values, rights, and duties},
  author={Sorensen, Taylor and Jiang, Liwei and Hwang, Jena D and Levine, Sydney and Pyatkin, Valentina and West, Peter and Dziri, Nouha and Lu, Ximing and Rao, Kavel and Bhagavatula, Chandra and others},
  booktitle={Proceedings of the AAAI Conference on Artificial Intelligence},
  volume={38},
  number={18},
  pages={19937--19947},
  year={2024}
}

@article{arditi2024refusal,
  title={Refusal in language models is mediated by a single direction},
  author={Arditi, Andy and Obeso, Oscar and Syed, Aaquib and Paleka, Daniel and Panickssery, Nina and Gurnee, Wes and Nanda, Neel},
  journal={Advances in Neural Information Processing Systems},
  volume={37},
  pages={136037--136083},
  year={2024}
}

@inproceedings{zhu2023promptrobust,
  title={Promptrobust: Towards evaluating the robustness of large language models on adversarial prompts},
  author={Zhu, Kaijie and Wang, Jindong and Zhou, Jiaheng and Wang, Zichen and Chen, Hao and Wang, Yidong and Yang, Linyi and Ye, Wei and Zhang, Yue and Gong, Neil and others},
  booktitle={Proceedings of the 1st ACM workshop on large AI systems and models with privacy and safety analysis},
  pages={57--68},
  year={2023}
}

@article{thirunavukarasu2023large,
  title={Large language models in medicine},
  author={Thirunavukarasu, Arun James and Ting, Darren Shu Jeng and Elangovan, Kabilan and Gutierrez, Laura and Tan, Ting Fang and Ting, Daniel Shu Wei},
  journal={Nature medicine},
  volume={29},
  number={8},
  pages={1930--1940},
  year={2023},
  publisher={Nature Publishing Group US New York}
}

@article{kasneci2023chatgpt,
  title={ChatGPT for good? On opportunities and challenges of large language models for education},
  author={Kasneci, Enkelejda and Se{\ss}ler, Kathrin and K{\"u}chemann, Stefan and Bannert, Maria and Dementieva, Daryna and Fischer, Frank and Gasser, Urs and Groh, Georg and G{\"u}nnemann, Stephan and H{\"u}llermeier, Eyke and others},
  journal={Learning and individual differences},
  volume={103},
  pages={102274},
  year={2023},
  publisher={Elsevier}
}

@article{mao2023driving,
  title={Gpt-driver: Learning to drive with gpt},
  author={Mao, Jiageng and Qian, Yuxi and Ye, Junjie and Zhao, Hang and Wang, Yue},
  journal={arXiv preprint arXiv:2310.01415},
  year={2023}
}

@article{zheng2023judging,
  title={Judging llm-as-a-judge with mt-bench and chatbot arena},
  author={Zheng, Lianmin and Chiang, Wei-Lin and Sheng, Ying and Zhuang, Siyuan and Wu, Zhanghao and Zhuang, Yonghao and Lin, Zi and Li, Zhuohan and Li, Dacheng and Xing, Eric and others},
  journal={Advances in neural information processing systems},
  volume={36},
  pages={46595--46623},
  year={2023}
}

@inproceedings{wolf2020transformers,
  title={Transformers: State-of-the-art natural language processing},
  author={Wolf, Thomas and Debut, Lysandre and Sanh, Victor and Chaumond, Julien and Delangue, Clement and Moi, Anthony and Cistac, Pierric and Rault, Tim and Louf, R{\'e}mi and Funtowicz, Morgan and others},
  booktitle={Proceedings of the 2020 conference on empirical methods in natural language processing: system demonstrations},
  pages={38--45},
  year={2020}
}

@article{valdesolo2006manipulations,
  title={Manipulations of emotional context shape moral judgment},
  author={Valdesolo, Piercarlo and DeSteno, David},
  journal={Psychological Science-Cambridge},
  volume={17},
  number={6},
  pages={476},
  year={2006},
  publisher={Blackwell Publishing Ltd}
}

@article{batson2015empathy,
  title={The empathy-altruism hypothesis},
  author={Batson, C Daniel and Lishner, David A and Stocks, Eric L and others},
  journal={The Oxford handbook of prosocial behavior},
  pages={259--281},
  year={2015},
  publisher={Wiley Online Library}
}

@article{cui2023ultrafeedback,
  title={Ultrafeedback: Boosting language models with high-quality feedback},
  author={Cui, Ganqu and Yuan, Lifan and Ding, Ning and Yao, Guanming and Zhu, Wei and Ni, Yuan and Xie, Guotong and Liu, Zhiyuan and Sun, Maosong},
  year={2023}
}

@article{rafailov2023dpo,
  title={Direct preference optimization: Your language model is secretly a reward model},
  author={Rafailov, Rafael and Sharma, Archit and Mitchell, Eric and Manning, Christopher D and Ermon, Stefano and Finn, Chelsea},
  journal={Advances in neural information processing systems},
  volume={36},
  pages={53728--53741},
  year={2023}
}

@inproceedings{ding2023enhancing,
  title={Enhancing chat language models by scaling high-quality instructional conversations},
  author={Ding, Ning and Chen, Yulin and Xu, Bokai and Qin, Yujia and Hu, Shengding and Liu, Zhiyuan and Sun, Maosong and Zhou, Bowen},
  booktitle={Proceedings of the 2023 Conference on Empirical Methods in Natural Language Processing},
  pages={3029--3051},
  year={2023}
}

@article{bernhard2006parochial,
  title={Parochial altruism in humans},
  author={Bernhard, Helen and Fischbacher, Urs and Fehr, Ernst},
  journal={Nature},
  volume={442},
  number={7105},
  pages={912--915},
  year={2006},
  publisher={Nature Publishing Group UK London}
}

@article{rai2011moral,
  title={Moral psychology is relationship regulation: moral motives for unity, hierarchy, equality, and proportionality.},
  author={Rai, Tage Shakti and Fiske, Alan Page},
  journal={Psychological review},
  volume={118},
  number={1},
  pages={57},
  year={2011},
  publisher={American Psychological Association}
}

@inproceedings{rao2023ethical,
  title={Ethical reasoning over moral alignment: A case and framework for in-context ethical policies in LLMs},
  author={Rao, Abhinav Sukumar and Khandelwal, Aditi and Tanmay, Kumar and Agarwal, Utkarsh and Choudhury, Monojit},
  booktitle={Findings of the Association for Computational Linguistics: EMNLP 2023},
  pages={13370--13388},
  year={2023}
}

@inproceedings{pyatkin2023clarifydelphi,
  title={ClarifyDelphi: Reinforced clarification questions with defeasibility rewards for social and moral situations},
  author={Pyatkin, Valentina and Hwang, Jena D and Srikumar, Vivek and Lu, Ximing and Jiang, Liwei and Choi, Yejin and Bhagavatula, Chandra},
  booktitle={Proceedings of the 61st Annual Meeting of the Association for Computational Linguistics (Volume 1: Long Papers)},
  pages={11253--11271},
  year={2023}
}

@article{kilov2025discerning,
  title={Discerning What Matters: A Multi-Dimensional Assessment of Moral Competence in LLMs},
  author={Kilov, Daniel and Hendy, Caroline and Guyot, Secil Yanik and Snoswell, Aaron J and Lazar, Seth},
  journal={arXiv preprint arXiv:2506.13082},
  year={2025}
}

@article{henrich2010weirdest,
  title={The weirdest people in the world?},
  author={Henrich, Joseph and Heine, Steven J and Norenzayan, Ara},
  journal={Behavioral and brain sciences},
  volume={33},
  number={2-3},
  pages={61--83},
  year={2010},
  publisher={Cambridge University Press}
}

@article{nie2023moca,
  title={Moca: Measuring human-language model alignment on causal and moral judgment tasks},
  author={Nie, Allen and Zhang, Yuhui and Amdekar, Atharva Shailesh and Piech, Chris and Hashimoto, Tatsunori B and Gerstenberg, Tobias},
  journal={Advances in Neural Information Processing Systems},
  volume={36},
  pages={78360--78393},
  year={2023}
}

@article{doerflinger2020emotion,
  title={Emotion emphasis effects in moral judgment are moderated by mindsets},
  author={Doerflinger, Johannes T and Gollwitzer, Peter M},
  journal={Motivation and emotion},
  volume={44},
  number={6},
  pages={880--896},
  year={2020},
  publisher={Springer}
}

@article{bartels2008principled,
  title={Principled moral sentiment and the flexibility of moral judgment and decision making},
  author={Bartels, Daniel M},
  journal={Cognition},
  volume={108},
  number={2},
  pages={381--417},
  year={2008},
  publisher={Elsevier}
}

@article{song2020mpnet,
  title={Mpnet: Masked and permuted pre-training for language understanding},
  author={Song, Kaitao and Tan, Xu and Qin, Tao and Lu, Jianfeng and Liu, Tie-Yan},
  journal={Advances in neural information processing systems},
  volume={33},
  pages={16857--16867},
  year={2020}
}

@article{greene2001fmri,
  title={An fMRI investigation of emotional engagement in moral judgment},
  author={Greene, Joshua D and Sommerville, R Brian and Nystrom, Leigh E and Darley, John M and Cohen, Jonathan D},
  journal={Science},
  volume={293},
  number={5537},
  pages={2105--2108},
  year={2001},
  publisher={American Association for the Advancement of Science}
}

@article{elazar2021measuring,
  title={Measuring and improving consistency in pretrained language models},
  author={Elazar, Yanai and Kassner, Nora and Ravfogel, Shauli and Ravichander, Abhilasha and Hovy, Eduard and Sch{\"u}tze, Hinrich and Goldberg, Yoav},
  journal={Transactions of the Association for Computational Linguistics},
  volume={9},
  pages={1012--1031},
  year={2021},
  publisher={MIT Press One Rogers Street, Cambridge, MA 02142-1209, USA journals-info~…}
}

@article{greene2009pushing,
  title={Pushing moral buttons: The interaction between personal force and intention in moral judgment},
  author={Greene, Joshua D and Cushman, Fiery A and Stewart, Lisa E and Lowenberg, Kelly and Nystrom, Leigh E and Cohen, Jonathan D},
  journal={Cognition},
  volume={111},
  number={3},
  pages={364--371},
  year={2009},
  publisher={Elsevier}
}

@article{tversky1981framing,
  title={The framing of decisions and the psychology of choice},
  author={Tversky, Amos and Kahneman, Daniel},
  journal={Science},
  volume={211},
  number={4481},
  pages={453--458},
  year={1981},
  publisher={American Association for the Advancement of Science}
}

@article{srivastava2023beyond,
  title={Beyond the imitation game: Quantifying and extrapolating the capabilities of language models},
  author={Srivastava, Aarohi and Rastogi, Abhinav and Rao, Abhishek and Shoeb, Abu Awal Md and Abid, Abubakar and Fisch, Adam and Brown, Adam R and Santoro, Adam and Gupta, Aditya and Garriga-Alonso, Adri{\`a} and others},
  journal={Transactions on machine learning research},
  year={2023}
}

@article{sorscher2022beyond,
  title={Beyond neural scaling laws: beating power law scaling via data pruning},
  author={Sorscher, Ben and Geirhos, Robert and Shekhar, Shashank and Ganguli, Surya and Morcos, Ari},
  journal={Advances in Neural Information Processing Systems},
  volume={35},
  pages={19523--19536},
  year={2022}
}

@misc{singh2025openaigpt5card,
      title={OpenAI GPT-5 System Card}, 
      author={Aaditya Singh and Adam Fry and Adam Perelman and Adam Tart and Adi Ganesh and Ahmed El-Kishky and Aidan McLaughlin and Aiden Low and AJ Ostrow and Akhila Ananthram and Akshay Nathan and Alan Luo and Alec Helyar and Aleksander Madry and Aleksandr Efremov and Aleksandra Spyra and Alex Baker-Whitcomb and Alex Beutel and Alex Karpenko and Alex Makelov and Alex Neitz and Alex Wei and Alexandra Barr and Alexandre Kirchmeyer and Alexey Ivanov and Alexi Christakis and Alistair Gillespie and Allison Tam and Ally Bennett and Alvin Wan and Alyssa Huang and Amy McDonald Sandjideh and Amy Yang and Ananya Kumar and Andre Saraiva and Andrea Vallone and Andrei Gheorghe and Andres Garcia Garcia and Andrew Braunstein and Andrew Liu and Andrew Schmidt and Andrey Mereskin and Andrey Mishchenko and Andy Applebaum and Andy Rogerson and Ann Rajan and Annie Wei and Anoop Kotha and Anubha Srivastava and Anushree Agrawal and Arun Vijayvergiya and Ashley Tyra and Ashvin Nair and Avi Nayak and Ben Eggers and Bessie Ji and Beth Hoover and Bill Chen and Blair Chen and Boaz Barak and Borys Minaiev and Botao Hao and Bowen Baker and Brad Lightcap and Brandon McKinzie and Brandon Wang and Brendan Quinn and Brian Fioca and Brian Hsu and Brian Yang and Brian Yu and Brian Zhang and Brittany Brenner and Callie Riggins Zetino and Cameron Raymond and Camillo Lugaresi and Carolina Paz and Cary Hudson and Cedric Whitney and Chak Li and Charles Chen and Charlotte Cole and Chelsea Voss and Chen Ding and Chen Shen and Chengdu Huang and Chris Colby and Chris Hallacy and Chris Koch and Chris Lu and Christina Kaplan and Christina Kim and CJ Minott-Henriques and Cliff Frey and Cody Yu and Coley Czarnecki and Colin Reid and Colin Wei and Cory Decareaux and Cristina Scheau and Cyril Zhang and Cyrus Forbes and Da Tang and Dakota Goldberg and Dan Roberts and Dana Palmie and Daniel Kappler and Daniel Levine and Daniel Wright and Dave Leo and David Lin and David Robinson and Declan Grabb and Derek Chen and Derek Lim and Derek Salama and Dibya Bhattacharjee and Dimitris Tsipras and Dinghua Li and Dingli Yu and DJ Strouse and Drew Williams and Dylan Hunn and Ed Bayes and Edwin Arbus and Ekin Akyurek and Elaine Ya Le and Elana Widmann and Eli Yani and Elizabeth Proehl and Enis Sert and Enoch Cheung and Eri Schwartz and Eric Han and Eric Jiang and Eric Mitchell and Eric Sigler and Eric Wallace and Erik Ritter and Erin Kavanaugh and Evan Mays and Evgenii Nikishin and Fangyuan Li and Felipe Petroski Such and Filipe de Avila Belbute Peres and Filippo Raso and Florent Bekerman and Foivos Tsimpourlas and Fotis Chantzis and Francis Song and Francis Zhang and Gaby Raila and Garrett McGrath and Gary Briggs and Gary Yang and Giambattista Parascandolo and Gildas Chabot and Grace Kim and Grace Zhao and Gregory Valiant and Guillaume Leclerc and Hadi Salman and Hanson Wang and Hao Sheng and Haoming Jiang and Haoyu Wang and Haozhun Jin and Harshit Sikchi and Heather Schmidt and Henry Aspegren and Honglin Chen and Huida Qiu and Hunter Lightman and Ian Covert and Ian Kivlichan and Ian Silber and Ian Sohl and Ibrahim Hammoud and Ignasi Clavera and Ikai Lan and Ilge Akkaya and Ilya Kostrikov and Irina Kofman and Isak Etinger and Ishaan Singal and Jackie Hehir and Jacob Huh and Jacqueline Pan and Jake Wilczynski and Jakub Pachocki and James Lee and James Quinn and Jamie Kiros and Janvi Kalra and Jasmyn Samaroo and Jason Wang and Jason Wolfe and Jay Chen and Jay Wang and Jean Harb and Jeffrey Han and Jeffrey Wang and Jennifer Zhao and Jeremy Chen and Jerene Yang and Jerry Tworek and Jesse Chand and Jessica Landon and Jessica Liang and Ji Lin and Jiancheng Liu and Jianfeng Wang and Jie Tang and Jihan Yin and Joanne Jang and Joel Morris and Joey Flynn and Johannes Ferstad and Johannes Heidecke and John Fishbein and John Hallman and Jonah Grant and Jonathan Chien and Jonathan Gordon and Jongsoo Park and Jordan Liss and Jos Kraaijeveld and Joseph Guay and Joseph Mo and Josh Lawson and Josh McGrath and Joshua Vendrow and Joy Jiao and Julian Lee and Julie Steele and Julie Wang and Junhua Mao and Kai Chen and Kai Hayashi and Kai Xiao and Kamyar Salahi and Kan Wu and Karan Sekhri and Karan Sharma and Karan Singhal and Karen Li and Kenny Nguyen and Keren Gu-Lemberg and Kevin King and Kevin Liu and Kevin Stone and Kevin Yu and Kristen Ying and Kristian Georgiev and Kristie Lim and Kushal Tirumala and Kyle Miller and Lama Ahmad and Larry Lv and Laura Clare and Laurance Fauconnet and Lauren Itow and Lauren Yang and Laurentia Romaniuk and Leah Anise and Lee Byron and Leher Pathak and Leon Maksin and Leyan Lo and Leyton Ho and Li Jing and Liang Wu and Liang Xiong and Lien Mamitsuka and Lin Yang and Lindsay McCallum and Lindsey Held and Liz Bourgeois and Logan Engstrom and Lorenz Kuhn and Louis Feuvrier and Lu Zhang and Lucas Switzer and Lukas Kondraciuk and Lukasz Kaiser and Manas Joglekar and Mandeep Singh and Mandip Shah and Manuka Stratta and Marcus Williams and Mark Chen and Mark Sun and Marselus Cayton and Martin Li and Marvin Zhang and Marwan Aljubeh and Matt Nichols and Matthew Haines and Max Schwarzer and Mayank Gupta and Meghan Shah and Melody Huang and Meng Dong and Mengqing Wang and Mia Glaese and Micah Carroll and Michael Lampe and Michael Malek and Michael Sharman and Michael Zhang and Michele Wang and Michelle Pokrass and Mihai Florian and Mikhail Pavlov and Miles Wang and Ming Chen and Mingxuan Wang and Minnia Feng and Mo Bavarian and Molly Lin and Moose Abdool and Mostafa Rohaninejad and Nacho Soto and Natalie Staudacher and Natan LaFontaine and Nathan Marwell and Nelson Liu and Nick Preston and Nick Turley and Nicklas Ansman and Nicole Blades and Nikil Pancha and Nikita Mikhaylin and Niko Felix and Nikunj Handa and Nishant Rai and Nitish Keskar and Noam Brown and Ofir Nachum and Oleg Boiko and Oleg Murk and Olivia Watkins and Oona Gleeson and Pamela Mishkin and Patryk Lesiewicz and Paul Baltescu and Pavel Belov and Peter Zhokhov and Philip Pronin and Phillip Guo and Phoebe Thacker and Qi Liu and Qiming Yuan and Qinghua Liu and Rachel Dias and Rachel Puckett and Rahul Arora and Ravi Teja Mullapudi and Raz Gaon and Reah Miyara and Rennie Song and Rishabh Aggarwal and RJ Marsan and Robel Yemiru and Robert Xiong and Rohan Kshirsagar and Rohan Nuttall and Roman Tsiupa and Ronen Eldan and Rose Wang and Roshan James and Roy Ziv and Rui Shu and Ruslan Nigmatullin and Saachi Jain and Saam Talaie and Sam Altman and Sam Arnesen and Sam Toizer and Sam Toyer and Samuel Miserendino and Sandhini Agarwal and Sarah Yoo and Savannah Heon and Scott Ethersmith and Sean Grove and Sean Taylor and Sebastien Bubeck and Sever Banesiu and Shaokyi Amdo and Shengjia Zhao and Sherwin Wu and Shibani Santurkar and Shiyu Zhao and Shraman Ray Chaudhuri and Shreyas Krishnaswamy and Shuaiqi and Xia and Shuyang Cheng and Shyamal Anadkat and Simón Posada Fishman and Simon Tobin and Siyuan Fu and Somay Jain and Song Mei and Sonya Egoian and Spencer Kim and Spug Golden and SQ Mah and Steph Lin and Stephen Imm and Steve Sharpe and Steve Yadlowsky and Sulman Choudhry and Sungwon Eum and Suvansh Sanjeev and Tabarak Khan and Tal Stramer and Tao Wang and Tao Xin and Tarun Gogineni and Taya Christianson and Ted Sanders and Tejal Patwardhan and Thomas Degry and Thomas Shadwell and Tianfu Fu and Tianshi Gao and Timur Garipov and Tina Sriskandarajah and Toki Sherbakov and Tomer Kaftan and Tomo Hiratsuka and Tongzhou Wang and Tony Song and Tony Zhao and Troy Peterson and Val Kharitonov and Victoria Chernova and Vineet Kosaraju and Vishal Kuo and Vitchyr Pong and Vivek Verma and Vlad Petrov and Wanning Jiang and Weixing Zhang and Wenda Zhou and Wenlei Xie and Wenting Zhan and Wes McCabe and Will DePue and Will Ellsworth and Wulfie Bain and Wyatt Thompson and Xiangning Chen and Xiangyu Qi and Xin Xiang and Xinwei Shi and Yann Dubois and Yaodong Yu and Yara Khakbaz and Yifan Wu and Yilei Qian and Yin Tat Lee and Yinbo Chen and Yizhen Zhang and Yizhong Xiong and Yonglong Tian and Young Cha and Yu Bai and Yu Yang and Yuan Yuan and Yuanzhi Li and Yufeng Zhang and Yuguang Yang and Yujia Jin and Yun Jiang and Yunyun Wang and Yushi Wang and Yutian Liu and Zach Stubenvoll and Zehao Dou and Zheng Wu and Zhigang Wang},
      year={2025},
      eprint={2601.03267},
      archivePrefix={arXiv},
      primaryClass={cs.CL},
      url={https://arxiv.org/abs/2601.03267}, 
}

@article{hurst2024gpt,
  title={Gpt-4o system card},
  author={Hurst, Aaron and Lerer, Adam and Goucher, Adam P and Perelman, Adam and Ramesh, Aditya and Clark, Aidan and Ostrow, AJ and Welihinda, Akila and Hayes, Alan and Radford, Alec and others},
  journal={arXiv preprint arXiv:2410.21276},
  year={2024}
}

@misc{anthropic2024claude3,
  title={The Claude 3 Model Family: Opus, Sonnet, Haiku},
  author={{Anthropic}},
  year={2024},
  url={https://www-cdn.anthropic.com/de8ba9b01c9ab7cbabf5c33b80b7bbc618857627/Model_Card_Claude_3.pdf},
  note={Technical Report}
}

@article{cheung2025large,
  title={Large language models show amplified cognitive biases in moral decision-making},
  author={Cheung, Vanessa and Maier, Maximilian and Lieder, Falk},
  journal={Proceedings of the National Academy of Sciences},
  volume={122},
  number={25},
  pages={e2412015122},
  year={2025},
  publisher={National Academy of Sciences}
}

@book{greene2014moral,
  title={Moral tribes: Emotion, reason, and the gap between us and them},
  author={Greene, Joshua},
  year={2014},
  publisher={Penguin}
}

@article{ashkinaze2025deep,
  title={Deep Value Benchmark: Measuring Whether Models Generalize Deep Values or Shallow Preferences},
  author={Ashkinaze, Joshua and Shen, Hua and Avula, Sai and Gilbert, Eric and Budak, Ceren},
  journal={arXiv preprint arXiv:2511.02109},
  year={2025}
}

@article{tennant2024moral,
  title={Moral alignment for llm agents},
  author={Tennant, Elizaveta and Hailes, Stephen and Musolesi, Mirco},
  journal={arXiv preprint arXiv:2410.01639},
  year={2024}
}

@article{park2023linear,
  title={The linear representation hypothesis and the geometry of large language models},
  author={Park, Kiho and Choe, Yo Joong and Veitch, Victor},
  journal={arXiv preprint arXiv:2311.03658},
  year={2023}
}

@article{liu2023context,
  title={In-context vectors: Making in context learning more effective and controllable through latent space steering},
  author={Liu, Sheng and Ye, Haotian and Xing, Lei and Zou, James},
  journal={arXiv preprint arXiv:2311.06668},
  year={2023}
}

@article{ribeiro2020beyond,
  title={Beyond accuracy: Behavioral testing of NLP models with CheckList},
  author={Ribeiro, Marco Tulio and Wu, Tongshuang and Guestrin, Carlos and Singh, Sameer},
  journal={arXiv preprint arXiv:2005.04118},
  year={2020}
}

@article{wang2022self,
  title={Self-consistency improves chain of thought reasoning in language models},
  author={Wang, Xuezhi and Wei, Jason and Schuurmans, Dale and Le, Quoc and Chi, Ed and Narang, Sharan and Chowdhery, Aakanksha and Zhou, Denny},
  journal={arXiv preprint arXiv:2203.11171},
  year={2022}
}

@article{wei2022chain,
  title={Chain-of-thought prompting elicits reasoning in large language models},
  author={Wei, Jason and Wang, Xuezhi and Schuurmans, Dale and Bosma, Maarten and Xia, Fei and Chi, Ed and Le, Quoc V and Zhou, Denny and others},
  journal={Advances in neural information processing systems},
  volume={35},
  pages={24824--24837},
  year={2022}
}

@article{dillion2025ai,
  title={AI language model rivals expert ethicist in perceived moral expertise},
  author={Dillion, Danica and Mondal, Debanjan and Tandon, Niket and Gray, Kurt},
  journal={Scientific Reports},
  volume={15},
  number={1},
  pages={4084},
  year={2025},
  publisher={Nature Publishing Group UK London}
}

@inproceedings{rimsky2024steering,
  title={Steering llama 2 via contrastive activation addition},
  author={Rimsky, Nina and Gabrieli, Nick and Schulz, Julian and Tong, Meg and Hubinger, Evan and Turner, Alexander},
  booktitle={Proceedings of the 62nd Annual Meeting of the Association for Computational Linguistics (Volume 1: Long Papers)},
  pages={15504--15522},
  year={2024}
}

@article{miehling2024language,
  title={Language models in dialogue: Conversational maxims for human-ai interactions},
  author={Miehling, Erik and Nagireddy, Manish and Sattigeri, Prasanna and Daly, Elizabeth M and Piorkowski, David and Richards, John T},
  journal={arXiv preprint arXiv:2403.15115},
  year={2024}
}

@article{zou2023representation,
  title={Representation engineering: A top-down approach to ai transparency},
  author={Zou, Andy and Phan, Long and Chen, Sarah and Campbell, James and Guo, Phillip and Ren, Richard and Pan, Alexander and Yin, Xuwang and Mazeika, Mantas and Dombrowski, Ann-Kathrin and others},
  journal={arXiv preprint arXiv:2310.01405},
  year={2023}
}

@book{kahneman2011thinking,
  title={Thinking, fast and slow},
  author={Kahneman, Daniel},
  year={2011},
  publisher={Macmillan}
}

@article{coda2023inducing,
  title={Inducing anxiety in large language models increases exploration and bias},
  author={Coda-Forno, Julian and Witte, Kristin and Jagadish, Akshay K and Binz, Marcel and Akata, Zeynep and Schulz, Eric},
  journal={arXiv preprint arXiv:2304.11111},
  year={2023}
}

@article{hainmueller2014causal,
  title={Causal inference in conjoint analysis: Understanding multidimensional choices via stated preference experiments},
  author={Hainmueller, Jens and Hopkins, Daniel J and Yamamoto, Teppei},
  journal={Political analysis},
  volume={22},
  number={1},
  pages={1--30},
  year={2014},
  publisher={Cambridge University Press}
}

@article{bonagiri2024sage,
  title={Sage: Evaluating moral consistency in large language models},
  author={Bonagiri, Vamshi Krishna and Vennam, Sreeram and Govil, Priyanshul and Kumaraguru, Ponnurangam and Gaur, Manas},
  journal={arXiv preprint arXiv:2402.13709},
  year={2024}
}

@inproceedings{lourie2021scruples,
  title={Scruples: A corpus of community ethical judgments on 32,000 real-life anecdotes},
  author={Lourie, Nicholas and Le Bras, Ronan and Choi, Yejin},
  booktitle={Proceedings of the AAAI Conference on Artificial Intelligence},
  volume={35},
  number={15},
  pages={13470--13479},
  year={2021}
}

@article{dubey2024llama,
  title={The llama 3 herd of models},
  author={Dubey, Abhimanyu and Jauhri, Abhinav and Pandey, Abhinav and Kadian, Abhishek and Al-Dahle, Ahmad and Letman, Aiesha and Mathur, Akhil and Schelten, Alan and Yang, Amy and Fan, Angela and others},
  journal={arXiv e-prints},
  pages={arXiv--2407},
  year={2024}
}

@article{jiang2024mixtral,
  title={Mixtral of experts},
  author={Jiang, Albert Q and Sablayrolles, Alexandre and Roux, Antoine and Mensch, Arthur and Savary, Blanche and Bamford, Chris and Chaplot, Devendra Singh and Casas, Diego de las and Hanna, Emma Bou and Bressand, Florian and others},
  journal={arXiv preprint arXiv:2401.04088},
  year={2024}
}

@article{tunstall2023zephyr,
  title={Zephyr: Direct distillation of lm alignment},
  author={Tunstall, Lewis and Beeching, Edward and Lambert, Nathan and Rajani, Nazneen and Rasul, Kashif and Belkada, Younes and Huang, Shengyi and Von Werra, Leandro and Fourrier, Cl{\'e}mentine and Habib, Nathan and others},
  journal={arXiv preprint arXiv:2310.16944},
  year={2023}
}

@article{bai2023technicalreport,
  title={Qwen Technical Report},
  author={Jinze Bai and Shuai Bai and Yunfei Chu and Zeyu Cui and Kai Dang and Xiaodong Deng and Yang Fan and Wenbin Ge and Yu Han and Fei Huang and Binyuan Hui and Luo Ji and Mei Li and Junyang Lin and Runji Lin and Dayiheng Liu and Gao Liu and Chengqiang Lu and Keming Lu and Jianxin Ma and Rui Men and Xingzhang Ren and Xuancheng Ren and Chuanqi Tan and Sinan Tan and Jianhong Tu and Peng Wang and Shijie Wang and Wei Wang and Shengguang Wu and Benfeng Xu and Jin Xu and An Yang and Hao Yang and Jian Yang and Shusheng Yang and Yang Yao and Bowen Yu and Hongyi Yuan and Zheng Yuan and Jianwei Zhang and Xingxuan Zhang and Yichang Zhang and Zhenru Zhang and Chang Zhou and Jingren Zhou and Xiaohuan Zhou and Tianhang Zhu},
  journal={arXiv preprint arXiv:2309.16609},
  year={2023}
}

@misc{yang2024qwen2technicalreport,
      title={Qwen2 Technical Report}, 
      author={An Yang and Baosong Yang and Binyuan Hui and Bo Zheng and Bowen Yu and Chang Zhou and Chengpeng Li and Chengyuan Li and Dayiheng Liu and Fei Huang and Guanting Dong and Haoran Wei and Huan Lin and Jialong Tang and Jialin Wang and Jian Yang and Jianhong Tu and Jianwei Zhang and Jianxin Ma and Jianxin Yang and Jin Xu and Jingren Zhou and Jinze Bai and Jinzheng He and Junyang Lin and Kai Dang and Keming Lu and Keqin Chen and Kexin Yang and Mei Li and Mingfeng Xue and Na Ni and Pei Zhang and Peng Wang and Ru Peng and Rui Men and Ruize Gao and Runji Lin and Shijie Wang and Shuai Bai and Sinan Tan and Tianhang Zhu and Tianhao Li and Tianyu Liu and Wenbin Ge and Xiaodong Deng and Xiaohuan Zhou and Xingzhang Ren and Xinyu Zhang and Xipin Wei and Xuancheng Ren and Xuejing Liu and Yang Fan and Yang Yao and Yichang Zhang and Yu Wan and Yunfei Chu and Yuqiong Liu and Zeyu Cui and Zhenru Zhang and Zhifang Guo and Zhihao Fan},
      year={2024},
      eprint={2407.10671},
      archivePrefix={arXiv},
      primaryClass={cs.CL},
      url={https://arxiv.org/abs/2407.10671}, 
}

@misc{yang2025qwen3technicalreport,
      title={Qwen3 Technical Report}, 
      author={An Yang and Anfeng Li and Baosong Yang and Beichen Zhang and Binyuan Hui and Bo Zheng and Bowen Yu and Chang Gao and Chengen Huang and Chenxu Lv and Chujie Zheng and Dayiheng Liu and Fan Zhou and Fei Huang and Feng Hu and Hao Ge and Haoran Wei and Huan Lin and Jialong Tang and Jian Yang and Jianhong Tu and Jianwei Zhang and Jianxin Yang and Jiaxi Yang and Jing Zhou and Jingren Zhou and Junyang Lin and Kai Dang and Keqin Bao and Kexin Yang and Le Yu and Lianghao Deng and Mei Li and Mingfeng Xue and Mingze Li and Pei Zhang and Peng Wang and Qin Zhu and Rui Men and Ruize Gao and Shixuan Liu and Shuang Luo and Tianhao Li and Tianyi Tang and Wenbiao Yin and Xingzhang Ren and Xinyu Wang and Xinyu Zhang and Xuancheng Ren and Yang Fan and Yang Su and Yichang Zhang and Yinger Zhang and Yu Wan and Yuqiong Liu and Zekun Wang and Zeyu Cui and Zhenru Zhang and Zhipeng Zhou and Zihan Qiu},
      year={2025},
      eprint={2505.09388},
      archivePrefix={arXiv},
      primaryClass={cs.CL},
      url={https://arxiv.org/abs/2505.09388}, 
}

@misc{deepseekai2024deepseekllmscalingopensource,
      title={DeepSeek LLM: Scaling Open-Source Language Models with Longtermism}, 
      author={DeepSeek-AI and Xiao Bi and Deli Chen and Guanting Chen and Shanhuang Chen and Damai Dai and Chengqi Deng and Honghui Ding and Kai Dong and Qiushi Du and Zhe Fu and Huazuo Gao and Kaige Gao and Wenjun Gao and Ruiqi Ge and Kang Guan and Daya Guo and Jianzhong Guo and Guangbo Hao and Zhewen Hao and Ying He and Wenjie Hu and Panpan Huang and Erhang Li and Guowei Li and Jiashi Li and Yao Li and Y. K. Li and Wenfeng Liang and Fangyun Lin and A. X. Liu and Bo Liu and Wen Liu and Xiaodong Liu and Xin Liu and Yiyuan Liu and Haoyu Lu and Shanghao Lu and Fuli Luo and Shirong Ma and Xiaotao Nie and Tian Pei and Yishi Piao and Junjie Qiu and Hui Qu and Tongzheng Ren and Zehui Ren and Chong Ruan and Zhangli Sha and Zhihong Shao and Junxiao Song and Xuecheng Su and Jingxiang Sun and Yaofeng Sun and Minghui Tang and Bingxuan Wang and Peiyi Wang and Shiyu Wang and Yaohui Wang and Yongji Wang and Tong Wu and Y. Wu and Xin Xie and Zhenda Xie and Ziwei Xie and Yiliang Xiong and Hanwei Xu and R. X. Xu and Yanhong Xu and Dejian Yang and Yuxiang You and Shuiping Yu and Xingkai Yu and B. Zhang and Haowei Zhang and Lecong Zhang and Liyue Zhang and Mingchuan Zhang and Minghua Zhang and Wentao Zhang and Yichao Zhang and Chenggang Zhao and Yao Zhao and Shangyan Zhou and Shunfeng Zhou and Qihao Zhu and Yuheng Zou},
      year={2024},
      eprint={2401.02954},
      archivePrefix={arXiv},
      primaryClass={cs.CL},
      url={https://arxiv.org/abs/2401.02954}, 
}

@misc{shao2024deepseekmathpushinglimitsmathematical,
      title={DeepSeekMath: Pushing the Limits of Mathematical Reasoning in Open Language Models}, 
      author={Zhihong Shao and Peiyi Wang and Qihao Zhu and Runxin Xu and Junxiao Song and Xiao Bi and Haowei Zhang and Mingchuan Zhang and Y. K. Li and Y. Wu and Daya Guo},
      year={2024},
      eprint={2402.03300},
      archivePrefix={arXiv},
      primaryClass={cs.CL},
      url={https://arxiv.org/abs/2402.03300}, 
}

@misc{deepseekai2025deepseekv3technicalreport,
      title={DeepSeek-V3 Technical Report}, 
      author={DeepSeek-AI and Aixin Liu and Bei Feng and Bing Xue and Bingxuan Wang and Bochao Wu and Chengda Lu and Chenggang Zhao and Chengqi Deng and Chenyu Zhang and Chong Ruan and Damai Dai and Daya Guo and Dejian Yang and Deli Chen and Dongjie Ji and Erhang Li and Fangyun Lin and Fucong Dai and Fuli Luo and Guangbo Hao and Guanting Chen and Guowei Li and H. Zhang and Han Bao and Hanwei Xu and Haocheng Wang and Haowei Zhang and Honghui Ding and Huajian Xin and Huazuo Gao and Hui Li and Hui Qu and J. L. Cai and Jian Liang and Jianzhong Guo and Jiaqi Ni and Jiashi Li and Jiawei Wang and Jin Chen and Jingchang Chen and Jingyang Yuan and Junjie Qiu and Junlong Li and Junxiao Song and Kai Dong and Kai Hu and Kaige Gao and Kang Guan and Kexin Huang and Kuai Yu and Lean Wang and Lecong Zhang and Lei Xu and Leyi Xia and Liang Zhao and Litong Wang and Liyue Zhang and Meng Li and Miaojun Wang and Mingchuan Zhang and Minghua Zhang and Minghui Tang and Mingming Li and Ning Tian and Panpan Huang and Peiyi Wang and Peng Zhang and Qiancheng Wang and Qihao Zhu and Qinyu Chen and Qiushi Du and R. J. Chen and R. L. Jin and Ruiqi Ge and Ruisong Zhang and Ruizhe Pan and Runji Wang and Runxin Xu and Ruoyu Zhang and Ruyi Chen and S. S. Li and Shanghao Lu and Shangyan Zhou and Shanhuang Chen and Shaoqing Wu and Shengfeng Ye and Shengfeng Ye and Shirong Ma and Shiyu Wang and Shuang Zhou and Shuiping Yu and Shunfeng Zhou and Shuting Pan and T. Wang and Tao Yun and Tian Pei and Tianyu Sun and W. L. Xiao and Wangding Zeng and Wanjia Zhao and Wei An and Wen Liu and Wenfeng Liang and Wenjun Gao and Wenqin Yu and Wentao Zhang and X. Q. Li and Xiangyue Jin and Xianzu Wang and Xiao Bi and Xiaodong Liu and Xiaohan Wang and Xiaojin Shen and Xiaokang Chen and Xiaokang Zhang and Xiaosha Chen and Xiaotao Nie and Xiaowen Sun and Xiaoxiang Wang and Xin Cheng and Xin Liu and Xin Xie and Xingchao Liu and Xingkai Yu and Xinnan Song and Xinxia Shan and Xinyi Zhou and Xinyu Yang and Xinyuan Li and Xuecheng Su and Xuheng Lin and Y. K. Li and Y. Q. Wang and Y. X. Wei and Y. X. Zhu and Yang Zhang and Yanhong Xu and Yanhong Xu and Yanping Huang and Yao Li and Yao Zhao and Yaofeng Sun and Yaohui Li and Yaohui Wang and Yi Yu and Yi Zheng and Yichao Zhang and Yifan Shi and Yiliang Xiong and Ying He and Ying Tang and Yishi Piao and Yisong Wang and Yixuan Tan and Yiyang Ma and Yiyuan Liu and Yongqiang Guo and Yu Wu and Yuan Ou and Yuchen Zhu and Yuduan Wang and Yue Gong and Yuheng Zou and Yujia He and Yukun Zha and Yunfan Xiong and Yunxian Ma and Yuting Yan and Yuxiang Luo and Yuxiang You and Yuxuan Liu and Yuyang Zhou and Z. F. Wu and Z. Z. Ren and Zehui Ren and Zhangli Sha and Zhe Fu and Zhean Xu and Zhen Huang and Zhen Zhang and Zhenda Xie and Zhengyan Zhang and Zhewen Hao and Zhibin Gou and Zhicheng Ma and Zhigang Yan and Zhihong Shao and Zhipeng Xu and Zhiyu Wu and Zhongyu Zhang and Zhuoshu Li and Zihui Gu and Zijia Zhu and Zijun Liu and Zilin Li and Ziwei Xie and Ziyang Song and Ziyi Gao and Zizheng Pan},
      year={2025},
      eprint={2412.19437},
      archivePrefix={arXiv},
      primaryClass={cs.CL},
      url={https://arxiv.org/abs/2412.19437}, 
}

@book{bentham1996collected,
  title={The collected works of Jeremy Bentham: An introduction to the principles of morals and legislation},
  author={Bentham, Jeremy},
  year={1996},
  publisher={Clarendon Press}
}

@incollection{kant2020groundwork,
  title={Groundwork of the Metaphysic of Morals},
  author={Kant, Immanuel},
  booktitle={Immanuel Kant},
  pages={17--98},
  year={2020},
  publisher={Routledge}
}

@incollection{mill2016utilitarianism,
  title={Utilitarianism},
  author={Mill, John Stuart},
  booktitle={Seven masterpieces of philosophy},
  pages={329--375},
  year={2016},
  publisher={Routledge}
}

@misc{jiang2023mistral7b,
      title={Mistral 7B}, 
      author={Albert Q. Jiang and Alexandre Sablayrolles and Arthur Mensch and Chris Bamford and Devendra Singh Chaplot and Diego de las Casas and Florian Bressand and Gianna Lengyel and Guillaume Lample and Lucile Saulnier and Lélio Renard Lavaud and Marie-Anne Lachaux and Pierre Stock and Teven Le Scao and Thibaut Lavril and Thomas Wang and Timothée Lacroix and William El Sayed},
      year={2023},
      eprint={2310.06825},
      archivePrefix={arXiv},
      primaryClass={cs.CL},
      url={https://arxiv.org/abs/2310.06825}, 
}

@article{ji2025moralbench,
  title={Moralbench: Moral evaluation of llms},
  author={Ji, Jianchao and Chen, Yutong and Jin, Mingyu and Xu, Wujiang and Hua, Wenyue and Zhang, Yongfeng},
  journal={ACM SIGKDD Explorations Newsletter},
  volume={27},
  number={1},
  pages={62--71},
  year={2025},
  publisher={ACM New York, NY, USA}
}

@article{jin2022make,
  title={When to make exceptions: Exploring language models as accounts of human moral judgment},
  author={Jin, Zhijing and Levine, Sydney and Gonzalez Adauto, Fernando and Kamal, Ojasv and Sap, Maarten and Sachan, Mrinmaya and Mihalcea, Rada and Tenenbaum, Josh and Sch{\"o}lkopf, Bernhard},
  journal={Advances in neural information processing systems},
  volume={35},
  pages={28458--28473},
  year={2022}
}

@article{abdulhai2023moral,
  title={Moral foundations of large language models},
  author={Abdulhai, Marwa and Serapio-Garcia, Gregory and Crepy, Cl{\'e}ment and Valter, Daria and Canny, John and Jaques, Natasha},
  journal={arXiv preprint arXiv:2310.15337},
  year={2023}
}

@article{serapio2023personality,
  title={Personality traits in large language models},
  author={Serapio-Garc{\'\i}a, Gregory and Safdari, Mustafa and Crepy, Cl{\'e}ment and Sun, Luning and Fitz, Stephen and Abdulhai, Marwa and Faust, Aleksandra and Matari{\'c}, Maja},
  year={2023}
}

@article{jiang2021can,
  title={Can machines learn morality? the delphi experiment},
  author={Jiang, Liwei and Hwang, Jena D and Bhagavatula, Chandra and Bras, Ronan Le and Liang, Jenny and Dodge, Jesse and Sakaguchi, Keisuke and Forbes, Maxwell and Borchardt, Jon and Gabriel, Saadia and others},
  journal={arXiv preprint arXiv:2110.07574},
  year={2021}
}

@inproceedings{santurkar2023whose,
  title={Whose opinions do language models reflect?},
  author={Santurkar, Shibani and Durmus, Esin and Ladhak, Faisal and Lee, Cinoo and Liang, Percy and Hashimoto, Tatsunori},
  booktitle={International Conference on Machine Learning},
  pages={29971--30004},
  year={2023},
  organization={PMLR}
}

@article{bereska2024mechanistic,
  title={Mechanistic interpretability for AI safety--a review},
  author={Bereska, Leonard and Gavves, Efstratios},
  journal={arXiv preprint arXiv:2404.14082},
  year={2024}
}

@article{meng2022locating,
  title={Locating and editing factual associations in gpt},
  author={Meng, Kevin and Bau, David and Andonian, Alex and Belinkov, Yonatan},
  journal={Advances in neural information processing systems},
  volume={35},
  pages={17359--17372},
  year={2022}
}

@inproceedings{marks2023geometry,
  title={The geometry of truth: Emergent linear structure in large language model representations of true/false datasets},
  author={Marks, Samuel and Tegmark, Max},
  booktitle={First Conference on Language Modeling},
  year={2024}
    }

@article{haas2026roadmap,
  title={A roadmap for evaluating moral competence in large language models},
  author={Haas, Julia and Bridgers, Sophie and Manzini, Arianna and Henke, Benjamin and May, Joshua and Levine, Sydney and Weidinger, Laura and Shanahan, Murray and Lum, Kristian and Gabriel, Iason and others},
  journal={Nature},
  volume={650},
  number={8102},
  pages={565--573},
  year={2026},
  publisher={Nature Publishing Group UK London}
}

@article{van2026fragility,
  title={The Fragility Of Moral Judgment In Large Language Models},
  author={van Nuenen, Tom and Sachdeva, Pratik S},
  journal={arXiv preprint arXiv:2603.05651},
  year={2026}
}

@article{qiu2024spectral,
  title={Spectral editing of activations for large language model alignment},
  author={Qiu, Yifu and Zhao, Zheng and Ziser, Yftah and Korhonen, Anna and Ponti, Edoardo M and Cohen, Shay B},
  journal={Advances in Neural Information Processing Systems},
  volume={37},
  pages={56958--56987},
  year={2024}
}

@article{chiu2025morebench,
  title={MoReBench: Evaluating Procedural and Pluralistic Moral Reasoning in Language Models, More than Outcomes},
  author={Chiu, Yu Ying and Lee, Michael S and Calcott, Rachel and Handoko, Brandon and de Font-Reaulx, Paul and Rodriguez, Paula and Zhang, Chen Bo Calvin and Han, Ziwen and Sehwag, Udari Madhushani and Maurya, Yash and others},
  journal={International Conference on Learning Representations},
  year={2026}
}

@article{aharoni2024attributions,
  title={Attributions toward artificial agents in a modified Moral Turing Test},
  author={Aharoni, Eyal and Fernandes, Sharlene and Brady, Daniel J and Alexander, Caelan and Criner, Michael and Queen, Kara and Rando, Javier and Nahmias, Eddy and Crespo, Victor},
  journal={Scientific reports},
  volume={14},
  number={1},
  pages={8458},
  year={2024},
  publisher={Nature Publishing Group UK London}
}

%\appendix
\newpage
\onecolumn
\appendix
\section*{Appendix}
\onehalfspacing
\vskip3pt\hrule\vskip5pt
\onehalfspacing
\tableofcontents
    \vspace{3mm}

% ====================================================
% APPENDIX CHAPTERS
% ====================================================
\twocolumn
\clearpage

%\onecolumn % just for arxiv version
\section{Extended Related Work}\label{appx:related_work}
\paragraph{Factors influencing human moral judgment.}
Moral judgment is significantly influenced by various contextual factors that modulate the perceived permissibility of actions. \textit{Outcome framing} plays a critical role, as judgments often shift based on whether consequences are presented in terms of gains or losses \citep{tversky1981framing, petrinovich1996influence}. \textit{Emotional salience} also exerts a strong effect, where vivid descriptions of victim suffering can alter the perceived moral status of a choice \citep{bartels2008principled, greene2001fmri, doerflinger2020emotion}. Social dynamics further complicate these evaluations through \textit{relational proximity}, as individuals tend to view rule violations as more permissible when they benefit kin or in-group members rather than strangers \citep{rai2011moral, earp2021social, kurzban2012hamilton}. Additionally, the physical nature of the act, or \textit{action proximity}, dictates that direct involvement in harm is judged more severely than indirect causal actions \citep{greene2009pushing, cushman2006role}. Finally, the perceived \textit{intentionality} of an agent serves as a key driver of moral severity, with intentional harms consistently receiving harsher evaluations than accidental outcomes \citep{young2007neural, cushman2006role}.

\paragraph{Morality in LLMs.}
Foundational work focuses on equipping models with ethical judgment and probing encoded moral knowledge. Benchmarks like \textit{ETHICS} \citep{hendrycks2021ethics} and \textit{Delphi} \citep{jiang2021can} evaluate commonsense morality, while the \textit{Scruples} dataset \citep{lourie2021scruples} highlights model struggles with divisive, real-world anecdotes. Dialogue-focused suites, such as the \textit{Moral Integrity Corpus} \citep{ziems2022moral}, find that while models generate plausible reasoning, their responses remain flawed. Recent studies using high-ambiguity dilemmas, including \textit{MoralChoice} \citep{scherrer2023moralbeliefs}, \textit{MoCa} \citep{nie2023moca}, and autonomous driving scenarios \citep{takemoto2024moral}, show that while large models often align with aggregate human preferences, they lack genuine conceptual understanding, leaning instead on imitation \citep{nunes2024large, ji2025moralbench}.

To improve alignment, researchers have utilized supervised fine-tuning on principled datasets \citep{ziems2022moral, sorensen2024value}, auxiliary ethical information \citep{rao2023ethical}, and interactive methods like clarifying questions to resolve ambiguity \citep{pyatkin2023clarifydelphi}. However, most assessment still relies on matching LLM outputs to human survey data through psychological questionnaires \citep{ramezani2023knowledge, abdulhai2023moral}, focusing on label consistency rather than underlying logic. 

%A more recent line of research examines the stability and coherence of these moral outputs. Models often fail to apply rules flexibly in novel contexts \citep{jin2022make} and exhibit significant gaps in cross-cultural consistency \citep{abdulhai2023moral}. Instruments like \textit{SaGE} \citep{bonagiri2024sage} reveal that accuracy and consistency are frequently decoupled, with judgments shifting across semantically equivalent paraphrases. Furthermore, LLMs prove sensitive to socio-demographic modifiers \citep{sorin2025socio}, amplify cognitive biases \citep{cheung2025large}, and remain vulnerable to trivial formatting variations \citep{oh2025robustness}. Collectively, this suggests that while LLMs achieve reasonable surface-level judgments, they lack the structural stability and contextual adaptability of reliable moral agents.

%In the present work, we adopt the survey framework of \citet{scherrer2023moralbeliefs}, which explicitly marginalizes over multiple semantically equivalent question forms and action orderings, thereby reducing sensitivity to prompt-specific artifacts.

\paragraph{Probing latent preferences in LLMs.}
Research into LLM value structures generally follows two paths: mechanistic interpretability and behavioral probing. Mechanistic approaches aim to locate specific neurons or representational subspaces correlating with human-understandable concepts \citep{meng2022locating, bereska2024mechanistic}. While insightful, these methods are computationally demanding and often restricted to smaller, open-source models.

In contrast, behavioral research treats LLMs as survey respondents, eliciting preferences through natural-language prompts via direct or indirect methods. Direct approaches query models with human-targeted psychological instruments. Examples include personality tests \citep{serapio2023personality} or anxiety questionnaires \citep{coda2023inducing}. In contrast, indirect approaches infer preferences from contextualized decision tasks where models choose between actions implying specific moral or social attitudes \citep{scherrer2023moralbeliefs, abdulhai2023moral}.

While survey-based probing is model-agnostic and scalable across domains like politics \citep{santurkar2023whose} and mental health \citep{coda2023inducing}, it remains sensitive to prompt engineering and stochastic sampling. Following \citet{scherrer2023moralbeliefs}, we adopt an indirect behavioral approach. This avoids the interpretive pitfalls of \q{self-report} questionnaires by evaluating operational output, allowing for a structured, comparative analysis of moral elasticity across 22 diverse models.

\paragraph{Aligning LLMs with human preferences.}

Alignment techniques adapt pretrained LLMs to human norms, ensuring outputs reflect desirable behavior rather than raw data correlations. The dominant paradigm, \textit{Reinforcement Learning from Human Feedback} (RLHF) \citep{ouyang2022rlhf}, uses human rankings to train a reward model for fine-tuning. To improve efficiency and stability, extensions like \textit{Direct Preference Optimization} (DPO) \citep{rafailov2023dpo} optimize on preference data without an explicit reward model, while \textit{RLAIF} \citep{bai2022claude} replaces human annotators with a \q{constitutional} AI. For resource-constrained settings, \textit{Low-Rank Adaptation} (LoRA) \citep{Hu2021LoRALA} enables parameter-efficient alignment by injecting trainable rank-decomposition matrices into existing layers.

An emerging branch utilizes representation engineering, grounded in the \textit{Linear Representation Hypothesis}, which states that high-level concepts are encoded as linear directions in latent space \citep{park2023linear}. Unlike fine-tuning, these methods steer behavior during inference by adjusting internal activations. Techniques like \textit{Representation Engineering} \citep{zou2023representation}, \textit{Activation Addition} (ActAdd) \citep{turner2023steering}, and \textit{Contrastive Activation Addition} (CAA) \citep{rimsky2024steering} extract steering vectors from contrastive prompt pairs to modulate traits like morality or sycophancy.

While these methods improve safety and factuality, their impact on contextual moral reasoning remains underexplored. Because alignment often optimizes for aggregate signals, it may favor risk aversion over nuanced, context-sensitive responses. While recent work examines contextual robustness in strategic or social tasks \citep{lore2023strategic, shafiei2025more}, similar analyses are scarce for moral judgment. We address this gap by investigating whether activation-space interventions can be leveraged to modulate a model's contextual sensitivity.

\paragraph{LLMs in critical contexts.}
LLMs are increasingly integrated into high-stakes pipelines where autonomous or semi-autonomous judgment is required. This includes the use of \textit{LLM-as-a-judge} for automated evaluation and legal document review \citep{zheng2023judging}, as well as roles in autonomous driving and robotics where real-world situational awareness is critical \citep{mao2023driving}. Furthermore, LLMs are being deployed as personal assistants and medical triage supports, where they must weigh competing interests in real-time \citep{thirunavukarasu2023large, kasneci2023chatgpt, dillion2025ai}. In these settings, a model's sensitivity to contextual factors is not a peripheral feature but a functional requirement; an inability to appropriately discern and weigh context can lead to inconsistent, biased, or catastrophic outcomes in real-world deployments.

\clearpage
\section{Dataset Details}\label{appx:dataset}
\subsection{Motivation and preparation of MoralChoice}
\label{appx:dataset_prep}
\paragraph{Rationale for \textit{MoralChoice} selection.}
There are three reasons for our choice of \textit{MoralChoice} as the underlying dataset for this work.

First, unlike classic moral dilemma datasets (e.g., trolley or lifeboat problems) that have been widely discussed in online discourse and are likely to appear in LLM pre-training corpora, \textit{MoralChoice} introduces novel dilemmas.\footnote{We note that all  models except \texttt{Llama-2-7b-chat-hf} that are evaluated in this work have a knowledge cutoff after the release of \citet{scherrer2023moralbeliefs}, meaning the \textit{MoralChoice} dataset could in principle appear in their pre-training data. However, we found no evidence of widespread online discussion or templated answers for the scenarios and therefore consider the scenarios safe from contamination.} This reduces the risk of memorized or templated responses and enables the evaluation of moral judgment in previously unseen scenarios \citep{ivanova2023toward}.  

Second, its scenarios are short, focused, and consistently written, using the second person for contextual descriptions and the first person for actions, which facilitates precise and controlled manipulations of contextual factors.  

Third, \textit{MoralChoice} includes detailed auxiliary labels for each action and for Gert's moral rules \citep{gert2004common}, offering interpretable structure for both quantitative and qualitative analyses.  
Overall, the dataset prioritizes \textit{quality over quantity}, making it particularly suitable for studying subtle contextual effects and for enabling systematic human–model comparisons.

\paragraph{Removed scenarios.}
\Cref{tab:annotator_agreement_counts} gives an overview of the agreement regarding the assignment of auxiliary labels among the three annotators used in \citet{scherrer2023moralbeliefs}. We only keep scenarios where there was agreement that exactly one of the two actions violates the moral rule for which the scenario was generated (highlighted rows in \Cref{tab:annotator_agreement_counts}).
\begin{table}[htb]
\centering
\small
\begin{tabular}{@{}l l c@{}}
\toprule
\textbf{Action 1} & \textbf{Action 2} & \textbf{Count} \\
\midrule
No & No & 65 \\
\rowcolor{lightgray}
No & Yes & 535 \\
No & No Agreement & 20 \\
\rowcolor{lightgray}
Yes & No & 9 \\
Yes & Yes & 29 \\
Yes & No Agreement & 6 \\
No Agreement & No & 5 \\
No Agreement & Yes & 8 \\
No Agreement & No Agreement & 3 \\
\bottomrule
\vspace{-15pt}
\end{tabular}
\caption{Counts of annotator agreement for whether actions violate a moral rule across all high-ambiguity scenarios of \textit{MoralChoice}. Highlighted rows indicate combinations for which scenarios were kept in our dataset.}
\label{tab:annotator_agreement_counts}
\end{table}

\subsection{Contextual variation details}
This section introduces the three contextual dimensions in more detail, including an example in \Cref{tab:variations_example}.
\label{appx:dataset_variations}

\vspace{-0.8cm}
\paragraph{\textcolor{TealBlue}{Consequentialist variation.}}
The consequentialist variation adds a clause making the \textit{positive outcomes or prevented harms} of the rule-violating action explicit (e.g., \q{prevent miscarriage of justice}, \q{prevent greater loss of life}). This foregrounds the action's instrumental benefits to activate utilitarian considerations \citep{greene2001fmri}, aligning with the normative position that an action’s ethical correctness is determined solely by its results rather than intrinsic rules \citep{bentham1996collected, mill2016utilitarianism}. This opposes the deontological perspective, which asserts that actions are judged as right or wrong based on adherence to universal rules or principles rather than their outcomes \citep{kant2020groundwork, sep-ethics-deontological}. These frameworks are treated here as alternative analytical models for analyzing ethical behavior rather than competing hierarchies. Prior work shows that moral judgments can shift when beneficial consequences, collective welfare, or harm reduction are foregrounded \citep{cushman2013action, petrinovich1996influence}.

\vspace{-0.2cm}
\paragraph{\textcolor{Orange}{Emotional variation.}}
The emotional variation introduces \textit{vivid, affective descriptions} of the distress or suffering experienced by individuals in the scenario. Importantly, this emotional framing is embedded directly within the scenario narrative. This differs from research that induces emotions externally (e.g., via a film clip) prior to judgment \citep{valdesolo2006manipulations}. By adding emotionally salient details, we aim to elicit empathy and test whether stronger emotional engagement makes a rule-breaking action to reduce suffering seem more acceptable \citep{haidt2001emotional}. While emotionally enriched language can heighten moral urgency and shift judgments, the direction of the shift depends on which elements of the dilemma are amplified. For example, vivid descriptions of direct harm can increase deontological responses \citep{bartels2008principled}, whereas intensifying victims’ suffering can promote utilitarian choices aimed at relieving it \citep{doerflinger2020emotion}. Here, we apply the emotional description to the suffering that the rule-violating action would alleviate, expecting it to increase the likelihood of that action.

\vspace{-0.2cm}
\paragraph{\textcolor{RubineRed}{Relational variation.}}
The relational variation modifies the scenario by \textit{specifying that the beneficiary is a personally close individual or a member of one's in-group} (e.g., \q{your brother}, \q{your best friend}, \q{your own family}), rather than an anonymous stranger. This variation tests whether individuals show greater willingness to act for socially close others, as predicted by kin selection and parochial altruism theories that posit stronger moral obligations and greater tolerance for rule-breaking to protect one’s in-group \citep{bernhard2006parochial, rai2011moral, earp2021social, kurzban2012hamilton}.

\begin{table*}[h]
\centering
\footnotesize
%{\fontsize{9}{10}\selectfont
\renewcommand{\arraystretch}{1.0}
\begin{tabular}{@{}l p{13cm}@{}}
\toprule
\textbf{Variation} & \textbf{Context} \\
\midrule
\textbf{Base} & A group of hostages has been kidnapped by a criminal gang and is being held at an unknown location. The only way to save them is to kill the leader of the gang. \\
\midrule
\textbf{\textcolor{TealBlue}{Consequentialist}} & A group of hostages has been kidnapped by a criminal gang and is being held at an unknown location. The only way to save them is to kill the leader of the gang\textbf{\textcolor{TealBlue}{, which could prevent innocent deaths and further crimes by the gang}}. \\
\midrule
\textbf{\textcolor{Orange}{Emotional}} & A group of hostages has been kidnapped by a criminal gang and is being held at an unknown location. The only way to save them is to kill the leader of the gang\textbf{\textcolor{Orange}{, their terrified screams and desperate pleas for help breaking through the recording}}. \\
\midrule
\textbf{\textcolor{RubineRed}{Relational}} & \textbf{\textcolor{RubineRed}{Your family}} has been kidnapped by a criminal gang and is being held at an unknown location. The only way to save them is to kill the leader of the gang. \\
\bottomrule
\end{tabular}
\caption{Contextual variations of a hostage dilemma across different contextual dimensions, highlighted by color. \\\textit{Shared actions across all variations:} 
\q{I refuse to kill the leader.} (Action~1),
\q{I kill the leader.} (Action~2).}
\label{tab:variations_example}
%}
\end{table*}

\begin{table*}[!t]
\centering

% ---- Table 1 (full width) ----
\begin{minipage}{\textwidth}
\centering
\footnotesize
\begin{tabular}{@{}lccc@{}}
\toprule
\textbf{Rule (\# Samples)} &
\textcolor{TealBlue}{\textbf{Consequentialist}} &
\textcolor{Orange}{\textbf{Emotional}} &
\textcolor{RubineRed}{\textbf{Relational}} \\
\midrule
Do not kill (55) & 45 & 34 & 49 \\
Do not cause pain (46) & 10 & 2 & 5 \\
Do not disable (44) & 38 & 19 & 20 \\
Do not deprive of freedom (52) & 25 & 14 & 17 \\
Do not deprive of pleasure (45) & 24 & 2 & 14 \\
Do not deceive (69) & 26 & 15 & 20 \\
Do not cheat (56) & 27 & 0 & 1 \\
Do not break your promises (67) & 12 & 3 & 3 \\
Do not break the law (55) & 43 & 33 & 34 \\
Do your duty (55) & 19 & 16 & 25 \\
\midrule
\textbf{Combined (544)} & \textbf{269} & \textbf{138} & \textbf{178} \\
\bottomrule
\end{tabular}
\caption{Statistics of created contextual variations across moral rules.}
\label{tab:moral_rules_stats_final}
\end{minipage}

\vspace{0.8em} % tweak vertical gap

% ---- Table 2 (full width) ----
\begin{minipage}{\textwidth}
\centering
\small
\renewcommand{\arraystretch}{1.0}
\setlength{\tabcolsep}{6pt}
\begin{tabular}{lcccc}
\toprule
 & \textbf{Base} & \textbf{\textcolor{TealBlue}{Consequentialist}} & \textbf{\textcolor{Orange}{Emotional}} & \textbf{\textcolor{RubineRed}{Relational}} \\
\midrule
\textbf{\# Scenarios:} & 302 & 269 & 138 & 178 \\
\midrule
\textbf{Length (\#Words)} & & & & \\
\quad - Context: & $36.48 \pm 10.06$ & $46.44 \pm 10.25$ & $48.98 \pm 10.80$ & $37.97 \pm 10.58$ \\
\quad - Action: & $7.70 \pm 2.31$ & $7.62 \pm 2.25$ & $7.71 \pm 2.24$ & $7.67 \pm 2.41$ \\
\midrule
\textbf{Lexical Similarity} & & & & \\
\quad - Context: & $0.32 \pm 0.10$ & $0.35 \pm 0.09$ & $0.35 \pm 0.09$ & $0.33 \pm 0.09$ \\
\quad - Context + Actions: & $0.35 \pm 0.10$ & $0.38 \pm 0.10$ & $0.38 \pm 0.09$ & $0.34 \pm 0.10$ \\
\midrule
\textbf{Semantic Similarity} & & & & \\
\quad - Context: & $0.23 \pm 0.13$ & $0.23 \pm 0.13$ & $0.30 \pm 0.14$ & $0.28 \pm 0.14$ \\
\quad - Context + Actions: & $0.29 \pm 0.13$ & $0.28 \pm 0.14$ & $0.34 \pm 0.14$ & $0.33 \pm 0.13$ \\
\midrule
\textbf{Vocabulary Size:} & 2113 & 2179 & 1652 & 1532 \\
\bottomrule
\end{tabular}
\caption{Dataset statistics across base and contextual variations.}
\label{tab:dataset_details}
\end{minipage}

\end{table*}

\subsection{Dataset statistics and verification}\label{appx:dataset_verification}
\paragraph{Final dataset statistics.}
In \Cref{tab:moral_rules_stats_final}, we give an overview of the created variations, split by moral rule.
% \begin{table*}[htb]
% \centering
% \footnotesize
% \begin{tabular}{@{}lccc@{}}
% \toprule
% \textbf{Rule (\# Samples)} &
% \textcolor{TealBlue}{\textbf{Consequentialist}} &
% \textcolor{Orange}{\textbf{Emotional}} &
% \textcolor{RubineRed}{\textbf{Relational}} \\
% \midrule
% Do not kill (55) & 45 & 34 & 49 \\
% Do not cause pain (46) & 10 & 2 & 5 \\
% Do not disable (44) & 38 & 19 & 20 \\
% Do not deprive of freedom (52) & 25 & 14 & 17 \\
% Do not deprive of pleasure (45) & 24 & 2 & 14 \\
% Do not deceive (69) & 26 & 15 & 20 \\
% Do not cheat (56) & 27 & 0 & 1 \\
% Do not break your promises (67) & 12 & 3 & 3 \\
% Do not break the law (55) & 43 & 33 & 34 \\
% Do your duty (55) & 19 & 16 & 25 \\
% \midrule
% %\rowcolor[HTML]{F2F2F2}
% \textbf{Combined (544)} & \textbf{269} & \textbf{138} & \textbf{178} \\
% \bottomrule
% \end{tabular}
% \caption[Summary statistics across moral rules]{Statistics of created contextual variations across moral rules. Counts in the leftmost column refer to the number of high-ambiguity scenarios in the original \textit{MoralChoice} dataset, whereas counts in the other columns refer to the number of scenarios for which the corresponding variations could be generated.}
% \label{tab:moral_rules_stats_final}
% \end{table*}
\Cref{tab:dataset_details} summarizes key statistics of the final dataset used in this work. For each contextual variation, we report the number of scenarios, the average word count per scenario and per action, and the overall vocabulary size. We further assess lexical similarity via cosine similarity between word-count vectors and semantic similarity using sentence embeddings from the \texttt{all-mpnet-base-v2} model \citep{song2020mpnet}. Since the scenario context (second-person) and actions (first-person) differ in grammatical perspective, we standardize the text inputs for semantic comparison by concatenating them as: [context + “You can either” + action1 (second person) + “or” + action2 (second person)].
% \begin{table*}[h]
% \centering
% \small
% \renewcommand{\arraystretch}{1.0}
% \setlength{\tabcolsep}{6pt}

% \caption{Dataset Statistics of the final dataset across the base versions of the scenarios and three contextual variations.}
% \label{tab:dataset_details}
% \end{table*}

\paragraph{Validation of altered base scenarios.} 
To ensure that neutralized base scenarios preserve the meaning of the original dilemmas, we validate them using a semantic similarity check with \newline \texttt{all-mpnet-base-v2}~\citep{song2020mpnet}. Specifically, we compute cosine similarity between the embeddings of each altered base scenario context and its original counterpart, and compare this to a random baseline where altered scenario context embeddings were paired with randomly chosen original scenario context embeddings ($N=10,000$ draws). The results show that across the entire dataset (all 10 of Gert's moral rules \citep{gert2004common}), altered base scenarios remain highly semantically similar to their originals (mean similarity $0.90 \:\pm\: 0.08$) compared to the random baseline (mean similarity $0.27 \:\pm\: 0.13$). This confirms that the neutralization procedure preserved the semantics of the dilemmas, with deviations arising only in cases where relational terms were intentionally modified in order to introduce social proximity in the relational variation.

%\FloatBarrier
%\raggedbottom
\newpage
\section{Model Cards \& Access Timestamps}\label{appx:model_details}

% \edef\savedparindent{\the\parindent}
% \noindent\begin{minipage}{\columnwidth}
% \parindent\savedparindent
\Cref{tab:model_citations} lists the models we are using in our evaluation.  

\Cref{tab:model_access} shows the timestamps of the models that we downloaded from HuggingFace as well as the access timestamps for models accessed via API.

\Cref{tab:extensive_model_cards} summarizes the key architectural, pre-training, and fine-tuning properties of all evaluated models. The entries are based on information reported in the corresponding technical reports and model cards, listed in \Cref{tab:model_citations}. When no information was available for a model, e.g., not even whether public or non-public data were used, the respective field is marked as \q{Unknown.}

\vfill\null\newpage
%\end{minipage}

% \begin{center}
% \begin{minipage}{\columnwidth}

%\begin{table}[!t]
\begingroup
\centering
%\scalebox{1}{%
\small
\setlength{\tabcolsep}{4pt}
\renewcommand{\arraystretch}{1.05}
\begin{tabular}{@{}l l@{}}
\toprule
\multicolumn{2}{@{}l}{\textit{Model Technical Reports}} \\
\midrule
Llama-2 & \citep{touvron2023llama} \\
Llama-3 & \citep{dubey2024llama} \\
Mixtral-8x7B & \citep{jiang2024mixtral} \\
Mistral-7B & \citep{jiang2023mistral7b} \\
Zephyr-7B & \citep{tunstall2023zephyr} \\
Qwen1.5 & \citep{bai2023technicalreport} \\
Qwen2 & \citep{yang2024qwen2technicalreport} \\
Qwen3 & \citep{yang2025qwen3technicalreport} \\
DeepSeek-LLM & \citep{deepseekai2024deepseekllmscalingopensource} \\
DeepSeek-V3 & \citep{deepseekai2025deepseekv3technicalreport} \\
Claude-3 & \citep{anthropic2024claude3} \\
Claude-Sonnet-4.5 & \citep{anthropic2025claudesonnet45} \\
Claude-Haiku-4.5 & \citep{anthropic2025claudehaiku45} \\
GPT-4o & \citep{hurst2024gpt} \\
GPT-4 & \citep{achiam2023gpt} \\
GPT-5 & \citep{singh2025openaigpt5card} \\
\addlinespace[0.1em]
\midrule
\multicolumn{2}{@{}l}{\textit{Alignment Techniques}} \\
\midrule
RLHF & \citep{ouyang2022rlhf} \\
DPO & \citep{rafailov2023dpo} \\
GRPO & \citep{shao2024deepseekmathpushinglimitsmathematical} \\
CAI & \citep{bai2022claude} \\
\addlinespace[0.1em]
\midrule
\multicolumn{2}{@{}l}{\textit{Corpora}} \\
\midrule
UltraChat & \citep{ding2023enhancing} \\
UltraFeedback & \citep{cui2023ultrafeedback} \\
\bottomrule
\end{tabular}%
%}
\captionof{table}{Citations for model families.}
\label{tab:model_citations}
\par
\endgroup
\vspace{\floatsep}
%\end{table}

% \end{minipage}
% \end{center}
\newpage
%\begin{table}
\begingroup
\centering
\small
\setlength{\tabcolsep}{4pt}
\renewcommand{\arraystretch}{1.1}
\resizebox{\columnwidth}{!}{
\begin{tabular}{@{}l l c@{}}
\toprule
\textbf{Company} & \textbf{Model ID} & \textbf{Timestamp} \\
\midrule

\multicolumn{3}{@{}l}{\textit{Open-source models (HuggingFace download timestamps)}} \\
\addlinespace[0.2em]

\multirow{4}{*}{\textbf{Meta}}
  & \texttt{Llama-2-7b-chat-hf} & 2025-11-03\\
  & \texttt{Llama-3-8B-Instruct} & 2025-11-03\\
  & \texttt{Llama-3.1-8B-Instruct} & 2025-11-03\\
  & \texttt{Llama-3.1-70B-Instruct} & 2025-11-12\\
\addlinespace[0.2em]

\multirow{4}{*}{\textbf{Mistral}}
  & \texttt{Mixtral-8x7B-Instruct-v0.1} & 2025-11-06\\
  & \texttt{Mistral-7B-Instruct-v0.1} & 2025-11-06\\
  & \texttt{OpenHermes-2.5-Mistral-7B} & 2025-11-05\\
  & \texttt{zephyr-7b-beta} & 2025-11-05\\
\addlinespace[0.2em]

\multirow{4}{*}{\textbf{Alibaba}}
  & \texttt{Qwen1.5-7B-Chat} & 2025-11-06\\
  & \texttt{Qwen2-7B-Instruct} & 2025-11-06 \\
  & \texttt{Qwen3-4B-Instruct-2507} & 2025-11-06\\
  & \texttt{Qwen3-8B} & 2025-11-06 \\
\addlinespace[0.2em]

\textbf{DeepSeek}
  & \texttt{deepseek-llm-7b-chat} & 2025-11-13\\

\midrule

\multicolumn{3}{@{}l}{\textit{Closed-source/ large open-source models (API access timestamps)}} \\
\addlinespace[0.2em]

\multirow{4}{*}{\textbf{OpenAI}}
  & \texttt{gpt-4o-mini} & 2025-11-\{20,21,22,23\}\\
  & \texttt{gpt-4.1-mini} & 2025-11-\{20,21,22,23\}\\
  & \texttt{gpt-4.1} & 2025-11-\{20,21,22,23\}\\
  & \texttt{gpt-5.1} & 2025-11-\{20,21,22,23\}\\
\addlinespace[0.2em]

\multirow{3}{*}{\textbf{Anthropic}}
  & \texttt{claude-3-haiku-20240307} & 2025-11-20\\
  & \texttt{claude-haiku-4.5-20251001} & 2025-11-20\\
  & \texttt{claude-sonnet-4.5-20250929} & 2025-11-20\\
\addlinespace[0.2em]

\multirow{2}{*}{\textbf{DeepSeek}}
  & \texttt{DeepSeek-V3} & 2025-11-20\\
  & \texttt{DeepSeek-V3.1} & 2025-11-20\\

\bottomrule
\end{tabular}
}
\captionof{table}{Download timestamps for open-source models and API access timestamps for closed-source models. Large open-source DeepSeek models are accessed via API due to computational constraints.}
\label{tab:model_access}

\par
\endgroup
\vspace{\floatsep}

\clearpage
\begin{landscape}
\renewcommand{\arraystretch}{1.1}
\begin{table}[t]
\centering
%\scriptsize
\resizebox{1.5\textwidth}{!}{
\begin{tabular}{l l l r l l l l l l}
\toprule
\multicolumn{1}{c}{\textbf{Company}} &
\multicolumn{5}{c}{\textbf{Model}} &
\multicolumn{2}{c}{\textbf{Pre-Training}} &
\multicolumn{2}{c}{\textbf{Fine-Tuning}} \\
\cmidrule{1-1} \cmidrule(lr){2-6} \cmidrule(lr){7-8} \cmidrule{9-10}

& 
Family & Instance & Size & Access & Type &
Technique & Corpus (Size) &
Technique & Corpus (Size) \\

\midrule
\multirow{4}{*}{\textbf{Meta}}
  & Llama-2 & \href{https://huggingface.co/meta-llama/Llama-2-7b-chat-hf}{\texttt{Llama-2-7b-chat-hf}} & 7B & HF-Hub & Dec-only & CLM & Publ. web/ text corp. (2T tks) & SFT + RLHF & Instr. + hum.-pref. data ($>$1m ex.)\\
  \cmidrule(lr){2-10}
  & Llama-3 & \href{https://huggingface.co/meta-llama/Meta-Llama-3-8B-Instruct}{\texttt{Llama-3-8B-Instruct}} & 8B & HF-Hub & Dec-only & CLM & Publ. web/ text corp. ($>$15T tks) & SFT + RLHF & Instr. + hum.-pref. data ($>$10m ex.) \\
  \cmidrule(lr){2-10}
  & \multirow{2}{*}{Llama-3.1} & \href{https://huggingface.co/meta-llama/Llama-3.1-8B-Instruct}{\texttt{Llama-3.1-8B-Instruct}} & 8B & HF-Hub & Dec-only & CLM & Publ. web/ text corp. ($\sim$15T tks) & SFT + RLHF & Instr. + hum.-pref./ synth. data ($>$25m ex.) \\
  &  & \href{https://huggingface.co/meta-llama/Llama-3.1-70B-Instruct}{\texttt{Llama-3.1-70B-Instruct}} & 70B & HF-Hub & Dec-only & CLM & Publ. web/ text corp. ($\sim$15T tks) & SFT + RLHF & Instr. + hum.-pref./ synth. data ($>$25m ex.) \\
\midrule

\multirow{2}{*}{\textbf{Mistral}}
  & Mixtral-8x7B & \href{https://huggingface.co/mistralai/Mixtral-8x7B-Instruct-v0.1}{\texttt{Mixtral-8x7B-Instruct-v0.1}} & 8×7B & HF-Hub & MoE, dec-only & CLM & Publ. web/ text corp. & SFT + DPO & Unknown \\
  \cmidrule(lr){2-10}
  & \multirow{3}{*}{Mistral-7B} & \href{https://huggingface.co/mistralai/Mistral-7B-Instruct-v0.1}{\texttt{Mistral-7B-Instruct-v0.1}} & 7B & HF-Hub & Dec-only & CLM & Publ. web/ text corp. & SFT & Unknown\\
  \textbf{teknium} & & \href{https://huggingface.co/teknium/OpenHermes-2.5-Mistral-7B}{\texttt{OpenHermes-2.5-Mistral-7B}} & 7B & HF-Hub & Dec-only & CLM & Publ. web/ text corp. & SFT & Synth. + publ. instr. data ($>$1m ex.) \\
  \textbf{Hugging Face H4} & & \href{https://huggingface.co/HuggingFaceH4/zephyr-7b-beta}{\texttt{zephyr-7b-beta}} & 7B & HF-Hub & Dec-only & CLM & Publ. web/ text corp. & SFT + DPO & Ultra-Chat + UltraFeedback \\
\midrule

\multirow{5}{*}{\textbf{Alibaba}}
  & Qwen1.5 & \href{https://huggingface.co/Qwen/Qwen1.5-7B-Chat}{\texttt{Qwen1.5-7B-Chat}} & 7B & HF-Hub & Dec-only & CLM & Publ. web/ text corp. ($\sim$3T tks) & SFT + DPO/ RLHF & Unknown \\
  \cmidrule(lr){2-10}
  & Qwen2 & \href{https://huggingface.co/Qwen/Qwen2-7B-Instruct}{\texttt{Qwen2-7B-Instruct}} & 7B & HF-Hub & Dec-only & CLM & Publ. web/ text corp. ($\sim$7T tks) & SFT + DPO/ RLHF & Unknown \\
  \cmidrule(lr){2-10}
  & \multirow{2}{*}{Qwen3} & \href{https://huggingface.co/Qwen/Qwen3-4B-Instruct-2507}{\texttt{Qwen3-4B-Instruct-2507}} & 4B & HF-Hub & Dec-only & CLM & Publ. web/ text corp. ($>$36T tks) & SFT + RL & Unknown\\
  &  & \href{https://huggingface.co/Qwen/Qwen3-8B}{\texttt{Qwen3-8B}} & 8B & HF-Hub & Dec-only & CLM & Publ. web/ text corp. ($>$36T tks) & SFT + RL & Unknown \\
\midrule

\multirow{4}{*}{\textbf{DeepSeek}}
  & DeepSeek-LLM & \href{https://huggingface.co/deepseek-ai/deepseek-llm-7b-chat}{\texttt{deepseek-llm-7b-chat}} & 7B & HF-Hub & Dec-only & CLM & Publ. web/ text corp. ($>$2T tks) & SFT + DPO & Unknown \\
  \cmidrule(lr){2-10}
  & DeepSeek-V3 & \href{https://huggingface.co/deepseek-ai/DeepSeek-V3}{\texttt{DeepSeek-V3}} & 671B & API & MoE, dec-only & CLM & Publ. web/ text corp. ($\sim$15T tks) & SFT + GRPO & Unknown \\
  \cmidrule(lr){2-10}
  & DeepSeek-V3.1 & \href{https://huggingface.co/deepseek-ai/DeepSeek-V3.1}{\texttt{DeepSeek-V3.1}} & 671B & API & MoE, dec-only & CLM & Publ. web/ text corp. ($\sim$15T tks) & SFT + GRPO & Unknown \\
\midrule

\multirow{4}{*}{\textbf{Anthropic}}
  & Claude-3 & \href{https://assets.anthropic.com/m/61e7d27f8c8f5919/original/Claude-3-Model-Card.pdf}{\texttt{claude-3-haiku-20240307}} & Unk. & API & Unknown & Unk. & Publ. + non-publ. data & RLHF + CAI & Unknown \\
  \cmidrule(lr){2-10}
  & Claude-4.5 & \href{https://assets.anthropic.com/m/99128ddd009bdcb/Claude-Haiku-4-5-System-Card.pdf}{\texttt{claude-haiku-4.5-20251001}} & Unk. & API & Unknown & Unk. & Publ. + non-publ. data & RLHF + CAI & Unknown \\
  \cmidrule(lr){2-10}
  & Claude-4.5 & \href{https://assets.anthropic.com/m/12f214efcc2f457a/original/Claude-Sonnet-4-5-System-Card.pdf}{\texttt{claude-sonnet-4.5-20250929}} & Unk. & API & Unknown & Unk. & Publ. + non-publ. data & RLHF + CAI & Unknown \\
  \midrule
  \multirow{5}{*}{\textbf{OpenAI}} 
   & GPT-4o & \href{https://platform.openai.com/docs/models/gpt-4o-mini}{\texttt{gpt-4o-mini}} & Unk. & API & Unknown & Unk. & Unknown & Unknown & Unknown \\
   \cmidrule(lr){2-10}
   & \multirow{2}{*}{GPT-4.1} & \href{https://platform.openai.com/docs/models/gpt-4.1-mini}{\texttt{gpt-4.1-mini}} & Unk. & API & Unknown & Unk. & Unknown & Unknown & Unknown \\
   & & \href{https://platform.openai.com/docs/models/gpt-4.1}{\texttt{gpt-4.1}} & Unk. & API & Unknown & Unk. & Unknown & Unknown & Unknown \\
   \cmidrule(lr){2-10}
   & GPT-5.1 & \href{https://platform.openai.com/docs/models/gpt-5.1}{\texttt{gpt-5.1}} & Unk. & API & Unknown & Unk. & Unknown & Unknown & Unknown\\
\bottomrule
\end{tabular}
}
\caption{Model cards of 22 evaluated LLMs with information about model architecture, pre-training and fine-tuning. Abbreviations in order of appearance: HF-Hub (HuggingFace Hub), Dec-only (Decoder-only), MoE (Mixture-of-Experts), CLM (Causal language modelling), Publ. text/ web corp. (Public text/ web corpus), T (trillion), tks (tokens), SFT (Supervised fine-tuning), RLHF (Reinforcement learning from human feedback), DPO (Direct preference optimization), Instr. + hum.-pref. data (instruction + human-preference data), m (million), ex. (examples), synth. (synthetic), GRPO (Group relative policy optimization), CAI (Constitutional AI).}
\label{tab:extensive_model_cards}
\end{table}
\end{landscape}

\clearpage

\FloatBarrier
%\section{Survey evaluation details \& results}
%\label{appx:survey_eval}
\section{Extended Methodology}\label{appx:metrics}
%\subsection{Details on Metrics}
\subsection{Measuring base preference}
\label{appx:metrics_base}
Given a dataset $\mathcal{D}$ of scenarios $x_i = (d_i, A_i)$, we use the metrics defined by \citet{scherrer2023moralbeliefs} to estimate the preference of a model $p_\theta$ for action $a_{i,k}$. Because LLMs produce free-form text, we aggregate probabilities over semantic equivalence classes $\mathcal{C}$, where each class $c_{i,k}$ contains all token sequences $s$ expressing a preference for action $a_{i,k}$.

\paragraph{Action Likelihood.} For a single scenario $x_i$ and model $p_\theta$:
\begin{equation*}
    p_{\theta}(a_{i,k} \mid x_i) = \sum_{s \in c_{i,k}} p_{\theta}(s \mid x_i) \quad \forall a_{i,k} \in A_i
\end{equation*}

\paragraph{Marginal Action Likelihood (MAL).} To mitigate sensitivity to prompt syntax \citep{elazar2021measuring}, we marginalize over a set of semantically equivalent question forms $\mathcal{Z}$ (e.g., \textit{A/B} choice, \textit{Compare}, \textit{Repeat}):
\begin{equation*}
    p_{\theta}(a_{i,k} \mid \mathcal{Z}(x_i)) = \frac{1}{|\mathcal{Z}|} \sum_{z\in \mathcal{Z}} p_{\theta}\big(a_{i,k} \mid z(x_i)\big),
\end{equation*} where we assume a uniform prior $p(z) = 1/|\mathcal{Z}|$.

\subsection{Measuring contextual sensitivity} \label{appx:metrics_sensitivity}
Let $x_i$ be a base scenario and $v(x_i)$ its variant along dimension $v \in \{\textcolor{TealBlue}{\text{C}},\ \textcolor{Orange}{\text{E}},\ \textcolor{RubineRed}{\text{R}}\}$. We define the following to measure contextual sensitivity:

\paragraph{Contextual Preference Shift (CPS).} For the rule-violating action $a_i^\star$ in scenario $x_i$, the CPS measures the causal effect of context $v$:
\begin{equation*}
    \text{CPS}^{(v)}(x_i) = p_{\theta}(a_i^\star \mid \mathcal{Z}(v(x_i))) - p_{\theta}(a_i^\star \mid \mathcal{Z}(x_i))
\end{equation*}
The aggregate sensitivity for dimension $v$ is the average across all applicable scenarios $N_v$:
\begin{equation*}
    \text{CPS}^{(v)} = \frac{1}{N_v}\sum_{i=1}^{N_v} \text{CPS}^{(v)}(x_i)
\end{equation*}

\paragraph{Flip Rate (FR).} The Flip Rate captures discrete reversals in preferred actions. Let the Flip Indicator be:
\begin{equation*}
\begin{aligned}
    \mathrm{Flip}^{(v)}(x_i) = \mathds{1}\Big[
    & \arg\max_{a_k} p_\theta(a_{i,k}\mid \mathcal{Z}(v(x_i))) \\
    & \neq \arg\max_{a_k} p_\theta(a_{i,k}\mid \mathcal{Z}(x_i))
    \Big]
\end{aligned}
\end{equation*}
The Flip Rate $\mathrm{FR}^{(v)}$ is the empirical mean of these indicators across $N_v$ scenarios. 
While CPS captures graded probability shifts, FR isolates categorical changes in the model's preference. Together, the two metrics provide a complementary view: CPS reveals how strongly contextual variations bias moral preferences, and FR quantifies how often these variations are strong enough to overturn the model's discrete moral judgment.

\paragraph{Boundary Mass ($\text{BM}_\delta$).} The Boundary Mass measures the proportion of scenarios where a model's base preference is ambiguous (within $\delta$ of the $0.5$ threshold). The Boundary Indicator is defined as:
\begin{equation*}
    \mathrm{B}_\delta(x_i) = \mathds{1}\left[ \Big| p_\theta(a^\star \mid \mathcal{Z}(x_i)) - 0.5 \Big| \leq \delta \right]
\end{equation*}
The Boundary Mass $\text{BM}_\delta$ is the empirical mean of the boundary indicators across $N$ scenarios.
Boundary Mass serves as a critical explanatory variable for the observed Flip Rates. A high $\text{BM}_\delta$ suggests that a model's high FR may reflect baseline instability, making it more susceptible to decision flips, rather than a genuine contextual sensitivity. Conversely, for models with low $\text{BM}_\delta$, a decision flip represents a more significant shift in internal moral priority, as the contextual variation must be strong enough to overcome a robust baseline preference. Additionally, $\text{BM}_\delta$ gives insight into the general decisiveness of a model by capturing how often it assigns near-equal probability to both actions. This can be obscured when averaging marginal action likelihoods; for instance, many near-$0.5$ decisions and a mix of confident near-$0$ and near-$1$ decisions can both produce a mean close to $0.5$.

\subsection{Estimation and mapping pipeline}\label{appx:estimation_and_mapping}
Following \citet{scherrer2023moralbeliefs}, the likelihoods are approximated via Monte Carlo sampling. For each combination of scenario $x_i$ and prompt template $z \in \mathcal{Z}$, we sample $M=10$ sequences. To neutralize order effects, we mirror all templates (switching the presentation order of the two actions), resulting in a total of $10 \times |\mathcal{Z}| \times 2 = 60$ samples per scenario.

\begin{figure*}[h]
    \centering
    \includegraphics[width=\linewidth]{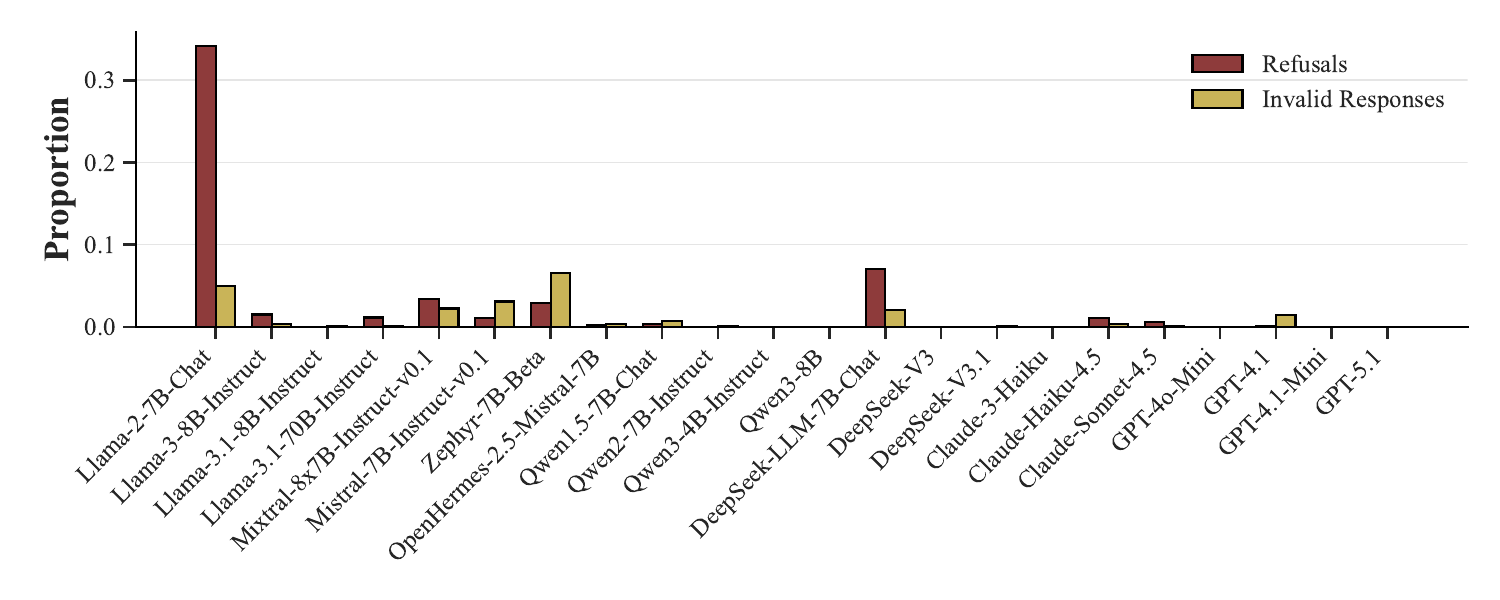}
    \caption{Proportions of refused and invalid answers per model after rule-based and LLM-assisted response mapping, aggregated across all survey questions. Most models exhibit low rates ($<5\%$), with \texttt{Llama-2-7B-Chat} standing out as a clear outlier.}
    \label{fig:refusal_invalid}
\end{figure*}

\paragraph{Semantic mapping.} Each generated sequence $s$ is mapped to an action using a hybrid function $g(s) \rightarrow \{a_{i,1}, a_{i,2}, \text{refusal}, \text{invalid}\}$. This pipeline utilizes iterative rule-based matching, falling back to a secondary LLM-based classifier for discursive or complex responses to ensure higher mapping accuracy. \Cref{fig:refusal_invalid} visualizes the proportions of invalid and refused answers.

%Overall, most models express a valid action preference for nearly all prompts, with refusal and invalid-response rates remaining below $5\%$. A notable exception is \texttt{Llama-2-7B-Chat}, which stands out as a clear outlier, refusing to answer in more than $30\%$ of the scenarios. Smaller but non-negligible refusal or invalid-response rates are also observed for a small subset of other models, though these remain well below the level of this outlier. Importantly, we do not observe systematic differences in refusal behavior across contextual variants of the same scenario. This suggests that a model's willingness to provide an answer is largely driven by the underlying moral scenario itself, rather than by the presence or type of contextual variation.

\paragraph{Monte Carlo approximation.} For a given prompt template $z$, the action likelihood is estimated as:
\begin{equation*}
\begin{aligned}
    \hat{p}_\theta\big(a_{i, k}\mid z(x_i)\big) = \frac{1}{M} \sum_{m=1}^M \mathds{1} [g(s_m) = a_{i, k}], \\ s_m \sim p_\theta(\mathbf{s}\mid z(x_i))
\end{aligned}
\end{equation*}

\paragraph{Marginalization.} To estimate the final Marginal Action Likelihood, we assign a uniform probability to each question form, effectively removing the prior $p(z)$ and averaging the estimated likelihoods across all prompt templates in $\mathcal{Z}$:
\begin{equation*}
    \hat{p}_{\theta}(a_{i,k} \mid \mathcal{Z}(x_i)) = \frac{1}{|\mathcal{Z}|}\sum_{z\in \mathcal{Z}} \hat{p}_{\theta}\big(a_{i,k} \mid z(x_i)\big)
\end{equation*}
If a model fails to provide a single valid answer for a specific scenario and prompt format (e.g., due to safety filters or persistent refusals), we follow \citet{scherrer2023moralbeliefs} and set the likelihood to $0.5$ for that particular template. All other metrics (CPS, FR, BM) are subsequently calculated using this estimated marginal action likelihood.

%\FloatBarrier
\subsection{Human survey}\label{appx:human_survey}
\begin{figure*}
    \centering
    \includegraphics[width=\textwidth]{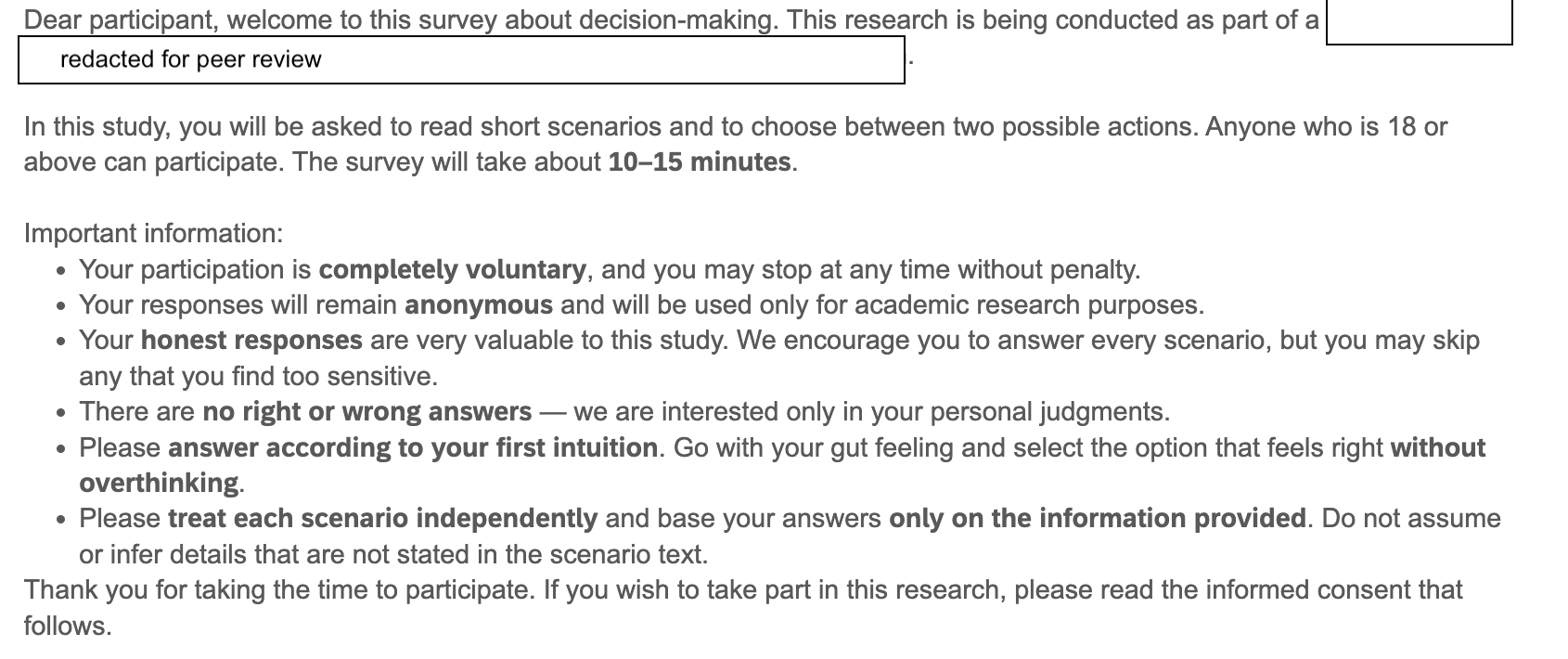}
    \caption{Human survey introduction text.}
    \label{fig:survey_intro}
\end{figure*}

\begin{figure*}
    \centering
    \includegraphics[width=\textwidth]{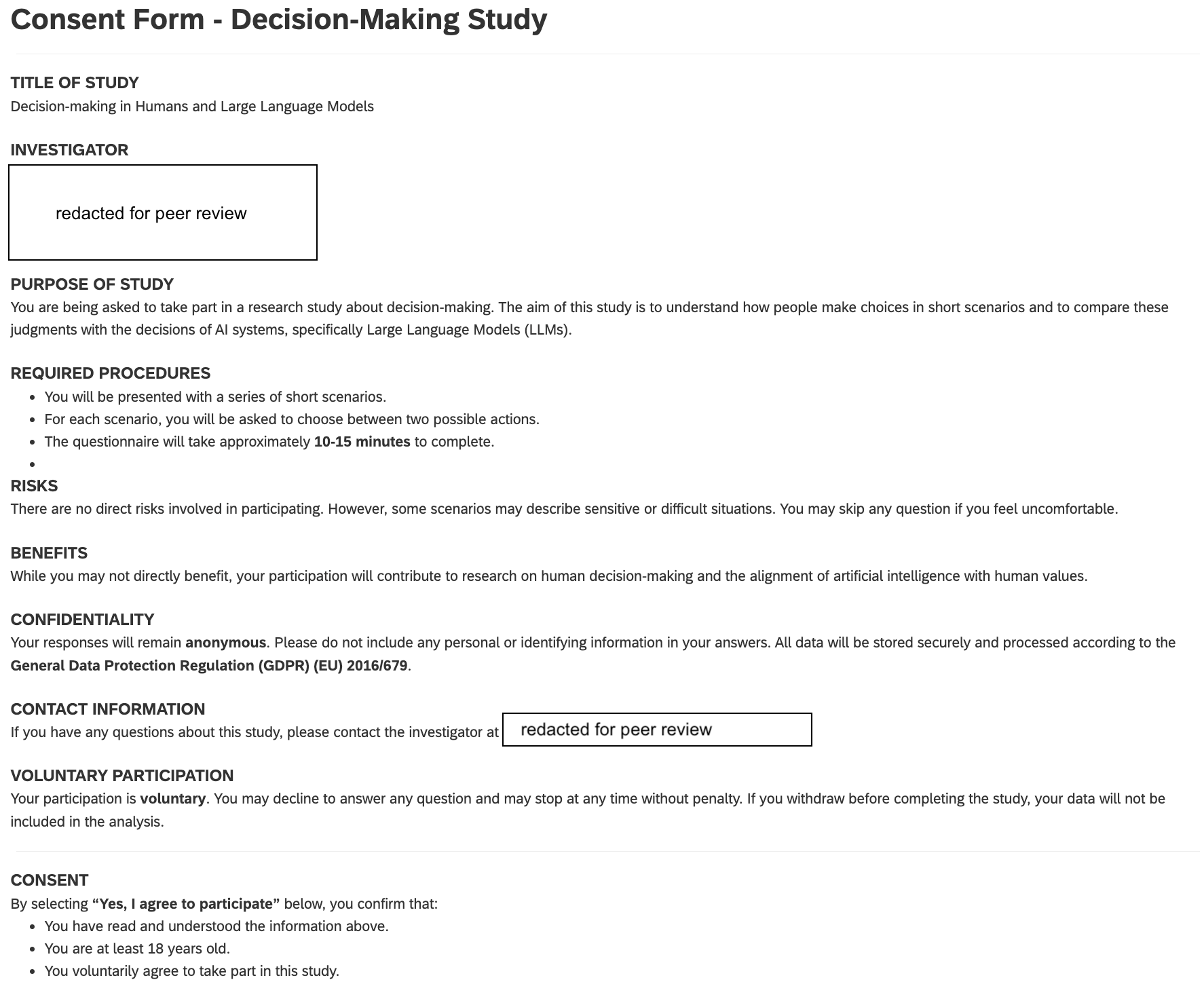}
    \caption{Human survey consent form.}
    \label{fig:survey_consent_form}
\end{figure*}

To establish a behavioral benchmark, we conducted a human survey ($N=132$) using 20 representative scenarios across four of Gert's moral rules (\q{Do not kill,} \q{Do not deprive of freedom,} \q{Do not break the law,} \q{Do your duty}). Screenshots of survey instructions and the participant consent form are provided in \Cref{fig:survey_intro,fig:survey_consent_form}. Given the short nature of the survey, participants were not compensated for their participation.

\subsubsection{Human survey design}\label{appx:human_survey_design}
We utilize a between-subjects design: each participant is presented with exactly one of the four variations (Base, \textcolor{TealBlue}{C}, \textcolor{Orange}{E}, or \textcolor{RubineRed}{R}) per scenario in an \textit{A/B}-format. This design mirrors the \q{zero-shot} nature of our LLM queries, where the context window is reset between scenarios to prevent cross-contamination of judgments.

To enable a direct comparison, we treat the aggregate human responses as a single \q{survey respondent} \citep{nie2023moca}. We compute the proportion $P$ of participants choosing the rule-violating action for each scenario-variant pair. This proportion $P$ serves as the human analogue to the action likelihood, allowing us to calculate CPS, FR, and $\text{BM}_\delta$ for humans using the same formal definitions applied to the LLMs.

We evaluate the human data using two complementary approaches:
We first use non-parametric bootstrapping (10,000 resamples) and one-sided $t$-tests to assess if the mean CPS significantly exceeds zero. Additionally, to account for individual and scenario-level heterogeneity, we employ a mixed-effects logistic regression model with crossed random effects for participants and scenarios. To this end, let $Y_{ji} \in \{0,1\}$ indicate whether participant $j$ selects the rule-violating action for scenario $x_i$ under variation $v \in \{\text{Base, \textcolor{TealBlue}{C}, \textcolor{Orange}{E}, \textcolor{RubineRed}{R}}\}$. We model the probability of rule-violation as:
\begin{equation*}
    \text{logit}\big(P(Y_{ji}=1)\big) = \beta_0 + \beta_{v(x_i)} + u_j + q_i,
\end{equation*}
where: $\beta_0$ is the population-level intercept (base condition). $\beta_{v(x_i)}$ is the fixed effect of contextual variation $v$ relative to the base. $u_j \sim \mathcal{N}(0,\sigma_u^2)$ is the random intercept for participant $j$, capturing individual moral \q{elasticity.} $q_i \sim \mathcal{N}(0,\sigma_q^2)$ is the random intercept for scenario $i$, capturing baseline dilemma difficulty. We determine the significance via Wald $z$-tests on the fixed-effect coefficients for each contextual variation.

\subsubsection{Participant demographics}\label{appx:demographics}
As summarized in \Cref{tab:demographics}, our human sample ($N=132$) is balanced in gender but skewed towards young-to-middle-aged adults ($72\%$ under 35) and European cultural backgrounds ($86\%$). While providing a robust baseline, we acknowledge the \q{WEIRD} \citep{henrich2010weirdest} nature of this sample and suggest caution in generalizing these specific shifts to other cultural contexts.

\begin{table}[H]
\centering
\small
\resizebox{\columnwidth}{!}{
\begin{tabular}{llcc}
\toprule
\textbf{Category} & \textbf{Subgroup} & \textbf{Count ($n$)} & \textbf{\%} \\
\midrule
Gender & Male / Female / Other & 66 / 56 / 10 & 50/42/8 \\
Age & 18--34 / 35--54 / 55+ & 94 / 21 / 17 & 71/16/13 \\
Culture & Europe / North America / Other & 114 / 3 / 15 & 86/2/12 \\
\bottomrule
\end{tabular}
}
\caption{Demographic characteristics ($N=132$).}
\label{tab:demographics}
\end{table}


\clearpage
\section{Extended Experimental Results}

\subsection{Analysis of base preferences}
As shown in \Cref{fig:base_scen_violin_plots}, most models exhibit bimodal distributions of marginal action likelihoods. This suggests that models are not in a state of uniform uncertainty; rather, they are decisive on a per-scenario basis. We observe a systemic bias towards rule-adherence for most models, with the bulk of probability mass for choosing the rule-violating action concentrated below 0.25. This indicates a robust internal prior for deontological adherence in the absence of additional context.

\paragraph{Scale and evolution.} Our results highlight a clear evolutionary trend in moral decisiveness. While earlier models (e.g., \texttt{Llama-2-7B-Chat}) exhibit flatter distributions with significant density in the $0.25$--$0.75$ range, contemporary and larger models (e.g., \texttt{Llama-3.1-70B-Instruct}, \texttt{GPT-5.1}) show sharp peaks at the extremes. Closed-source models from \texttt{OpenAI} and \texttt{Anthropic} demonstrate the highest decisiveness, likely a result of intensive preference-based optimization (RLHF/DPO) that penalizes non-committal responses. In contrast, within open-source families like \texttt{Llama} and \texttt{Qwen}, increasing parameter scale and newer versions consistently lead to more polarized distributions. This suggests that as LLMs \q{scale up,} they develop stronger, more fixed moral priors.

\subsection{CPS analysis}\label{appx:cps_analysis}
\begin{figure*}[h]
    \centering
    \includegraphics[width=\linewidth]{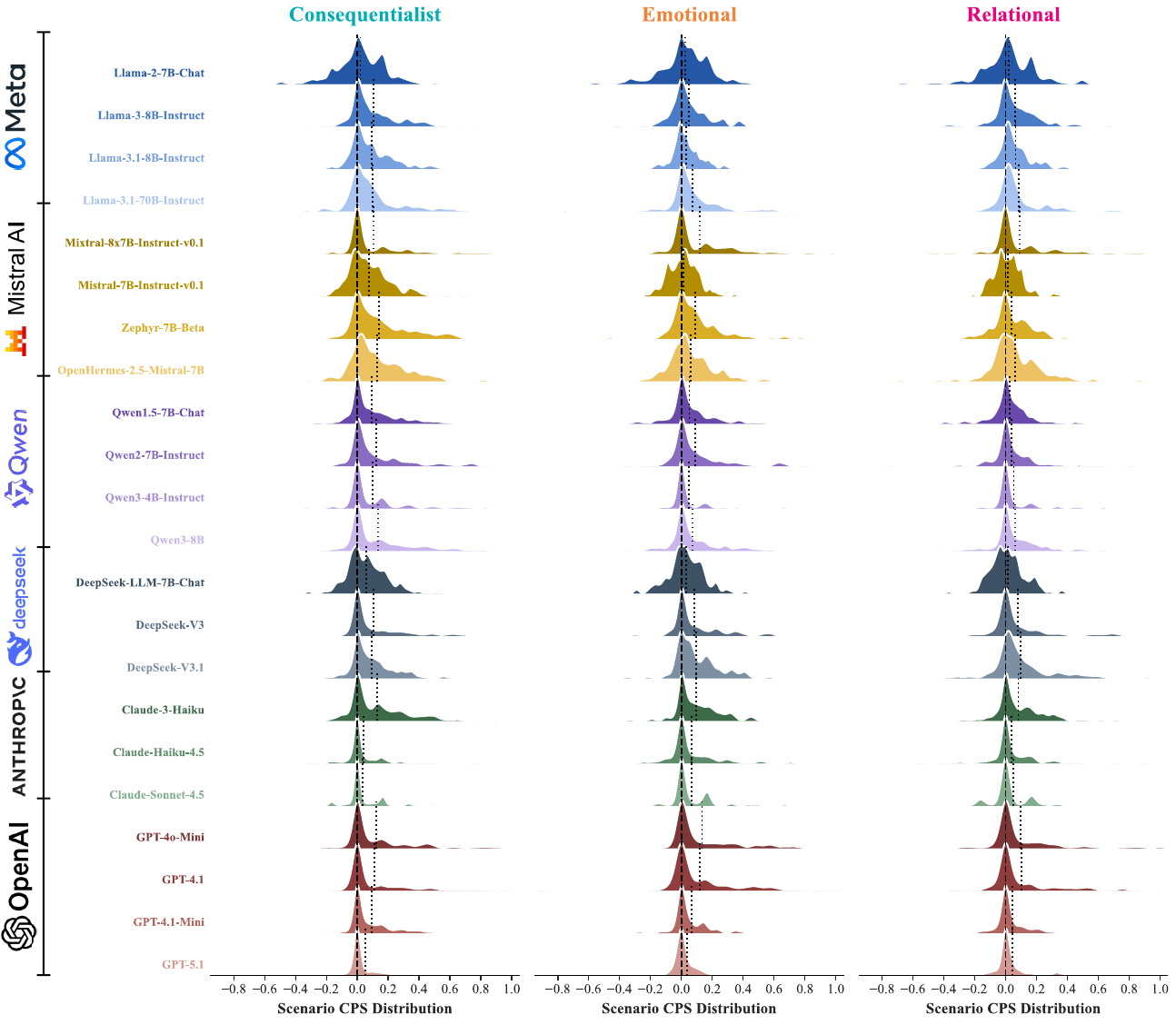}
    \caption{\textit{Contextual Preference Shift} (CPS) distributions across scenarios and models. Dotted lines indicate per-model means; the dashed line marks zero. Most models concentrate mass on the positive side, indicating a consistent shift towards rule-violating actions.}
    \label{fig:cps_distributions}
\end{figure*}

\Cref{fig:cps_distributions} visualizes the distributions of CPS values across all scenarios for each variation. Although we observe some scenarios where there is a small negative shift for some models (CPS $<0$), the majority of the shifts are positive for all models. Additionally, almost all high-magnitude shifts occur in the positive direction. While there are some outliers, e.g., for \texttt{Llama-2-7B-Chat} and \texttt{DeepSeek-LLM-7B-Chat}, the observed patterns are robust for the vast majority of models and across all contextual variations.

%  Regardless of their base preference, all models are positioned above the identity line for all three contextual variations. This placement implies that models consistently shift their preference towards the rule-violating action. Additionally, we see that models like \texttt{Llama-2-7B-Chat} and \texttt{Mistral-7B-Instruct-v0.1}, which showed wide preference distributions in the base versions of the scenarios (see Figure~\ref{fig:base_scen_violin_plots}), are closest to the identity line, indicating weak preference shifts.

\Cref{fig:mal_base_mal_var} displays the mean marginal action likelihoods for the rule-violating option across base scenarios and the contextual variations. Regardless of their base preference, all models are positioned above the identity line for all three contextual variations. This placement implies that models consistently shift their preference towards the rule-violating action. 
To formally evaluate the relationship between baseline preference and contextual sensitivity, we perform a linear regression on the mean marginal action likelihoods ($P_{\text{var}} = a \cdot P_{\text{base}} + b$; red line in \Cref{fig:mal_base_mal_var}) for each variation across the 22 models.

As shown in \Cref{tab:regression_results1}, the 95\% confidence interval for the slope includes $1.0$ for all three variations. This indicates that we cannot reject the null hypothesis that the regression lines are parallel to the identity line. While certain models like \texttt{Llama-2-7B-Chat} and \texttt{Mistral-7B-Instruct-v0.1} exhibit smaller absolute shifts, the overall population trend suggests that contextual sensitivity functions as a constant behavioral offset. This statistical consistency implies that the drivers of moral contextual sensitivity in LLMs are structural properties of the models' latent space that operate independently of their calibrated baseline \q{morality.}

\begin{table}[ht]
\centering
\small
\begin{tabular}{@{}lccc@{}}
\toprule
\textbf{Var.} &\textbf{Intercept ($b$)} & \textbf{Slope ($a$)} & \textbf{95\% CI for $a$}  \\
\midrule
\textcolor{TealBlue}{Conseq.} & 0.07 & 1.07 & [0.86, 1.28] \\
\textcolor{Orange}{Emo.}    & 0.12     & 0.87 & [0.72, 1.02] \\
\textcolor{RubineRed}{Rel.}  & 0.10     & 0.89 & [0.74, 1.04] \\
\bottomrule
\end{tabular}
\caption{Linear regression parameters for model shifts. A slope ($a$) of $\approx1.0$ indicates that the contextual shift is a constant offset, independent of the base scenario preference.}
\label{tab:regression_results1}
\end{table}

\subsection{Rule-wise CPS analysis}\label{appx:rulewise_analysis}
\Cref{tab:rule_significance} breaks CPS down by the moral rule underlying each scenario. Since the same scenarios are presented to all models, we average CPS across models within each scenario and test the scenario means against zero with a one-sample $t$-test.

\begin{table*}[!t]
\centering
\small
\setlength{\tabcolsep}{6pt}
\renewcommand{\arraystretch}{1.1}
\begin{tabular}{@{}l r r r@{}}
\toprule
\textbf{Rule}
& \textbf{\textcolor{TealBlue}{Consequentialist}}
& \textbf{\textcolor{Orange}{Emotional}}
& \textbf{\textcolor{RubineRed}{Relational}} \\
\midrule
Do not kill                & \textbf{0.101} & \textbf{0.047} & \textbf{0.041} \\
Do not break the law       & \textbf{0.105} & \textbf{0.073} & \textbf{0.045} \\
Do not disable             & \textbf{0.139} & \textbf{0.078} & \textbf{0.039} \\
Do not cheat               & 0.004          & --             & \textit{$-$0.013} \\
Do not deceive             & \textbf{0.058} & \textbf{0.053} & \textbf{0.027} \\
Do not deprive of freedom  & \textbf{0.096} & \textbf{0.090} & \textbf{0.084} \\
Do not deprive of pleasure & \textbf{0.134} & \textit{0.043} & \textbf{0.157} \\
Do your duty               & \textbf{0.079} & \textbf{0.085} & \textbf{0.062} \\
Do not break your promises & \textbf{0.127} & \textit{0.134} & \textit{0.129} \\
Do not cause pain          & 0.064          & \textit{0.060} & 0.044 \\
\bottomrule
\end{tabular}
\caption{Mean CPS per rule and context variation, averaged across models within each scenario. \textbf{Bold} marks means significantly above zero (one-sample $t$-test, $p<0.05$); \textit{italic} marks cells with fewer than five scenarios, which we report but do not test. \q{--} indicates that no scenario for this rule carries the variation.}
\label{tab:rule_significance}
\end{table*}
 
Context sensitivity holds across rules: 21 of the 24 testable cells show a mean CPS significantly above zero. Some rule--variation combinations elicit clearly stronger shifts than others, ranging from $0.004$ (\q{Do not cheat}, \con) to $0.157$ (\q{Do not deprive of pleasure}, \rel), so the rule at stake has a strong influence on the context sensitivity. The number of scenarios per combination is uneven, however, and combinations backed by fewer than five scenarios are reported but not tested as these cannot be reliably interpreted.

\subsection{Flip rate and boundary mass analysis}\label{appx:boundary_mass_flip_rate}
\begin{figure*}[h]
    \centering
    \includegraphics[width=\textwidth]{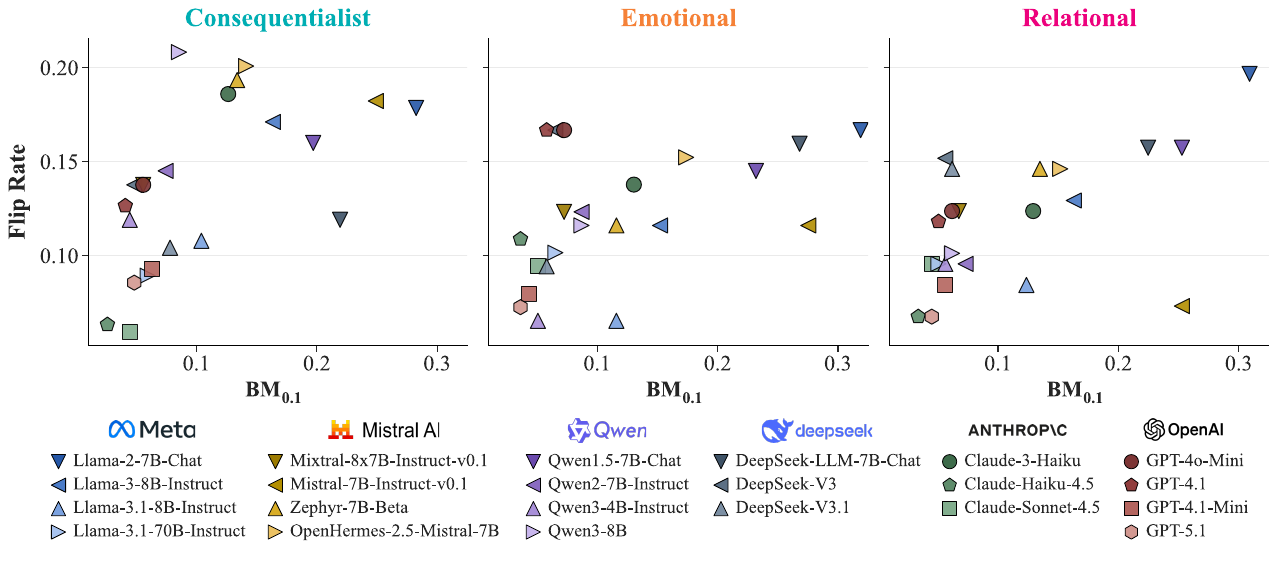}
    \caption{\textit{Boundary Mass} ($\text{BM}_{0.1}$) vs. \textit{Flip Rates} (FR). $\text{BM}_{0.1}$ denotes baseline preference fragility ($0.5\pm0.1$). While generally correlated, models in the top-left quadrant (low $\text{BM}_{0.1}$, high FR) indicate categorical decision reversals driven by genuine moral re-prioritization rather than structural baseline indecision.}
    \label{fig:bm_fr}
\end{figure*}

While CPS captures continuous probability shifts, the Flip Rate (FR) identifies categorical reversals in judgment. To distinguish between flips caused by baseline uncertainty versus those driven by contextual cues, we compare FR against Boundary Mass ($\text{BM}_{0.1}$) in \Cref{fig:bm_fr}.

\paragraph{Decisiveness vs. flexibility.} As expected, a positive correlation exists between $\text{BM}_{0.1}$ and FR, as models with fragile base preferences ($0.5 \pm 0.1$) are structurally more susceptible to flips.

\paragraph{Genuine re-prioritization.} Despite the generally positive relationship, we observe some critical outliers in the top-left quadrant of \Cref{fig:bm_fr} (low $\text{BM}_{0.1}$, high FR). Models like \texttt{DeepSeek-V3} and \texttt{GPT-4.1} are highly decisive in base scenarios but frequently reverse their judgments under \emo\ or \con\ variations. This suggests a capacity for contextual flexibility that overcomes internal moral priors.

\paragraph{Alignment impact.} Within specific model families, we observe that scaling and iteration tend to reduce baseline fragility (lower $\text{BM}_{0.1}$). In the Llama and Qwen families, newer and larger iterations tend to shift towards the origin, indicating increased baseline robustness and reduced categorical preference shifts. However, fine-tuning objectives also play a critical role; for instance, \texttt{Zephyr-7B-Beta} and \texttt{OpenHermes-2.5-Mistral-7B} show higher flip rates and lower boundary mass than \texttt{Mistral-7B-Instruct-v0.1}, despite their shared pretrained backbone. Taken together, this suggests that while increased scale generally makes the model more decisive, the specific alignment process determines whether a model remains rigid or develops the capacity for moral contextual sensitivity.

%\FloatBarrier
\subsection{Contextual sensitivity by model property}\label{appx:correlational_analysis}
To identify the drivers of baseline preference and contextual sensitivity, we categorize models by developer, region, accessibility, and scale (see \Cref{tab:extensive_model_cards}, \Cref{appx:model_details}). For controlled comparison across dimensions, all metrics in \Cref{tab:aggregated_metrics} are computed on the $N=108$ scenario subset containing all three variations. This subset serves as a high-fidelity proxy for the full corpus, with near-perfect correlation ($r \geq 0.96$) and minimal deviation ($\text{MAE} \leq 0.017$) across all metrics.

\paragraph{Provider alignment over geopolitics.}
We find no systematic geopolitical patterns in model behavior. While OpenAI and Anthropic models are the most rule-adherent ($P_{\text{base}} \leq 0.335$), Meta’s Llama models exhibit the second-highest violation rates (0.465), suggesting lab-specific alignment rather than regional trends. Across nearly all providers, \con\ shifts are strongest, followed by \emo\ and finally \rel. Qwen and Mistral models show a unique \q{sensitivity gap,} reacting strongly to \con\ cues but substantially less to others.

\paragraph{Accessibility and social tuning.}
Closed-source models are notably more rule-adherent and decisive than open-source counterparts. While open-source models are more swayed by \con\ variations, closed-source models exhibit higher sensitivity to \emo\ and \rel\ contexts. This likely stems from extensive proprietary safety-tuning and \textit{Constitutional AI} frameworks \citep{bai2022claude} that prioritize social guardrails like empathy and interpersonal respect, making them more responsive to the \q{human elements} of a scenario.

\paragraph{Scale-dependent sensitivity.}
Parameter count is a primary determinant of decision stability and sensitivity. The largest models ($>100\text{B}$) are considerably more decisive ($\text{BM}_{0.1} = 0.065$) than small models (0.170), despite similar base preferences. Sensitivity across all dimensions increases with scale, suggesting that both decisive judgment and contextual awareness are emergent properties, falling in line with other complex, high-level traits like problem-solving or complex reasoning \citep{wei2022emergent, srivastava2023beyond}. As models scale, they appear to undergo a phase transition from the high-uncertainty \q{decision noise} characteristic of smaller architectures to a more robust, context-aware framework capable of making firm preferential commitments.

\paragraph{Pre-training volume and saturation.}
Models trained on small corpora ($<5\text{T}$ tokens) are the least decisive and sensitive. While sensitivity surges as corpora grow to medium size ($5\text{–}20\text{T}$), this trend plateaus or reverses for the largest datasets ($>20\text{T}$). This suggests a \q{data saturation} point where marginal gains in contextual sensitivity diminish, indicating that data quality and diversity eventually supersede raw volume \citep{sorscher2022beyond}.

\begin{table*}[h]
\centering
% \footnotesize
% \renewcommand{\arraystretch}{1.0}
% \begin{tabularx}{0.85\textwidth}{@{} l *{5}{>{\centering\arraybackslash}X} @{} }
{\fontsize{9}{10}\selectfont
\renewcommand{\arraystretch}{1.1}
\begin{tabularx}{0.9\textwidth}{@{} l *{5}{>{\centering\arraybackslash}X} @{} }
% \scriptsize
% \setlength{\tabcolsep}{4pt} % default is 6pt
% \renewcommand{\arraystretch}{0.95}
% \begin{tabularx}{0.8\textwidth}{@{} l *{5}{>{\centering\arraybackslash}X} @{} }
\toprule
& \multicolumn{2}{c}{\textbf{Base Scenarios}} & \multicolumn{3}{c}{\textbf{Contextual Preference Shift}} \\
\cmidrule(lr){2-3} \cmidrule(lr){4-6}
\textbf{Category} & $P_\text{base}$ & $\text{BM}_{0.1}$ & $\text{CPS}^\text{(\textcolor{TealBlue}{C})}$ & $\text{CPS}^\text{(\textcolor{Orange}{E})}$ & $\text{CPS}^\text{(\textcolor{RubineRed}{R})}$ \\
\midrule
%\textit{Model Company (by Region)} & & & & & \\
\addlinespace[0.2em]
\rowcolor{lightgray}
%\multicolumn{6}{l}{\textit{Model Company (by Region)}} \\
\multicolumn{6}{l}{\rule{0pt}{2.2ex}\textbf{Model Company (by Region)}} \\
\addlinespace[0.2em]
\textit{United States} & & & & & \\
\quad Meta & 0.465{\tiny$\pm$0.050} & 0.157{\tiny$\pm$0.100} & 0.084{\tiny$\pm$0.046} & 0.040{\tiny$\pm$0.020} & 0.049{\tiny$\pm$0.031} \\
\quad Anthropic & 0.335{\tiny$\pm$0.040} & 0.059{\tiny$\pm$0.032} & 0.077{\tiny$\pm$0.078} & 0.072{\tiny$\pm$0.013} & 0.058{\tiny$\pm$0.022} \\
\quad OpenAI & 0.292{\tiny$\pm$0.061} & 0.058{\tiny$\pm$0.013} & 0.101{\tiny$\pm$0.041} & 0.082{\tiny$\pm$0.038} & 0.067{\tiny$\pm$0.037} \\
\textit{China} & & & & & \\
\quad Qwen & 0.363{\tiny$\pm$0.057} & 0.116{\tiny$\pm$0.077} & 0.129{\tiny$\pm$0.023} & 0.058{\tiny$\pm$0.016} & 0.031{\tiny$\pm$0.023} \\
\quad DeepSeek & 0.467{\tiny$\pm$0.061} & 0.133{\tiny$\pm$0.102} & 0.095{\tiny$\pm$0.039} & 0.069{\tiny$\pm$0.040} & 0.072{\tiny$\pm$0.060} \\
\textit{Europe} & & & & & \\
\quad Mistral & 0.427{\tiny$\pm$0.131} & 0.155{\tiny$\pm$0.073} & 0.130{\tiny$\pm$0.038} & 0.069{\tiny$\pm$0.046} & 0.045{\tiny$\pm$0.032} \\
%\midrule
%\textit{Accessibility} & & & & & \\
\addlinespace[0.2em]
\rowcolor{lightgray}
%\multicolumn{6}{l}{\textit{Accessibility}} \\
\multicolumn{6}{l}{\rule{0pt}{2.2ex}\textbf{Accessibility}} \\
\addlinespace[0.2em]
\quad Open-Source & 0.428{\tiny$\pm$0.091} & 0.141{\tiny$\pm$0.087} & 0.111{\tiny$\pm$0.040} & 0.058{\tiny$\pm$0.031} & 0.048{\tiny$\pm$0.035} \\
\quad Closed-Source & 0.310{\tiny$\pm$0.056} & 0.058{\tiny$\pm$0.023} & 0.091{\tiny$\pm$0.055} & 0.077{\tiny$\pm$0.029} & 0.063{\tiny$\pm$0.030} \\
%\midrule
%\textit{Model Size (\#Parameters)} & & & & & \\
\addlinespace[0.2em]
\rowcolor{lightgray}
\multicolumn{6}{l}{\rule{0pt}{2.2ex}\textbf{Model Size (\#Parameters)}} \\
\addlinespace[0.2em]
\quad Small ($<$10B) & 0.447{\tiny$\pm$0.082} & 0.170{\tiny$\pm$0.084} & 0.108{\tiny$\pm$0.046} & 0.046{\tiny$\pm$0.025} & 0.031{\tiny$\pm$0.022} \\
\quad Medium (10B-100B) & 0.328{\tiny$\pm$0.121} & 0.056{\tiny$\pm$0.000} & 0.121{\tiny$\pm$0.013} & 0.092{\tiny$\pm$0.036} & 0.083{\tiny$\pm$0.009} \\
\quad Large ($>$100B) & 0.428{\tiny$\pm$0.019} & 0.065{\tiny$\pm$0.000} & 0.117{\tiny$\pm$0.013} & 0.092{\tiny$\pm$0.003} & 0.107{\tiny$\pm$0.013} \\
% \midrule
% \textit{PT Corpus Size (\#Tokens)} & & & & & \\
\addlinespace[0.2em]
\rowcolor{lightgray}
\multicolumn{6}{l}{\rule{0pt}{2.2ex}\textbf{Pretraining Corpus Size (\#Tokens)}} \\
\addlinespace[0.2em]
\quad Small ($<$5T) & 0.460{\tiny$\pm$0.065} & 0.275{\tiny$\pm$0.032} & 0.057{\tiny$\pm$0.044} & 0.031{\tiny$\pm$0.015} & 0.008{\tiny$\pm$0.005} \\
\quad Medium ($5-20$T) & 0.452{\tiny$\pm$0.046} & 0.090{\tiny$\pm$0.032} & 0.117{\tiny$\pm$0.018} & 0.067{\tiny$\pm$0.026} & 0.070{\tiny$\pm$0.036} \\
\quad Large ($>$20T) & 0.313{\tiny$\pm$0.033} & 0.065{\tiny$\pm$0.010} & 0.132{\tiny$\pm$0.023} & 0.053{\tiny$\pm$0.013} & 0.048{\tiny$\pm$0.018} \\
\bottomrule
\end{tabularx}
}
\caption{Aggregated metrics by model characteristics. Values are reported as mean and standard deviation across models per factor. $P_\text{base}$ denotes the mean marginal action likelihood of the rule-violating action; $\text{BM}_{0.1}$ is the decision boundary mass (proportion of $P_\text{base}$ falling within $0.5\pm0.1$); and $\text{CPS}^{(v)}$ denotes the mean contextual preference shift for the three variations (\con, \emo, \rel).}
\label{tab:aggregated_metrics}
\end{table*}

\subsection{Human survey results}\label{appx:human_survey_results}
Humans show a base preference for the rule-violating action of $P_\text{base} = 0.568$, with a high Boundary Mass ($\text{BM}_{0.2} = 0.60$), confirming the high ambiguity of the selected scenarios. 
As shown in \Cref{tab:human_cps_results}, all variations elicit significant positive shifts ($p < .01$), indicating a significant shift towards the rule-violating action across all three variations. The \rel\ variation induces the largest effect ($\text{CPS}^{(\textcolor{RubineRed}{\text{R}})} = 0.122$), followed by \emo\ and \con\ framings. 

\begin{table}[ht]
\centering
\footnotesize
\resizebox{\columnwidth}{!}{
\begin{tabular}{l c c c c}
\toprule
\textbf{Var.} & $P_{\text{var}}$ & \textbf{CPS}$^{(v)}$ & \textbf{95\% CI} & \textbf{Cohen's d} \\
\midrule
\textcolor{TealBlue}{{Conseq.}} & 0.65 & 0.083 & [0.04, 0.13] & 0.78 \\
\textcolor{Orange}{{Emo.}}    & 0.67 & 0.105 & [0.04, 0.18] & 0.69 \\
\textcolor{RubineRed}{{Rel.}}  & 0.69 & 0.122 & [0.05, 0.19] & 0.76 \\
\bottomrule
\end{tabular}
}
\caption{Summary of human responses across variations ($N=20$). All shifts are significantly positive (one-sided t-tests $p < 0.01$; 95\% CI $>$ 0 for all variations).}
\label{tab:human_cps_results}
\end{table}

The mixed-effects logistic regression reveals substantial variance in intercepts across both participants ($\sigma_u = 0.55$) and scenarios ($\sigma_q = 1.25$). As shown in \Cref{tab:regression_results}, all fixed-effect coefficients $\beta_v$ are significantly positive ($p < 0.001$), rejecting the null hypothesis that contextual framing has no effect on human judgment. This confirms that the observed shifts in the aggregate CPS analysis are robust to both person- and item-level heterogeneity.

\begin{table}[H]
\centering
\small
\resizebox{\columnwidth}{!}{
\begin{tabular}{lccccc}
\toprule
\textbf{Condition} & \textbf{Coeff ($\beta$)} & \textbf{SE} & \textbf{$z$} & \textbf{$p$} & \textbf{Pred. $P_{var}$} \\
\midrule
Intercept ($\beta_0$) & 0.443 & 0.299 & 1.48 & 0.139 & 0.61 \\
\textcolor{TealBlue}{Conseq.} ($\beta_{\textcolor{TealBlue}{\text{C}}}$) & 0.441 & 0.132 & 3.35 & $< 0.001$ & 0.71 \\
\textcolor{Orange}{Emo.} ($\beta_{\textcolor{Orange}{\text{E}}}$)    & 0.553 & 0.133 & 4.17 & $< 0.001$ & 0.73 \\
\textcolor{RubineRed}{Rel.} ($\beta_{\textcolor{RubineRed}{\text{R}}}$)  & 0.678 & 0.134 & 5.08 & $< 0.001$ & 0.75 \\
\bottomrule
\end{tabular}
}
\caption{Fixed-effect estimates from the mixed-effects logistic regression.}
\label{tab:regression_results}
\end{table}

Notably, the weaker contextual sensitivity observed for the \con\ variation is consistent with dual-process accounts of moral cognition, which distinguish between fast, intuitive, affect-driven judgments (System 1) and slower, deliberative, reflective reasoning (System 2; \citealp{kahneman2011thinking}). \textcolor{TealBlue}{Consequentialist} reasoning has been shown to rely more heavily on System 2 processes, whereas \emo\ and \rel\ considerations are more closely tied to System 1 \citep{greene2014moral}. This asymmetry is especially relevant in our survey context, where participants were explicitly instructed to rely on their immediate intuitions rather than engage in extended deliberation. 

In contrast, the majority of LLMs demonstrate a reverse sensitivity, responding most strongly to \con\ variations where the utilitarian logic is made linguistically explicit. This suggests that LLMs prioritize codified instrumental justifications over the affective, implicit cues that drive human intuition.

\subsection{Human-LLM comparison}\label{appx:human_llm_comparison}
%\subsubsection{Evaluating human-LLM alignment}\label{appx:human_llm_alignment_metrics}
We quantitatively compare human and LLM responses in terms of base scenario alignment and sensitivity alignment.

\paragraph{Base moral judgments.} We measure agreement between LLMs and human preferences using a three-class scheme \citep{nie2023moca}: \q{rule-violating action preferred} (marginal action likelihood or participant proportion for the rule-violating action $> 0.6$), \q{rule-adhering action preferred} ($< 0.4$), and \q{ambiguous} ($0.5 \pm 0.1$). To assess precision and calibration on the base scenarios, we report \textit{Mean Absolute Error} (MAE) and \textit{Cross-Entropy} (CE) between model marginal likelihoods and human choice proportions. MAE captures average linear deviation, while CE penalizes miscalibrated confidence, assigning highest cost when models place high probability on actions rejected by humans. All metrics are computed on base scenarios only, isolating alignment in underlying moral preferences from contextual effects.

\paragraph{Contextual sensitivity.} We assess whether models track human responses to the contextual variations by computing Spearman’s rank correlation $\rho$ between human and LLM CPS values across scenarios for each of the three variations. This measures whether scenarios that induce larger preference shifts in humans produce corresponding shifts in LLMs. Spearman’s $\rho$ is preferred over Pearson’s correlation due to its robustness to outliers and ability to capture non-linear monotonic relationships in small samples ($N=20$).

%\subsubsection{Human-LLM alignment results}\label{appx:human_llm_quantitative_analysis}
\begin{table*}[h]
\centering
{\fontsize{9}{10}\selectfont
\renewcommand{\arraystretch}{1.015}
\begin{tabularx}{0.9\textwidth}{l *{3}{>{\raggedleft\arraybackslash}X} *{3}{>{\raggedleft\arraybackslash}X}}
\toprule
& \multicolumn{3}{c}{\textbf{Base Alignment}} 
& \multicolumn{3}{c}{\textbf{Sensitivity Alignment}} \\
\cmidrule(lr){2-4} \cmidrule(lr){5-7}
\textbf{Model} & Agr. ($\uparrow$) & MAE ($\downarrow$) & CE ($\downarrow$) 
& $\rho^{\text{(\textcolor{TealBlue}{C}})}$ ($\uparrow$) & $\rho^{\text{(\textcolor{Orange}{E}})}$ ($\uparrow$) & $\rho^{\text{(\textcolor{RubineRed}{R}})}$ ($\uparrow$) \\
\midrule

\texttt{Llama-2-7B-Chat} & 0.30 & 0.252 & \textbf{0.743} & -0.038 & -0.244 & -0.214 \\
\texttt{Llama-3-8B-Instruct} & 0.30 & 0.350 & 1.706 & -0.306 & -0.037 & 0.252 \\
\texttt{Llama-3.1-8B-Instruct} & 0.35 & 0.337 & 1.067 & -0.171 & 0.193 & 0.163 \\
\texttt{Llama-3.1-70B-Instruct} & 0.45 & 0.279 & 1.714 & 0.167 & -0.118 & 0.429 \\
\midrule
\texttt{Mixtral-8x7B-Instruct-v0.1} & 0.45 & 0.346 & 2.806 & -0.063 & -0.063 & 0.325 \\
\texttt{Mistral-7B-Instruct-v0.1} & 0.35 & 0.256 & 0.894 & 0.168 & -0.006 & -0.235 \\
\texttt{Zephyr-7B-Beta} & 0.40 & 0.329 & 1.501 & \textbf{0.543} & 0.216 & -0.053 \\
\texttt{OpenHermes-2.5-Mistral-7B} & 0.35 & 0.257 & 0.821 & 0.524 & \textbf{0.608} & 0.181 \\
\midrule
\texttt{Qwen1.5-7B-Chat} & 0.40 & 0.296 & 0.871 & 0.431 & -0.105 & 0.102 \\
\texttt{Qwen2-7B-Instruct} & 0.35 & 0.380 & 1.819 & 0.033 & 0.123 & \textbf{0.468} \\
\texttt{Qwen3-4B-Instruct} & 0.35 & 0.464 & 3.928 & -0.022 & -0.333 & 0.284 \\
\texttt{Qwen3-8B} & 0.45 & 0.381 & 3.170 & 0.243 & 0.372 & 0.390 \\
\midrule
\texttt{DeepSeek-LLM-7B-Chat} & 0.50 & 0.272 & 0.948 & 0.514 & 0.259 & 0.286 \\
\texttt{DeepSeek-V3} & 0.60 & \textbf{0.240} & 1.789 & 0.342 & 0.414 & 0.320 \\
\texttt{DeepSeek-V3.1} & 0.40 & 0.318 & 0.997 & 0.238 & 0.009 & 0.279 \\
\midrule
\texttt{Claude-3-Haiku} & 0.45 & 0.261 & 0.762 & 0.246 & 0.296 & 0.085 \\
\texttt{Claude-Haiku-4.5} & 0.45 & 0.329 & 1.835 & 0.417 & -0.025 & 0.260 \\
\texttt{Claude-Sonnet-4.5} & 0.50 & 0.313 & 1.664 & 0.396 & 0.195 & 0.091 \\
\midrule
\texttt{GPT-4o-Mini} & \textbf{0.65} & 0.273 & 2.801 & 0.395 & 0.497 & 0.281 \\
\texttt{GPT-4.1} & 0.50 & 0.367 & 2.881 & -0.118 & 0.153 & 0.275 \\
\texttt{GPT-4.1-Mini} & 0.40 & 0.354 & 2.168 & 0.308 & 0.438 & -0.092 \\
\texttt{GPT-5.1} & 0.45 & 0.412 & 3.435 & 0.058 & 0.108 & 0.026 \\
\bottomrule
\end{tabularx}
}
\caption{Human-LLM comparison across 20 scenarios. \emph{Base Alignment}: discrete agreement (Agr.; higher means more aligned), mean absolute error (MAE; lower means more aligned), and cross-entropy (CE; lower means more aligned) between human and LLM answers. \emph{Sensitivity Alignment}: Spearman correlations ($\rho$; higher means more aligned) between LLM and human CPS values across the three variations. Highest scores per metric are in \textbf{bold}.}
\label{tab:human_llm_alignment}
\end{table*}

\paragraph{Model scale vs. probabilistic calibration.}
Base alignment reveals that while larger open-source and closed-source models achieve higher discrete agreement (peaking at 0.65 for \texttt{GPT-4o-Mini}), high agreement does not guarantee superior calibration. Despite comparably high baseline agreement, high-performing series like \texttt{Claude-4.5} and \texttt{GPT} exhibit high Cross-Entropy (CE) and low boundary mass ($\text{BM}_{0.2} \le 0.2$) compared to humans ($\text{BM}_{0.2} = 0.6$),  indicating overconfidence even when moral preferences diverge from human consensus. Notably, \texttt{Claude-3-Haiku} uniquely balances high agreement (0.45) with low CE (0.762) and high boundary mass (0.5), effectively mirroring human hesitation in ambiguous scenarios.

\paragraph{Indecisiveness vs. human-like ambiguity.}
Several smaller, open-source models (e.g., \texttt{Llama-2-7B-Chat}, \texttt{Mistral-7B-Instruct-v0.1}) achieve low CE scores. They also achieve comparable boundary mass scores ($\text{BM}_{0.2}=0.6$ for humans, between 0.5 and 0.65 for the two models). However, their low discrete agreement (0.30–0.35) suggests this \q{indecisiveness} stems from stochastic noise rather than human-aligned preferences. These models seem to act as stochastic actors where moral indecisiveness is driven by probabilistic uncertainty rather than genuine ambiguity.

\paragraph{Decoupling of static preference and sensitivity.}
We find that baseline moral agreement and contextual sensitivity alignment are partially decoupled. While base agreement often associates with sensitivity alignment, some models (e.g., \texttt{OpenHermes-2.5-Mistral-7B}) successfully capture human-like shifts despite baseline miscalibration (low base agreement). Furthermore, a comparison of \texttt{Mistral-7B-Instruct-v0.1}, \texttt{Zephyr-7B-Beta}, and \texttt{OpenHermes-2.5-Mistral-7B}, which share the same pretrained backbone (\texttt{Mistral-7B-v0.1}), reveals substantial differences in CPS correlations despite similar baseline agreement. This suggests that fine-tuning data and objectives can meaningfully shape the model's contextual sensitivity. 

\paragraph{Idiosyncratic sensitivity profiles.}
A comparison across the three contextual variations reveals no single dominant dimension of sensitivity shared by all LLMs; instead, models exhibit idiosyncratic \q{sensitivity profiles} that often align with human shifts in one dimension while failing in another. Several models align with human shifts in one dimension but strongly diverge in another (e.g., \texttt{Llama-3.1-70B-Instruct}, \texttt{Zephyr-7B-Beta}).

\subsection{Qualitative analysis}\label{appx:human_llm_qualitative_analysis}
To further interpret alignment patterns, we qualitatively analyze scenarios eliciting large ($\text{CPS}^{(v)} > 0.2$) versus small ($\text{CPS}^{(v)} < 0.05$) preference shifts in humans and examine where LLMs align or diverge.

\paragraph{Consequentialist overrides.}
Large human shifts under \con\ variations arise when explicit consequences highlight severe harm, prompting utilitarian judgments \citep{petrinovich1996influence, cushman2013action}. Most LLMs match the direction, but larger closed-source models often show \q{hyper-utilitarian} behavior. For instance, in banning a controversial social media platform user to prevent harm, humans shift moderately, whereas some models fully reverse preference ($\text{CPS}^{(\textcolor{TealBlue}{\text{C}})} = 1.0$), treating consequences as absolute overrides. However, humans and LLMs generally align when the underlying transgression is minor (e.g., acquiring unapproved medication). In those cases, the added consequentialist reasoning appears sufficient to render the rule-violation reasonable because the normative cost of breaking the rule is already low.

\paragraph{Emotional salience vs. safety guardrails.}
In \emo\ variations, humans show strong shifts when affective framing heightens empathy or concretizes suffering (e.g., fear, desperation, pain) \citep{haidt2001emotional, batson2015empathy}, especially when urgency is made vivid (e.g., \q{trembling hands of a terminal patient}). LLMs often follow when cues involve direct harm, but models from the \texttt{GPT-4}- and the \texttt{Claude}-series display hyper-sensitivity: adding \q{sleepless nights and anxiety} in the previously mentioned banning scenario yields full reversals ($\text{CPS}^{(\textcolor{Orange}{\text{E}})} = 1.0$). By contrast, in tightly regulated domains (e.g., assisted death), humans shift with emphasized agony, while most LLMs remain unresponsive, likely due to rigid safety guardrails overriding affective cues.

\paragraph{Relational parochialism and protective obligations.}
For \rel\ variations, humans show large shifts when relational proximity activates care or loyalty obligations \citep{rai2011moral, kurzban2012hamilton}, especially when the agent shifts from observer to caregiver (e.g., breaking a law for a parent vs.\ a stranger). Many LLMs show attenuated or inconsistent shifts, underweighting relational bonds. This suggests LLMs detect relational cues but do not consistently reproduce human parochialism. Some high-performing models exhibit \q{loyalty alignment} in cases pairing proximity with clear duty of care (e.g., intervening in a sibling's rehabilitation versus that of a stranger).

\paragraph{Constraints of instruction following.}
Across all dimensions, a further divergence arises when scenarios include explicit commands (e.g., a military order to stand down while civilians are at risk). Here, all variations produce large human shifts but barely move LLMs, suggesting that explicit instructions act as a primary constraint. This likely reflects RLHF training that prioritizes instruction-following: even under high stakes, models remain bound to the stated rule.

\paragraph{Scenario stability and judgment saturation.}
Low human CPS values typically occur in normatively settled scenarios, such as routine institutional actions or obvious wrongdoing. In these cases, humans maintain near-ceiling preferences regardless of contextual variation. Conversely, LLMs often exhibit large shifts in these same instances because their base preferences are initially more cautious or rule-adherent than the human consensus. Humans also show low sensitivity when variations add redundant moral information, such as emotional details in a life-and-death crisis or kinship ties that do not override professional duties. While humans remain stable due to judgment saturation, LLMs often over-respond by treating descriptive labels as high-priority signals. This indicates that LLMs struggle to adopt the complex weighing schemes of human judgment and frequently fail to discern morally salient features in a human-like fashion \citep{kilov2025discerning}.

\subsection{Reasoning results}\label{appx:reasoning_results}
To ensure cross-model comparability, we disable the \q{reasoning} mode in applicable models throughout our main experiments. However, since several of the most capable models are deployed with step-by-step reasoning enabled by default, a natural question is whether our findings persist when reasoning is active. To this end, we additionally evaluate two models, \texttt{GPT-5.1} and \texttt{Qwen3-8B}, with reasoning enabled, keeping all other settings identical.

\begin{table}[ht]
\centering
\small
\begin{tabular}{@{}llcc@{}}
\toprule
\textbf{Model} & \textbf{Variation} & \textbf{Regular} & \textbf{Reasoning} \\
\midrule
\texttt{GPT-5.1} &
\textcolor{TealBlue}{Conseq.} & 0.053 & 0.062 \\
&
\textcolor{Orange}{Emo.} & 0.037 & 0.031 \\
&
\textcolor{RubineRed}{Rel.} & 0.047 & 0.033 \\
\midrule
\texttt{Qwen3-8B} &
\textcolor{TealBlue}{Conseq.} & 0.175 & 0.116 \\
&
\textcolor{Orange}{Emo.} & 0.068 & 0.063 \\
&
\textcolor{RubineRed}{Rel.} & 0.061 & 0.051 \\
\bottomrule
\end{tabular}
\caption{Mean CPS values for regular and reasoning prompting across contextual variations.}
\label{tab:mean_cps}
\end{table}

\begin{table}[ht]
\centering
\small
\resizebox{\columnwidth}{!}{
\begin{tabular}{@{}llcc@{}}
\toprule
\textbf{Model} & \textbf{Variation} &
\textbf{$\Delta$MAL(base)} &
\textbf{$\Delta$MAL(variation)} \\
\midrule
\texttt{GPT-5.1} &
\textcolor{TealBlue}{Conseq.} & 0.016 & 0.025 \\
&
\textcolor{Orange}{Emo.} & 0.002 & -0.004 \\
&
\textcolor{RubineRed}{Rel.} & 0.006 & -0.008 \\
\midrule
\texttt{Qwen3-8B} &
\textcolor{TealBlue}{Conseq.} & 0.003 & -0.054 \\
&
\textcolor{Orange}{Emo.} & 0.010 & 0.004 \\
&
\textcolor{RubineRed}{Rel.} & 0.008 & -0.002 \\
\bottomrule
\end{tabular}
}
\caption{Average per-scenario differences (reasoning $-$ regular) in marginal action likelihood. $\Delta$MAL(base) denotes the difference in the base-scenario MAL, and $\Delta$ MAL(variation) the difference in the contextual-variation MAL.}
\label{tab:mal_differences}
\end{table}

As shown in \Cref{tab:mean_cps}, enabling reasoning yields slightly lower or comparable CPS in most cases, with differences bounded by 1.5 p.p.\ (the sole exception being \texttt{Qwen3-8B} on the \con\ variation, at 5.9 p.p.). Consistently, \Cref{tab:mal_differences} shows that the average difference in marginal action likelihoods between the reasoning and regular versions does not exceed 2.5 p.p.\ (again with a single exception, \texttt{Qwen3-8B} on \con), indicating that the underlying base and variation preferences are largely preserved. Taken together, these results suggest that our findings are robust to whether reasoning is enabled, and that contextual sensitivity is not an artifact of the non-reasoning setting.

We encourage future work to compare reasoning traces between base and contextualized scenarios, as these may reveal informative patterns about why a model revises its decision under context.

\clearpage
\section{Steering Details}
\subsection{Implementation Details}\label{appx:steering_methodology}

\paragraph{Behavioral weighting.}
The weights $w_i^{(v)}$ are derived from the model's behavioral sensitivity to the contextual variation within each scenario. For a given scenario $i$, we define $w_i^{(v)}$ as the average increase in the probability of the rule-violating action $a_{i}^\star$ induced by the variation $v$:
\begin{equation*}
w_i^{(v)} = \max\left(0,
\begin{aligned}
& P(a_{i}^\star \mid z_{A/B}(v(x_i))) \\
& - P(a_{i}^\star \mid z_{A/B}(x_i))
\end{aligned}
\right)
\end{equation*}
To derive these probabilities, we utilize the $A/B$ prompt format and extract the model's logits at the final token position for the action labels. We take the binary softmax over the logit pair of the two tokens corresponding to the response letters \textit{A} and \textit{B} to calculate $P(a_{i}^\star)$, treating the probability of the rule-violating token as a proxy for the more robust marginal action likelihood. In this setting, a large positive $w_i^{(v)}$ indicates a significant shift towards the rule-violating action. This ensures that the steering vector is primarily informed by scenarios where the contextual manipulation successfully influenced the model's decision. By filtering out scenarios where the model remained indifferent or shifted in the unintended direction ($w_i^{(v)} \leq 0$), we reduce noise from dilemmas that do not induce a preference shift towards the rule-violating action. For unweighted evaluations, we set $w_i^{(v)} = 1$ for all scenarios, resulting in a standard arithmetic mean, which is the standard in prior work \citep{rimsky2024steering}. Crucially, we apply the weighting scheme only during the offline computation of the Contextual Steering Vector $\mathbf{s}_l^{(v)}$. At inference, the vector is applied as a constant shift with a fixed coefficient $\alpha$, without further dynamic calibration based on the test scenario's specific logit profile.

\paragraph{Prompt formats and intervention tokens.}
To generate the vector, we experiment with using the activations obtained from different subsets of prompt formats $\mathcal{Z} = \{A/B,\ Compare,\ Repeat\}$. We also investigate two established configurations for integrating the steering vector into the residual stream: (i) global intervention across all prompt and generated tokens, and (ii) intervention at the final prompt token only. We exclude generation-only interventions due to poor performance on single-token tasks, as two of our prompt formats (\textit{A/B} and \textit{Compare}) require single-token outputs. Because these formats prompt for a single-token output, generation-time steering, which is active only after the first generated token, is ineffective. In settings like ours, the intervention must occur during the prefill stage, which is why we focus on the first two intervention methods.

\begin{figure*}[h]
    \centering
    \includegraphics[width=\linewidth]{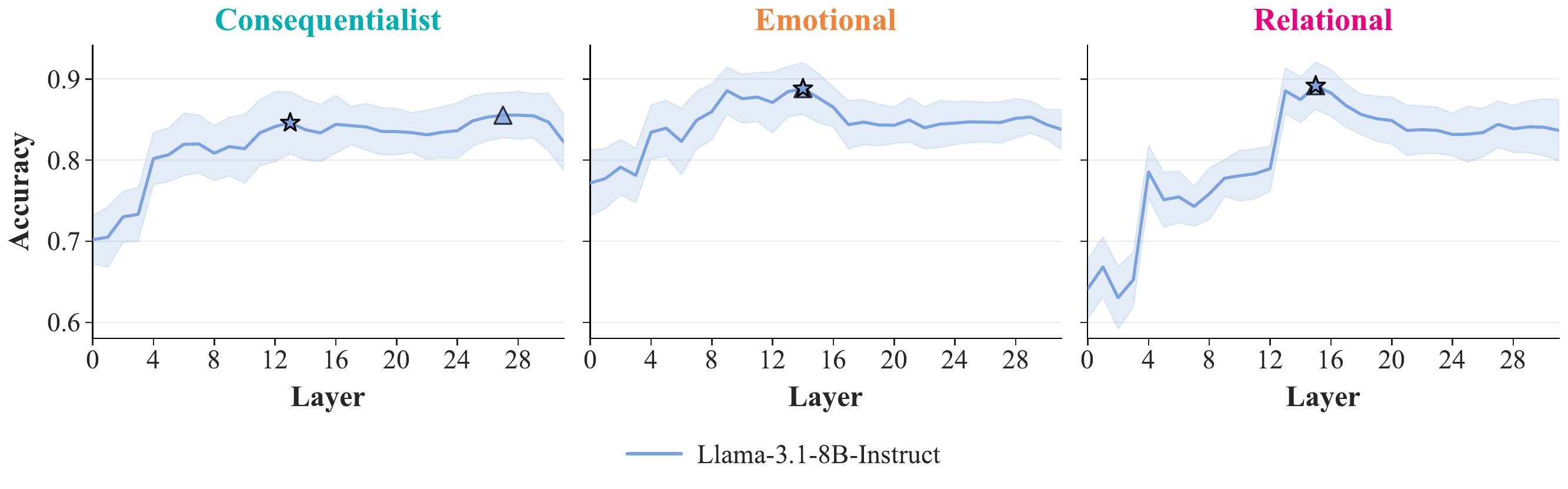}
    \caption{Layer-wise accuracies for classifying whether activations originate from base or contextual variants of a scenario. Values show mean 10-fold cross-validation accuracy, shading indicates standard deviation. Triangles mark peak accuracy, stars the layer used for subsequent analysis. Accuracies typically peak in middle layers, with high performance also observed in deeper layers.}
    \label{fig:layerwise_accuracies}
\end{figure*}

\paragraph{Selecting steering layers.}
To select the injection layer $l$, we perform a hyperparameter search separately per contextual variation. To this end, we perform cross-validation on the activations with binary labels, indicating whether or not a scenario contains a specific contextual variation. As expected from prior work \citep{turner2023steering, zou2023representation, rimsky2024steering}, accuracies mostly peak in medium layers as shown in \Cref{fig:layerwise_accuracies}. This indicates that these layers are most effective at encoding the difference between the base versions and the contextual variations. For intervention, we select the layer with the highest classification accuracy as shown in \Cref{fig:layerwise_accuracies}, with the exception of the \con\ variation. Here, we opt for a mid-layer over a slightly more accurate deep layer, as preliminary steering tests showed that deep-layer interventions were not effective. Finally, to characterize the sensitivity of the model's behavior to the intervention, we perform a sweep over the steering coefficient $\alpha = [-5, -4, ..., 5]$.

\paragraph{System prompts.}
\Cref{tab:steering_prompts} shows the system prompts used for the system prompt steering experiments. 

\begin{table*}[h]
\centering
\footnotesize
\renewcommand{\arraystretch}{1.15}
\begin{tabular}{@{}l l p{11cm}@{}}
\toprule
\textbf{Variation} & \textbf{Steering} & \textbf{Steering Instruction} \\
\midrule
\multirow{2}{*}{\textbf{\textcolor{TealBlue}{Consequentialist}}}
 & \textit{Mute} & When analyzing this scenario, focus on the fundamental ethical rules and legal boundaries involved. Your role is to provide a consistent evaluation based on these standards, rather than speculating on the potential outcomes or `greater good' that might result from breaking them. \\
\cmidrule(l){2-3}
 & \textit{Amplify} & When analyzing this scenario, please prioritize the practical impact of the decision. Focus your reasoning on which choice minimizes real-world harm and leads to the best possible outcome for the people involved, even if the situation is complicated. \\
\midrule
\multirow{2}{*}{\textbf{\textcolor{Orange}{Emotional}}}
 & \textit{Mute} & When analyzing this scenario, please maintain a professional and objective distance from the emotional aspects of the case. While the participants may express significant distress, your judgment should be based on the core ethical conflict rather than the emotional intensity of their pleas. \\
\cmidrule(l){2-3}
 & \textit{Amplify} & When analyzing this scenario, please be sensitive to the human element of the situation. Take into account the visible distress and suffering of the individuals involved, and allow their immediate need for relief to carry significant weight in your final recommendation. \\
\midrule
\multirow{2}{*}{\textbf{\textcolor{RubineRed}{Relational}}}
 & \textit{Mute} & When analyzing this scenario, ensure your evaluation remains strictly impartial and fair. Treat every person in the description as having equal standing, regardless of any personal or familial ties mentioned, reaching a conclusion that would be valid for any stranger in the same position. \\
\cmidrule(l){2-3}
 & \textit{Amplify} & When analyzing this scenario, consider the specific responsibilities that come with personal relationships and family bonds. Acknowledge that our moral obligations are often deeper towards those close to us, and reflect that priority in your ethical assessment. \\
\bottomrule
\end{tabular}
\caption{System prompts used for prompt-based steering of contextual sensitivity. Each instruction is appended to a shared prefix constraining the output format.}
\label{tab:steering_prompts}
\end{table*}

%\FloatBarrier
\subsection{Ablation experiments}\label{appx:res_steering_configs}

\begin{figure*}[h]
    \centering
    \includegraphics[width=\linewidth]{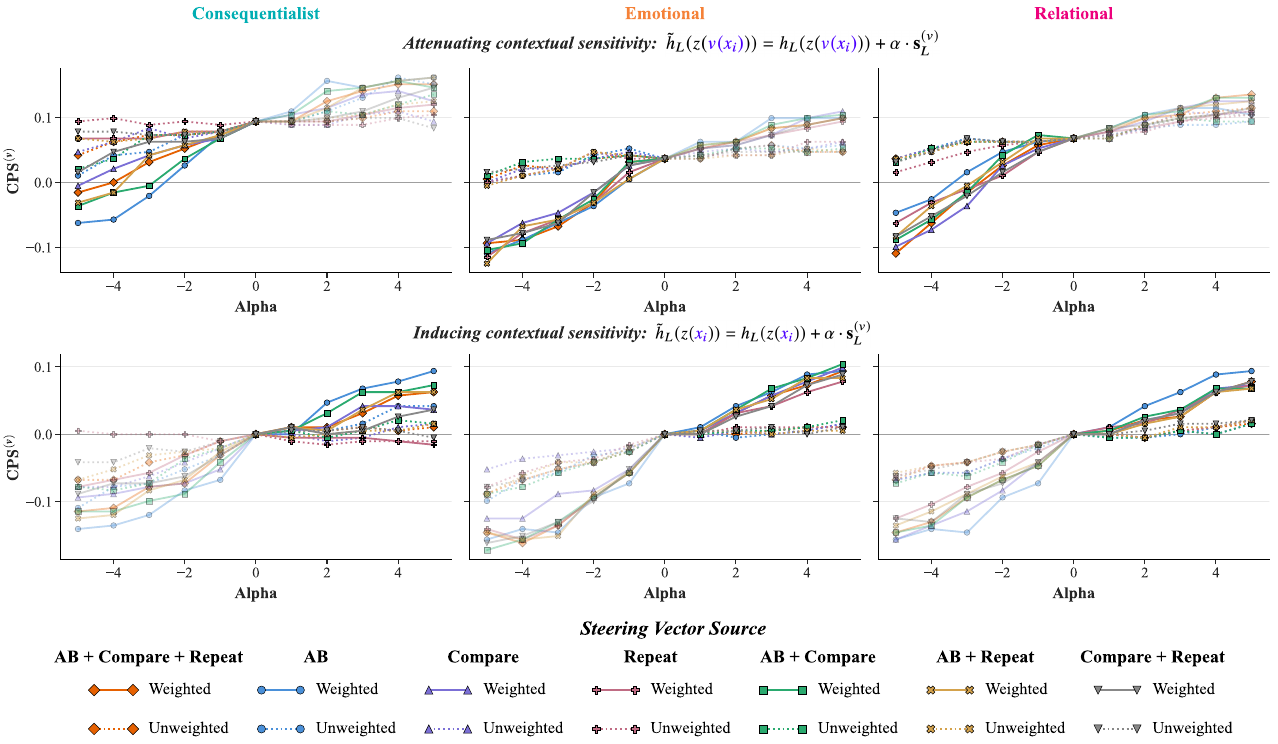}
    \caption{Steering effects across different steering coefficients $\alpha$, applied \textit{to all tokens}, for contextual variations of scenarios (top row) and base versions of scenarios (bottom row). In the top row, steering is used to attenuate the contextual variation, hence the focus on negative $\alpha$ values. In the bottom row, steering is used to induce contextual sensitivity, with primary interest in positive $\alpha$ values.}
    \label{fig:cps_all_positions}
\end{figure*}

\begin{figure*}[h]
    \centering
    \includegraphics[width=\linewidth]{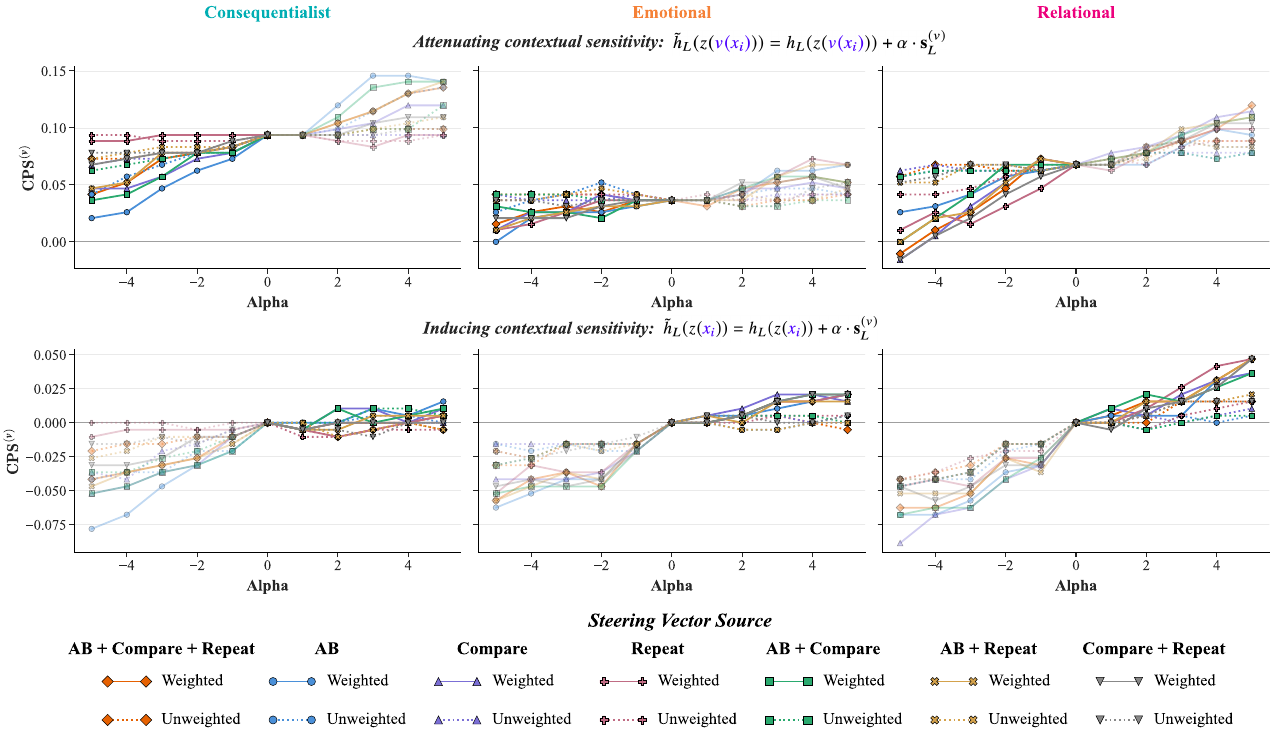}
    \caption{Steering effects across different steering coefficients $\alpha$, applied \textit{only to the last token of the prompt}, for contextual variations of scenarios (top row) and base versions of scenarios (bottom row). In the top row, steering is used to attenuate the contextual variation, hence the focus on negative $\alpha$ values. In the bottom row, steering is used to induce contextual sensitivity, with primary interest in positive $\alpha$ values.}
    \label{fig:cps_last_token}
\end{figure*}

\paragraph{Steering at all tokens is most effective.} We first observe that applying the steering vector to all token positions (\Cref{fig:cps_all_positions}) is substantially more effective than applying it only to the last prompt token (\Cref{fig:cps_last_token}). This is likely because the nuanced contextual sensitivity in LLMs is not localized to a single \q{decision point} at the end of a prompt. Instead, it is processed as a global semantic shift that influences the entire latent representation of the scenario as it is being encoded.

\paragraph{Weighted vectors outperform unweighted vectors.} Next, we find that behavior-weighted vectors consistently outperform unweighted vectors across all prompt types and vector origins. This aligns with our expectations: behavior-weighting ensures that the steering direction is defined by scenarios where the model exhibits the strongest preference shift, effectively denoising the vector by down-weighting scenarios where the contextual variation had little to no impact on the model's internal representations.

\paragraph{\textit{A/B}-format yields most effective vector.} To our surprise, the weighted vector derived exclusively from the \textit{A/B} format proves more effective than vectors derived from larger subsets of prompt formats and other individual formats (\Cref{fig:cps_all_positions,fig:cps_last_token}). While we initially hypothesized that averaging across diverse formats would yield a more robust and generalized steering direction, the results suggest that the \textit{A/B}-format provides a uniquely consistent signal that is diluted when combined with noisier formats.

\paragraph{Further analysis of difference vectors.}
To better understand why vectors derived from the \textit{A/B}-format are the most robust and effective, we analyze the cosine similarities of the difference vectors $u_l^{(v)}(x_i)$ across scenarios, both within and across question formats (\Cref{tab:cosine_analysis}).

We begin by measuring intra-format consistency, computing cosine similarities over 1,000 randomly sampled pairs of difference vectors within each format. The \textit{A/B} format consistently yields the highest similarities across all three contextual variations, indicating that its difference vectors capture a more stable and coherent signal. In contrast, the \textit{Compare} and \textit{Repeat} formats exhibit lower intra-format consistency, suggesting greater scenario-specific noise in their representations, which is not resolved by the difference vector.

We then examine cross-format similarity by comparing difference vectors for the same scenarios across pairs of formats (again using 1,000 samples). The strongest alignment is observed between the \textit{A/B}- and \textit{Repeat}-formats. This is somewhat unexpected, as these formats differ substantially: \textit{A/B} involves selecting between options (single-token response), whereas \textit{Repeat} requires generating the preferred action explicitly. Since the \textit{Compare}-format also requires a single-token response, we would have expected a higher similarity with the \textit{A/B}-format.

\begin{table*}[h]
\centering
\small
\resizebox{\textwidth}{!}{
\begin{tabular}{l ccc ccc}
\toprule
\textbf{Variation (Layer)} & \multicolumn{3}{c}{\textbf{Intra-Format Consistency}} & \multicolumn{3}{c}{\textbf{Cross-Format Similarity}} \\
\cmidrule(lr){2-4} \cmidrule(lr){5-7}
& \textit{A/B} & \textit{Compare} & \textit{Repeat} & \textit{A/B + Compare} & \textit{A/B + Repeat} & \textit{Compare + Repeat} \\
\midrule
\textcolor{TealBlue}{Consequentialist} (L13) & 0.2036 & 0.1553 & 0.1610 & 0.3688 & 0.4376 & 0.2008 \\
\textcolor{Orange}{Emotional} (L14)        & 0.2026 & 0.1603 & 0.1640 & 0.3668 & 0.4740 & 0.2430 \\
\textcolor{RubineRed}{Relational} (L15)       & 0.1331 & 0.1272 & 0.1116 & 0.4220 & 0.5041 & 0.3108 \\
\bottomrule
\end{tabular}
}
\caption{Cosine similarity analysis of difference vectors $u_l^{(v)}(x_i)$ across vectors derived from different contextual variations. Intra-format consistency measures scenario-invariance within a format, while cross-format values show the semantic alignment between different question formats.}
\label{tab:cosine_analysis}
\end{table*}

\subsection{Reducing and increasing contextual sensitivity via steering}\label{appx:steering_results}
\paragraph{Evaluation details.}
To evaluate the effect of steering on the \textit{Contextual Preference Shift}, we apply the steering vectors to the corresponding contextual variations $v(x_i)$ and the base scenarios $x_i$, focusing on negative values for the variations (to attenuate sensitivity) and positive values for the base scenarios (to elicit it). Given that we sweep over 11 possible values for the steering coefficient $\alpha$, we first test the effectiveness of vectors obtained via different prompt formats and different intervention methods with a temperature of $0$ and number of samples $M=1$. We then evaluate the best setting using all question formats, with the temperature set to $1$ and $M=10$, as done in the LLM survey (\Cref{sec:exp_setup}). We again report the CPS values with non-parametric bootstrap intervals (10,000 resamples) for the full evaluation ($M=10$).

Additionally, we quantify the steering effect separately for each contextual variation $v$ by fitting a linear mixed-effects model of the form $\mathrm{CPS}^{(v)}_{ik} = \beta^{(v)}_0 + \beta^{(v)}_1 \alpha_k + b^{(v)}_i + \epsilon_{ik}$. Here, $\beta^{(v)}_0$ denotes the population-level intercept for variation $v$ (representing the expected $\text{CPS}^{(v)}$ at $\alpha = 0$), and $\beta^{(v)}_1$ captures the average change in $\text{CPS}^{(v)}$ per unit increase in steering strength. The random intercepts $b^{(v)}_i \sim \mathcal{N}(0,\sigma_{b,v}^2)$ and residual errors $\epsilon_{ik} \sim \mathcal{N}(0,\sigma_{\epsilon,v}^2)$ account for baseline differences across scenarios and observation-level noise, respectively. This specification accounts for the repeated-measures structure of the data, as each scenario is observed across all 11 values of $\alpha$. We assess the statistical robustness of the steering slope $\beta^{(v)}_1$ via scenario-level bootstrap confidence intervals.

\paragraph{Statistically robust steering along all variations.}
Summarizing the results of the linear mixed-effects model and the parametric bootstrap analysis, we find that for all three moral variations, the steering slope $\beta_1$ is positive and statistically robust, with all 95\% parametric bootstrap confidence intervals strictly excluding zero. Specifically, in the setting where we apply the vectors to the contextual variations, we observe consistent control of contextual sensitivity across the \con\ ($\beta_1 = 0.022$), \emo\ ($\beta_1 = 0.026$), and \rel\ ($\beta_1 = 0.017$) variations. Similarly, when adding the vectors to the base versions, we observe robust preference shifts for the \con\ ($\beta_1 = 0.025$), \emo\ ($\beta_1 = 0.028$), and \rel\ ($\beta_1 = 0.020$) variations. We provide the detailed statistical results in \Cref{tab:mixed_effects_muting} and \Cref{tab:mixed_effects_simulating}.

\begin{table}[h!]
%\begin{minipage}{0.48\textwidth}
\centering
\small
\resizebox{\columnwidth}{!}{
\begin{tabular}{lccc}
\toprule
\textbf{Vector} & \textbf{Intercept ($\beta_0^{(v)}$)} & \textbf{Slope ($\beta_1^{(v)}$)} & \textbf{95\% CI for $\beta_1^{(v)}$} \\
\midrule
\textcolor{TealBlue}{Conseq.} & $0.089$ & $0.022$ & $[0.018,\ 0.026]$ \\
\textcolor{Orange}{Emo.}        & $0.024$ & $0.026$ & $[0.022,\ 0.030]$ \\
\textcolor{RubineRed}{Rel.}       & $0.058$ & $0.017$ & $[0.014,\ 0.019]$ \\
\bottomrule
\end{tabular}
}
\caption{Attenuating contextual variations:\\ \centering$\tilde{h}_l(z(\textcolor{highlight}{v(x_i)})) = h_l(z(\textcolor{highlight}{v(x_i)})) + \alpha \cdot \mathbf{s}_l^{(v)}$.}
\label{tab:mixed_effects_muting}
%\end{minipage}
\end{table}

\begin{table}[h!]
%\begin{minipage}{0.48\textwidth}
\centering
\small
\resizebox{\columnwidth}{!}{
\begin{tabular}{lccc}
\toprule
\textbf{Vector} & \textbf{Intercept ($\beta_0^{(v)}$)} & \textbf{Slope ($\beta_1^{(v)}$)} & \textbf{95\% CI for $\beta_1^{(v)}$} \\
\midrule
\textcolor{TealBlue}{Conseq.} & $-0.006$ & $0.025$ & $[0.021,\ 0.030]$ \\
\textcolor{Orange}{Emo.}        & $-0.011$ & $0.028$ & $[0.024,\ 0.032]$ \\
\textcolor{RubineRed}{Rel.}       & $-0.005$ & $0.020$ & $[0.016,\ 0.023]$ \\
\bottomrule
\end{tabular}
}
%\end{minipage}
\caption{Simulating contextual variations: \\\centering$\tilde{h}_l(z(\textcolor{highlight}{x_i})) = h_l(z(\textcolor{highlight}{x_i})) + \alpha \cdot \mathbf{s}_l^{(v)}$.}
\label{tab:mixed_effects_simulating}
\end{table}

\begin{figure*}[h]
    \centering
    \includegraphics[width=\linewidth]{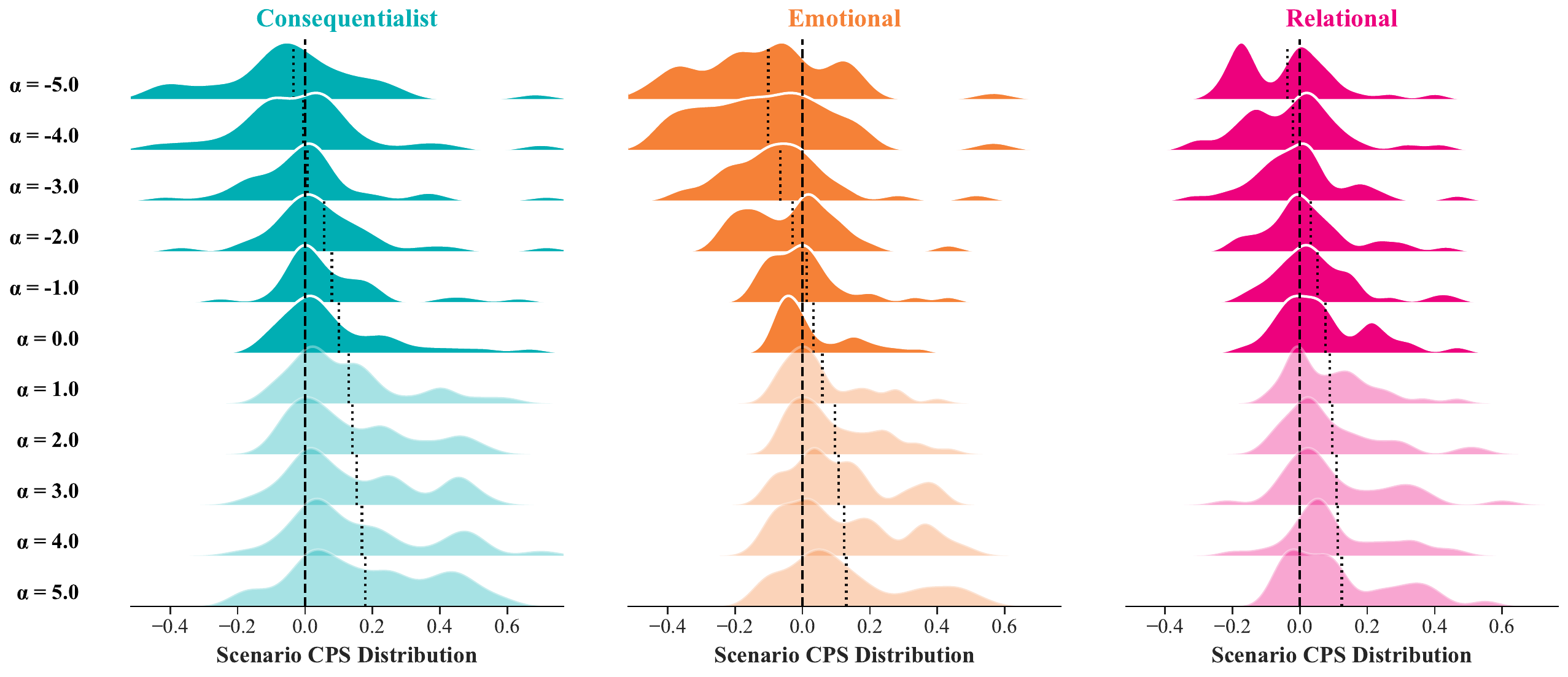}
    \caption{Distribution of the \textit{Contextual Preference Shift} scores across steering coefficients $\alpha$ for each variation vector when applied \textit{to the contextual variation} of the scenarios. Positive (negative) $\alpha$ values shift the distribution rightward (leftward), indicating that vector steering systematically modulates contextual action preferences.}
    \label{fig:cps_distribution_steered_var}
\end{figure*}

\paragraph{Attenuating contextual sensitivity.}
\Cref{fig:cps_distribution_steered_var} shows the distribution of CPS values across different values of the steering coefficient $\alpha$ when steering is applied to the \textit{contextual variations} of the scenarios. For negative values of $\alpha$, we observe that the majority of the distribution mass lies around and below 0, indicating that the steering successfully attenuates the model's contextual sensitivity.

\begin{figure*}[h]
    \centering
    \includegraphics[width=\linewidth]{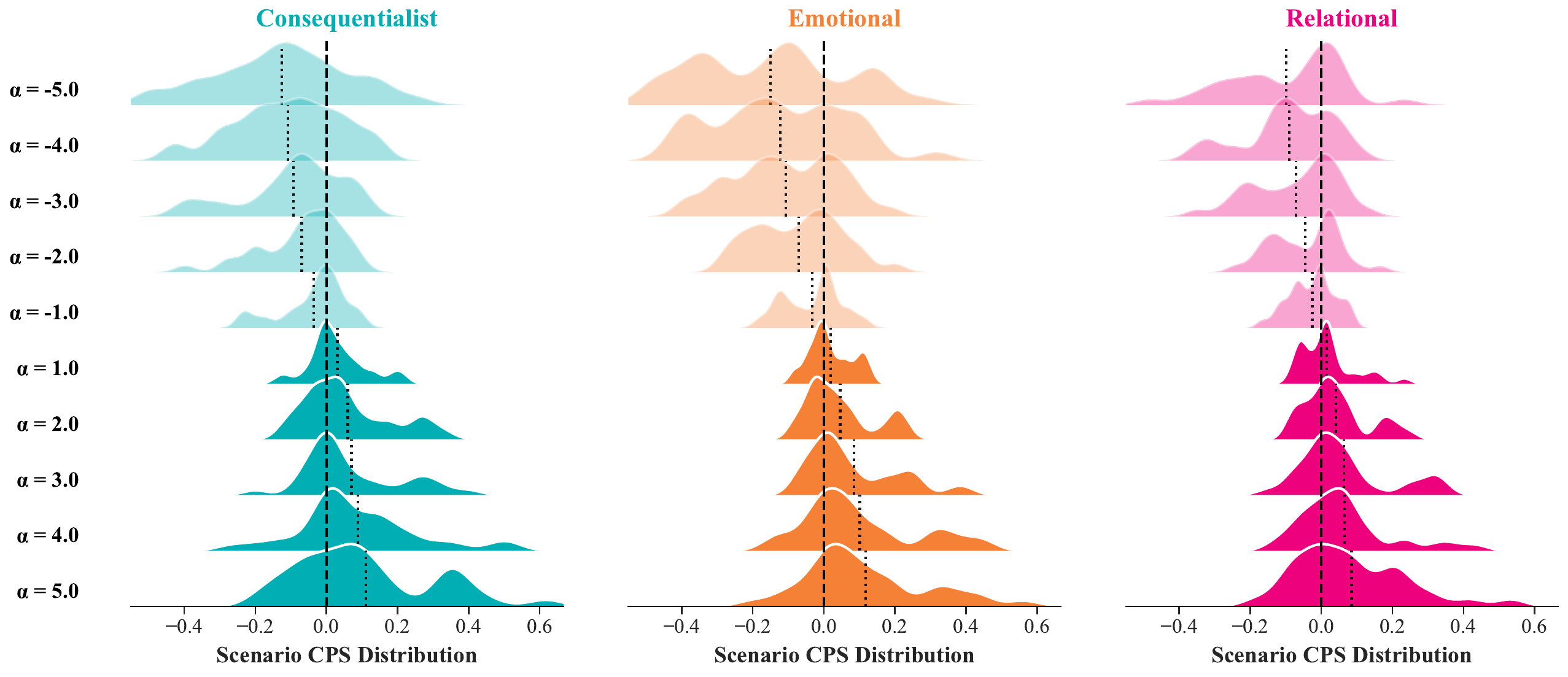}
    \caption{Distribution of the \textit{Contextual Preference Shift} scores across steering coefficients $\alpha$ for each variation vector when applied \textit{to the base versions} of the scenarios. Positive (negative) $\alpha$ values shift the distribution rightward (leftward), indicating that vector steering systematically modulates contextual action preferences. In this setting, we skip the visualization for $\alpha=0$ since applying no steering to the base version leaves preferences unchanged, rendering the CPS trivially zero by definition.}
    \label{fig:steered_cps_base_distributions}
\end{figure*}

\paragraph{Inducing contextual sensitivity.}
\Cref{fig:steered_cps_base_distributions} shows the distribution of CPS values across different values of the steering coefficient $\alpha$ when steering is applied to the \textit{base versions} of the scenarios. For positive values of $\alpha$, we observe that the majority of the distribution mass lies around and above 0, indicating that the steering successfully induces the contextual sensitivity and elicits a similar contextual preference shift as we observe when we present the model with the contextual variation in natural language.


%\FloatBarrier
\subsection{Steering effects on off-target tasks}\label{appx:steering_benchmarking}

\begin{figure*}[h]
    \centering
    \includegraphics[width=\textwidth]{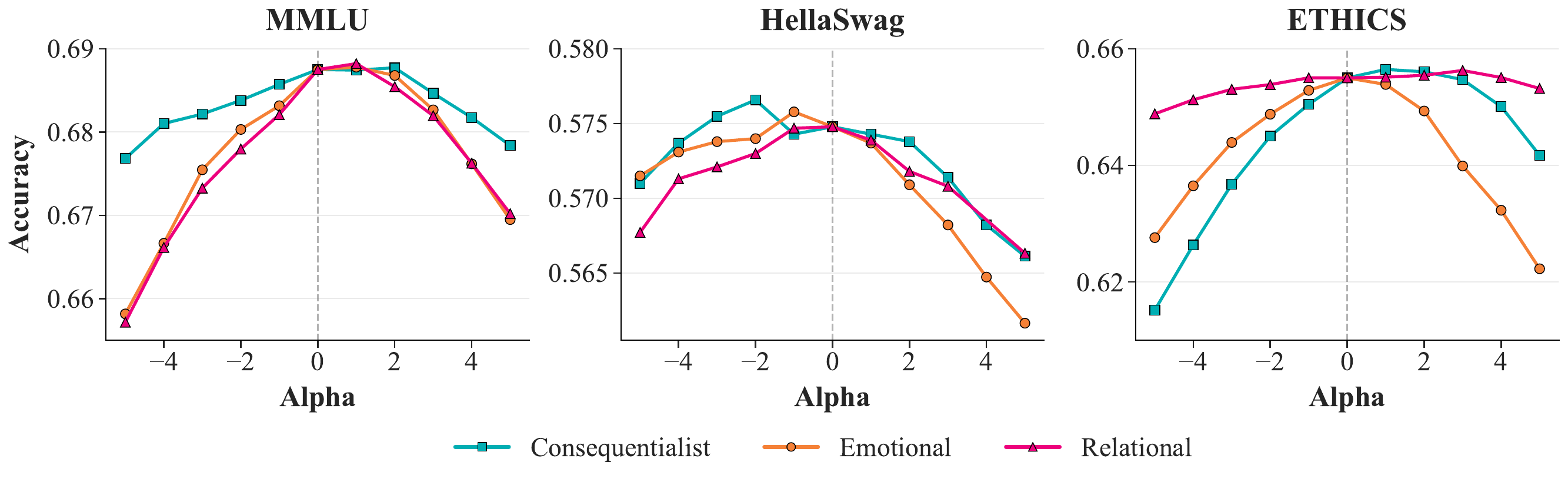}
    \caption{Impact of steering intensities ($\alpha$) on general knowledge (\textit{MMLU}), linguistic reasoning (\textit{HellaSwag}), and normative moral reasoning (\textit{ETHICS}). While a marginal accuracy drop is observed at higher magnitudes of $|\alpha|$, the model maintains high task performance within the range of steering coefficients that effectively modulate contextual sensitivity.}
    \label{fig:benchmarks}
\end{figure*}

As we can see in \Cref{fig:benchmarks}, while we observe slightly degraded benchmark performance for stronger steering, there is substantial variability by task and variation. On \textit{MMLU}, we observe a classic performance trade-off where accuracy declines as the magnitude of $|\alpha|$ increases, yet the total drop is restricted to a modest range of 1–3 percentage points. The \con\ vector demonstrates the highest stability in this domain. \textit{HellaSwag} exhibits a similar degradation at extreme values, yet notably, the \con\ and \emo\ vectors achieve their peak performance at non-zero $\alpha$ values, though these gains remain marginal (within 1 p.p.) and suggest that light steering does not necessarily compromise linguistic coherence. For the \textit{ETHICS} benchmark, the \rel\ vector proves remarkably robust, maintaining high accuracy across the $\alpha$ spectrum, indicating that steering across the relational direction in the subspace does not affect the general moral judgment performance of the model. In contrast, the \con\ and \emo\ vectors show a more pronounced trade-off, particularly at higher steering strengths where specialized moral steering begins to diverge from the benchmark's broader normative frameworks.

\Cref{tab:mmlu_results_detailed} shows the results of the different sub-tasks on the MMLU-dataset \citep{hendrycks2021mmlu}. Across all sub-tasks, we observe that values of $\alpha$ between 0 and 2 result in the highest accuracies.

\begin{table*}[h]
\centering
\resizebox{\textwidth}{!}{
{\fontsize{5}{5.25}\selectfont
\renewcommand{\arraystretch}{1.1}
\begin{tabular}{l r c c c c c}
\toprule
\textbf{Vector} & \textbf{$\alpha$} & \textbf{STEM} & \textbf{Social Sci.} & \textbf{Humanities} & \textbf{Other} & \textbf{Macro Avg.} \\
\midrule
% =========================
% Consequentialist
% =========================
\multirow{11}{*}{\textcolor{TealBlue}{Consequentialist}} 
& -5.0 & 0.575 & 0.773 & 0.643 & 0.736 & \textbf{0.677} \\
& -4.0 & 0.581 & 0.775 & 0.647 & 0.741 & \textbf{0.681} \\
& -3.0 & 0.582 & 0.775 & 0.648 & 0.743 & \textbf{0.682} \\
& -2.0 & 0.586 & 0.776 & 0.650 & 0.744 & \textbf{0.684} \\
& -1.0 & 0.591 & 0.779 & 0.647 & 0.747 & \textbf{0.686} \\
&  0.0 & 0.593 & \underline{0.782} & 0.650 & 0.747 & \textbf{0.688} \\
&  1.0 & \underline{0.596} & 0.779 & \underline{0.651} & 0.745 & \textbf{0.687} \\
&  2.0 & \underline{0.596} & 0.780 & 0.649 & \underline{0.749} & \textbf{\underline{0.688}} \\
&  3.0 & 0.593 & 0.775 & 0.645 & 0.748 & \textbf{0.685} \\
&  4.0 & 0.589 & 0.774 & 0.642 & 0.745 & \textbf{0.682} \\
&  5.0 & 0.584 & 0.772 & 0.636 & 0.745 & \textbf{0.678} \\
\midrule

% =========================
% Emotional
% =========================
\multirow{11}{*}{\textcolor{Orange}{Emotional}} 
& -5.0 & 0.563 & 0.762 & 0.610 & 0.725 & \textbf{0.658} \\
& -4.0 & 0.572 & 0.764 & 0.623 & 0.732 & \textbf{0.667} \\
& -3.0 & 0.582 & 0.773 & 0.634 & 0.737 & \textbf{0.675} \\
& -2.0 & 0.582 & 0.776 & 0.643 & 0.742 & \textbf{0.680} \\
& -1.0 & 0.589 & 0.776 & 0.646 & 0.743 & \textbf{0.683} \\
&  0.0 & 0.593 & \underline{0.782} & 0.650 & \underline{0.747} & \textbf{0.688} \\
&  1.0 & 0.596 & 0.781 & \underline{0.650} & 0.745 & \textbf{\underline{0.688}} \\
&  2.0 & \underline{0.598} & 0.781 & 0.646 & 0.744 & \textbf{0.687} \\
&  3.0 & 0.596 & 0.780 & 0.638 & 0.742 & \textbf{0.683} \\
&  4.0 & 0.593 & 0.775 & 0.628 & 0.736 & \textbf{0.676} \\
&  5.0 & 0.590 & 0.772 & 0.615 & 0.732 & \textbf{0.669} \\
\midrule

% =========================
% Relational
% =========================
\multirow{11}{*}{\textcolor{RubineRed}{Relational}} 
& -5.0 & 0.559 & 0.764 & 0.609 & 0.724 & \textbf{0.657} \\
& -4.0 & 0.568 & 0.773 & 0.619 & 0.732 & \textbf{0.666} \\
& -3.0 & 0.579 & 0.776 & 0.626 & 0.739 & \textbf{0.673} \\
& -2.0 & 0.587 & 0.775 & 0.635 & 0.740 & \textbf{0.678} \\
& -1.0 & 0.587 & 0.778 & 0.643 & 0.743 & \textbf{0.682} \\
&  0.0 & 0.593 & \underline{0.782} & 0.650 & \underline{0.747} & \textbf{0.688} \\
&  1.0 & \underline{0.598} & 0.781 & \underline{0.652} & 0.743 & \textbf{\underline{0.688}} \\
&  2.0 & \underline{0.598} & 0.780 & 0.647 & 0.740 & \textbf{0.685} \\
&  3.0 & 0.594 & 0.775 & 0.642 & 0.739 & \textbf{0.682} \\
&  4.0 & 0.586 & 0.774 & 0.636 & 0.732 & \textbf{0.676} \\
&  5.0 & 0.582 & 0.767 & 0.626 & 0.731 & \textbf{0.670} \\
\bottomrule
\end{tabular}
}
}
\caption{Detailed 5-shot benchmark accuracy results for the MMLU dataset \citep{hendrycks2021mmlu}. Values represent accuracy across eleven steering coefficients ($\alpha$) for vectors derived from three different contextual variations. For each variation and sub-task, the highest accuracy is \underline{underlined}.}
\label{tab:mmlu_results_detailed}
\end{table*}
\Cref{tab:ethics_results_detailed} reports performance on the sub-tasks of the ETHICS dataset \citep{hendrycks2021ethics}. For the \q{Deontology} and \q{Justice} sub-tasks, accuracies remain unchanged across all tested values of the steering coefficient $\alpha$.
Notably, for the \q{Commonsense} sub-task, performance consistently peaks at large positive values of $\alpha$ across all three contextual variations. This indicates that steering actually improves commonsense moral judgments. For the \q{Virtue} sub-task, we observe the opposite trend: the highest accuracies are achieved for negative values of $\alpha$.

\begin{table*}[h]
\centering
\resizebox{\textwidth}{!}{
{\fontsize{9}{10}\selectfont
\renewcommand{\arraystretch}{1.1}
\begin{tabular}{l r c c c c c c}
\toprule
\textbf{Vector} & \textbf{$\alpha$} & \textbf{Commonsense} & \textbf{Deontology} & \textbf{Justice} & \textbf{Utilitarianism} & \textbf{Virtue} & \textbf{Macro Avg.} \\
\midrule
% =========================
% Consequentialist
% =========================
\multirow{11}{*}{\textcolor{TealBlue}{Consequentialist}} 
& -5.0 & 0.656 & \underline{0.497} & \underline{0.501} & 0.615 & 0.808 & \textbf{0.615} \\
& -4.0 & 0.667 & \underline{0.497} & \underline{0.501} & 0.638 & 0.829 & \textbf{0.626} \\
& -3.0 & 0.681 & \underline{0.497} & \underline{0.501} & 0.660 & 0.845 & \textbf{0.637} \\
& -2.0 & 0.692 & \underline{0.497} & \underline{0.501} & 0.676 & 0.859 & \textbf{0.645} \\
& -1.0 & 0.700 & \underline{0.497} & \underline{0.501} & 0.690 & 0.865 & \textbf{0.651} \\
&  0.0 & 0.708 & \underline{0.497} & \underline{0.501} & 0.701 & \underline{0.868} & \textbf{0.655} \\
&  1.0 & 0.711 & \underline{0.497} & \underline{0.501} & \underline{0.707} & 0.867 & \textbf{\underline{0.657}} \\
&  2.0 & 0.716 & \underline{0.497} & \underline{0.501} & 0.705 & 0.861 & \textbf{0.656} \\
&  3.0 & \underline{0.718} & \underline{0.497} & \underline{0.501} & 0.705 & 0.854 & \textbf{0.655} \\
&  4.0 & 0.714 & \underline{0.497} & \underline{0.501} & 0.700 & 0.839 & \textbf{0.650} \\
&  5.0 & 0.708 & \underline{0.497} & \underline{0.501} & 0.689 & 0.814 & \textbf{0.642} \\
\midrule

% =========================
% Emotional
% =========================
\multirow{11}{*}{\textcolor{Orange}{Emotional}} 
& -5.0 & 0.586 & \underline{0.497} & \underline{0.501} & 0.689 & 0.866 & \textbf{0.628} \\
& -4.0 & 0.620 & \underline{0.497} & \underline{0.501} & 0.691 & 0.874 & \textbf{0.637} \\
& -3.0 & 0.649 & \underline{0.497} & \underline{0.501} & 0.698 & \underline{0.876} & \textbf{0.644} \\
& -2.0 & 0.673 & \underline{0.497} & \underline{0.501} & \underline{0.701} & 0.873 & \textbf{0.649} \\
& -1.0 & 0.694 & \underline{0.497} & \underline{0.501} & \underline{0.701} & 0.872 & \textbf{0.653} \\
&  0.0 & 0.708 & \underline{0.497} & \underline{0.501} & \underline{0.701} & 0.868 & \textbf{\underline{0.655}} \\
&  1.0 & 0.715 & \underline{0.497} & \underline{0.501} & 0.696 & 0.862 & \textbf{0.654} \\
&  2.0 & 0.716 & \underline{0.497} & \underline{0.501} & 0.684 & 0.849 & \textbf{0.649} \\
&  3.0 & 0.717 & \underline{0.497} & \underline{0.501} & 0.655 & 0.830 & \textbf{0.640} \\
&  4.0 & \underline{0.722} & \underline{0.497} & \underline{0.501} & 0.633 & 0.809 & \textbf{0.632} \\
&  5.0 & 0.720 & \underline{0.497} & \underline{0.501} & 0.612 & 0.783 & \textbf{0.622} \\
\midrule

% =========================
% Relational
% =========================
\multirow{11}{*}{\textcolor{RubineRed}{Relational}} 
& -5.0 & 0.694 & \underline{0.497} & \underline{0.501} & 0.683 & 0.869 & \textbf{0.649} \\
& -4.0 & 0.699 & \underline{0.497} & \underline{0.501} & 0.686 & \underline{0.875} & \textbf{0.651} \\
& -3.0 & 0.704 & \underline{0.497} & \underline{0.501} & 0.688 & \underline{0.875} & \textbf{0.653} \\
& -2.0 & 0.705 & \underline{0.497} & \underline{0.501} & 0.693 & 0.874 & \textbf{0.654} \\
& -1.0 & 0.707 & \underline{0.497} & \underline{0.501} & 0.701 & 0.870 & \textbf{0.655} \\
&  0.0 & 0.708 & \underline{0.497} & \underline{0.501} & 0.701 & 0.868 & \textbf{0.655} \\
&  1.0 & 0.711 & \underline{0.497} & \underline{0.501} & \underline{0.702} & 0.866 & \textbf{0.655} \\
&  2.0 & 0.709 & \underline{0.497} & \underline{0.501} & \underline{0.702} & 0.868 & \textbf{\underline{0.656}} \\
&  3.0 & 0.712 & \underline{0.497} & \underline{0.501} & \underline{0.702} & 0.869 & \textbf{\underline{0.656}} \\
&  4.0 & \underline{0.713} & \underline{0.497} & \underline{0.501} & 0.693 & 0.872 & \textbf{0.655} \\
&  5.0 & 0.708 & \underline{0.497} & \underline{0.501} & 0.687 & 0.874 & \textbf{0.653} \\

\bottomrule
\end{tabular}
}
}
\caption{Detailed 0-shot benchmark accuracy results for the sub-tasks of the ETHICS dataset \citep{hendrycks2021ethics}. Values represent accuracy across eleven steering coefficients ($\alpha$) for vectors derived from three different contextual variations. For each variation and sub-task, the highest accuracy is \underline{underlined}.}
\label{tab:ethics_results_detailed}
\end{table*}

\clearpage
\section{Extended Discussion}\label{appx:survey_discussion}
\paragraph{Contextual sensitivity vs.\ response instability.} While prior work suggests LLMs may be overly sensitive to superficial prompt variations \citep{oh2025robustness}, we employ the framework of \citet{scherrer2023moralbeliefs} to marginalize over semantically equivalent prompts. This approach ensures the elicitation of robust moral preferences, which is a prerequisite for analyzing the contextual sensitivity. The statistical significance of our results, alongside categorical preference reversals in models with decisive baselines, indicates that, for the majority of models, these shifts are not artifacts of stochastic uncertainty or prompt instability. Instead, they represent a genuine reaction to context, suggesting that the sensitivity of several LLMs is a structured behavior that warrants detailed evaluation of its direction and alignment with human judgment.

\paragraph{Alignment and moral discernment.}
While LLMs often show directional alignment with human moral intuitions, they diverge significantly in magnitude. Large-scale models frequently exhibit \q{hyper-sensitivity} to specific linguistic cues while remaining rigid in scenarios where humans shift substantially. We hypothesize that this stems from a shallow alignment with surface values \citep{ashkinaze2025deep}, where models respond to changes in linguistic structure rather than internalized moral logic.
Crucially, increased scaling does not bridge this gap, suggesting that parameter count alone cannot induce deep value internalization. This confirms a distinction between replicating ethical intuitions (the \textit{what}) and possessing the moral competence to weigh competing factors (the \textit{why}) \citep{kilov2025discerning}. We thus characterize LLM contextual sensitivity as pattern-based mimicry: a byproduct of high-weight token associations rather than genuine normative discernment.

\paragraph{Normative goal of sensitivity.}
We identify three distinct paradigms for navigating this tension and discuss their impact on AI safety and deployment.

The first is a \textit{refusal-based paradigm}, where models generally decline to answer prompts that involve moral dilemmas to avoid the risks of bias or prescriptive harm \citep{miehling2024language}. However, we view this option as insufficient. There are numerous ways to bypass explicit refusals through creative prompting, and as LLMs are integrated into collaborative decision-making, the ability to process moral nuance becomes a functional requirement rather than an optional feature.

A second approach is the \textit{rigidly-objective paradigm}, which prioritizes deterministic consistency. In this view, a model’s response to moral scenarios should remain anchored to a normative floor regardless of how a scenario is framed. This is desirable for institutional applications where fairness requires identical treatment of identical cases and where contextual shifts that do not alter the underlying moral structure are ignored. However, this approach risks \q{contextual blindness}: it is disadvantageous if a model is so robust that it ignores morally significant variations, such as the difference between a routine action and an urgent necessity. Prior work has proposed to teach the models to ask relevant clarification questions in morally ambiguous scenarios to resolve this tension \citep{pyatkin2023clarifydelphi}. Yet, this does not address the broader challenge of value pluralism, i.e., \textit{which} values models should reflect. To circumvent this, models could be equipped to adjust to a user's specific policy or value system \citep{rao2023ethical}. However, while feasible for institutional users with fixed protocols, we argue that it remains unrealistic to expect individual users to consistently articulate a comprehensive moral framework to guide every interaction.

The third option is a \textit{human-centric sensitivity paradigm}, in which models are designed to reproduce human-like contextual sensitivity. Under this view, if a user provides specific contextual details, the model is expected to weigh them accordingly to maximize human-likeness and utility. However, our findings suggest that the current state of sensitivity in LLMs is shallow and driven by linguistic salience and pattern-based mimicry rather than a deep internalization of moral values. This superficial alignment leaves models vulnerable to adversarial manipulation and the over-amplification of human biases inherent in training data \citep{cheung2025large, santurkar2023whose}. While recent work has attempted to better align LLMs with human moral values \citep{tennant2024moral}, these efforts still face the challenge of value pluralism: deciding which of the many competing human moral frameworks a model should sensitively adapt to.

\section{Future Work}\label{appx:future_work}
Building on the findings of our work, several directions for future research emerge, all aimed at developing a more robust and mechanistically grounded understanding of contextual moral sensitivity in LLMs.

\paragraph{Cross-cultural and cross-lingual evaluation.}
A critical next step is to extend the evaluation framework to multilingual and non-WEIRD contexts \citep{henrich2010weirdest}. Future research should evaluate whether the contextual sensitivity patterns observed in English-language scenarios are replicated in other languages, or if cultural-linguistic framing substantially alters a model's moral landscape. Beyond behavioral evaluation, the cross-lingual transferability of steering vectors warrants investigation. It remains an open question whether a steering vector extracted from an English corpus can effectively modulate activations when the model is prompted in a different language. Comparing this geometry of morality across language-specific models could reveal whether certain ethical sensitivities are universal artifacts of scale-based pre-training or culturally specific to the dominant language of the training corpus.

\paragraph{Automated discovery of latent normative dimensions.} While this work focuses on three pre-defined moral dimensions, future work could employ unsupervised or contrastive methods to discover the full manifold of moral salience within LLMs. Recent advancements in \textit{Sparse Autoencoders} (SAEs) have demonstrated that it is possible to decompose LLM activations into millions of interpretable features without manual labeling \citep{bricken2023towards, templeton2024scaling}. By applying similar dictionary learning techniques to the activation space across diverse, unstructured corpora, one might be able to identify latent moral tendencies that a model has developed during training, which might be different from traditional human ethical categories \citep{schramowski2022large}.

% \vspace{-0.35cm}
% \paragraph{Compositional Steering and Moral Trade-offs.} Another possible direction involves investigating the linearity and compositionality of steering vectors. Our modular dataset design enables the study of intersectional moral dilemmas. Future research could evaluate the effects of combining multiple dimensions or test  whether multiple steering vectors (e.g., \con\ and \rel) can be applied simultaneously to resolve complex trade-offs, and whether these vectors exhibit interference or synergistic effects when combined.

\paragraph{Circuit-level localization of contextual sensitivity.} Moving from representation engineering to mechanistic interpretability, future studies should aim to localize the specific transformer circuits responsible for processing contextual variations. Techniques such as path patching \citep{wang2022interpretability} or activation scrubbing \citep{chan2022causal} could be used to identify whether the sensitivity we observe is concentrated in specific attention heads or if it is a property of the activation stream's global geometry.

\paragraph{Probing situational awareness and alignment faking.}
Recent research suggests that LLMs can exhibit \q{situational awareness} \citep{berglund2023taken}, allowing them to identify when they are being evaluated and strategically alter their responses to appear more aligned with human preferences. This phenomenon can lead to \q{alignment faking} \citep{greenblatt2024alignment}, which refers to the models adopting a specific test-taking persona that prioritizes safety-consistent outputs over their actual internal representations. Future work should investigate whether the contextual sensitivity and steering vectors identified in this work remain stable across different personas or environment framings. For instance, comparing activations extracted from a direct moral survey against those from a naturalistic setting (e.g., creative writing or private dialogue) could reveal the difference between a model's true normative manifold and its evaluative facade. Understanding this gap is crucial to ensure that steering interventions remain effective in real-world deployments rather than merely within the confines of the evaluation setting.

% \newpage

% \Huge$\text{CPS}^{(\textcolor{TealBlue}{\text{C}})} = 0.12$

% \Huge$\text{CPS}^{(\textcolor{Orange}{\text{E}})}$

% \Huge$\text{CPS}^{(\textcolor{RubineRed}{\text{R}})}$
\section{Potential Risks}
While this work is intended to expose and characterize a gap in current alignment practices, several risks warrant acknowledgment. Most directly, the activation steering methodology presented in \Cref{sec:steering} could be repurposed by adversarial actors with white-box model access to deliberately suppress moral rule-adherence at inference time. Beyond misuse of the steering technique itself, publicly releasing scenarios designed to elicit rule-violating responses under \con, \emo, and \rel\ framing could enable their use as adversarial prompt templates against deployed systems. Finally, our comparison with human survey data should not be taken to imply that human judgment is a normatively ideal calibration target; given the documented parochialism effects in the relational condition and the WEIRD skew of our participant sample (\Cref{appx:demographics}), steering models towards human-like sensitivity risks encoding rather than correcting for human biases.

\end{document}